\documentclass{article}

    \usepackage[nonatbib, final]{neurips_2021}

\usepackage[utf8]{inputenc} 
\usepackage[T1]{fontenc}    
\usepackage{hyperref}       
\usepackage{url}            
\usepackage{booktabs}       
\usepackage{amsfonts}       
\usepackage{nicefrac}       
\usepackage{microtype}      
\usepackage{xcolor}         
\usepackage{multicol}
\usepackage{wrapfig}

\usepackage{enumitem}
\usepackage[T1]{fontenc}
\usepackage{tgpagella}
\usepackage{graphicx,amsmath,amsfonts,amscd,amssymb,bm,url,color,latexsym}
\usepackage{fullpage}
\usepackage[small,bf]{caption}
\usepackage{subfigure}
\usepackage{bbm}
\usepackage{microtype}
\usepackage{algorithm}
\usepackage{algorithmicx}
\usepackage{algpseudocode} 
\usepackage{hyperref}
\usepackage{verbatim}
\usepackage{framed}
\usepackage{cancel}

\algnewcommand{\LineComment}[1]{\State\textit{ \(\#\) #1}}

\newtheorem{theorem}{Theorem}[section]
\newtheorem{assumption}{Assumption}[section]
\newtheorem{lemma}[theorem]{Lemma}

\newtheorem{claim}[theorem]{Claim}

\usepackage{tikz}
\usepackage{fge}

\usepackage{mathdots} 

\renewcommand{\mathbf}{\boldsymbol}

\newcommand{\mb}{\mathbf}
\newcommand{\mc}{\mathcal}

\newcommand{\bb}{\mathbb}

\newcommand{\reals}{\bb R}

\newcommand{\R}{\reals}

\newcommand{\paren}[1]{ \left( #1 \right) }

\DeclareMathOperator{\sign}{sign}

\DeclareMathOperator{\prox}{prox}

\newcommand{\wh}{\widehat}

\newcommand{\innerprod}[2]{\left\langle #1,  #2 \right\rangle}

\numberwithin{equation}{section}

\def \endprf{\hfill {\vrule height6pt width6pt depth0pt}\medskip}
\newenvironment{proof}{\noindent {\bf Proof} }{\endprf\par}

            \newcommand{\rootpcp}{$\sqrt{\text{PCP}}$ }

           \newcommand{\stableU}{$\text{StablePCP}_u$}
           \newcommand{\stableC}{$\text{StablePCP}_c$}

\newcommand\subsetsim{\mathrel{%
  \ooalign{\raise0.2ex\hbox{$\subset$}\cr\hidewidth\raise-0.8ex\hbox{\scalebox{0.9}{$\sim$}}\hidewidth\cr}}}
  
  \newcommand\supsetsim{\mathrel{%
  \ooalign{\raise0.2ex\hbox{$\supset$}\cr\hidewidth\raise-0.8ex\hbox{\scalebox{0.9}{$\sim$}}\hidewidth\cr}}}

\definecolor{royalpurple}{rgb}{0.37, 0.22, 0.57}
\definecolor{lava}{rgb}{0.81, 0.06, 0.13}

\title{Square Root Principal Component Pursuit:\\
Tuning-Free Noisy Robust Matrix Recovery}

\author{%
  Junhui Zhang \\
  Department of Applied Physics and Applied Math\\
  Columbia University\\
  New York, NY 10027 \\
  \texttt{jz2903@columbia.edu} \\
   \And
   Jingkai Yan \\
   Department of Electrical Engineering \\
   Columbia University\\
   New York, NY 10027 \\
   \texttt{jy2927@columbia.edu} \\
   \AND
   John Wright \\
   Department of Electrical Engineering \\
   Columbia University\\
   New York, NY 10027 \\
   \texttt{jw2966@columbia.edu} \\
}

\begin{document}

\maketitle

\begin{abstract}
 We propose a new framework -- Square Root Principal Component Pursuit -- for low-rank matrix recovery from observations corrupted with noise and outliers. Inspired by the square root Lasso, this new formulation does not require prior knowledge of the noise level. We show that a single, universal choice of the regularization parameter suffices to achieve reconstruction error proportional to the (a priori unknown) noise level. In comparison, previous formulations such as stable PCP rely on noise-dependent parameters to achieve similar performance, and are therefore challenging to deploy in applications where the noise level is unknown. We validate the effectiveness of our new method through experiments on simulated and real datasets. Our simulations corroborate the claim that a universal choice of the regularization parameter yields near optimal performance across a range of noise levels, indicating that the proposed method outperforms the (somewhat loose) bound proved here. 
\end{abstract}

\section{Introduction}\label{sec:introduction}

The problem of recovering a low-rank matrix from unreliable observations arises in a wide range of engineering applications, including collaborative filtering \cite{su2009survey}, latent semantic indexing \cite{hofmann1999probabilistic}, image and video analysis \cite{haeffele2014structured, zhang2013hyperspectral, ji2010robust} and so on. 
This problem can be formalized in terms of the following observation model: given an observation $\mb D$ which is a superposition
\begin{equation}
\mb D = \underset{\color{lava}\text{\bf low-rank}}{\mb L_0} + \underset{\color{lava}\text{\bf sparse}}{\mb S_0} + \underset{\color{lava}\text{\bf noise}}{ \mb Z_0 }.
\end{equation}
of an unknown low-rank matrix $\mb L_0$, sparse corruptions $\mb S_0$ and dense noise $\mb Z_0$, our goal is to accurately estimate both $\mb L_0$ and $\mb S_0$. 

This model has been intensely studied, leading to algorithmic theory for methods based on both convex and nonconvex optimization \cite{candes2011robust, netrapalli2014non, chen2020bridging}. One virtue of the convex approach is that in the noise-free setting ($\mb Z_0 = \mb 0$), it is possible to exactly recover a broad range of low-rank and sparse pairs $(\mb L_0,\mb S_0)$, with a universal choice of regularization parameters, which does not depend on either the rank or sparsity. This makes it possible to deploy this method in a ``hands-free'' manner, provided the dataset of interest indeed has low-rank and sparse structure. 

In the presence of noise, however, the situation becomes more complicated: all efficient, guaranteed estimators require knowledge of the noise level (or the rank and sparsity) \cite{chen2020bridging, zhou2010stable, chen2015fast}. This is problematic, since in most applications the noise level is not known ahead of time. In standard convex formulations, the appropriate regularization parameter depends on the noise standard deviation, leaving the user with a painful and time-consuming task of tuning these parameters on a per-dataset basis.

Motivated by this issue, we revisit this classical matrix recovery problem. The main contribution of this paper is the proposal and analysis of a new formulation for robust matrix recovery, which stably recovers $\mb L_0$ and $\mb S_0$ without requiring prior knowledge of the rank, sparsity, or noise level. In particular, our approach admits a single, universal choice of regularization parameters, which under standard hypotheses on $\mb L_0$ and $\mb S_0$, yields an estimation error proportional to the noise standard deviation $\sigma$. To our knowledge, our method and analysis are the first to achieve this.

Our approach is based on a combination of two natural ideas. For matrix recovery, we draw on the {\em stable principal component pursuit} \cite{zhou2010stable}, a natural convex relaxation, which minimizes a combination of the nuclear norm of $\mb L$, the $\ell_1$ norm of $\mb S$ and the squared Frobenius norm $\|\mb Z\|_F^2$ of the noise. This is a principled approach to handling both the structured components $\mb L_0, \mb S_0$ and the noise: $\| \mb Z\|_F^2$ can be motivated naturally from the negative log-likelihood of the gaussian distribution. Moreover, under mild assumptions on the rank and singular vectors of $\mb L_0$ and the sparsity pattern of $\mb S_0$, the reconstruction error of stable PCP is $O(\|\mb Z_0\|_F)$ \cite{zhou2010stable}. 

On the other hand, optimally balancing these terms requires knowledge of the standard deviation $\sigma$ of the true noise distribution. To address this issue, we draw inspiration from the square root Lasso \cite{belloni2011square}. The square root Lasso is a sparse estimator which achieves minimax optimal estimation with a universal choice of parameters, which does not depend on the noise level. The core idea is very simple: instead of penalizing the squared Frobenius norm $\| \mb Z \|_F^2$ of the noise, one penalizes its square root, $\| \mb Z \|_F$. We call the resulting formulation {\em square root principal component pursuit} (\rootpcp). Our new formulation has the benefit that with a \textit{noise-independent} universal choice of regularization parameters, essentially the same level of reconstruction error can be achieved. This makes  \rootpcp a more practical approach to low-rank recovery in unknown noise. 

Due to the square root term, the objective function is no longer smooth or differentiable, and so we cannot apply algorithms such as the proximal gradient method\footnote{which requires the objective to be the sum of a smooth and a non-smooth function}. Nevertheless, our new formulation remains convex and separable, i.e. the objective is the sum of functions of different variables, making Alternating Direction Method of Multipliers (ADMM) a suitable solver \cite{boydADMM}. We test our new formulation with ADMM on both simulated data and real data in image processing. The experimental results show the effectiveness of our proposed new formulation in recovering the low-rank and the sparse matrix, and suggest that \rootpcp has better performance than anticipated by our loose upper bound of the reconstruction error. 

\subsection{Notations and Assumptions}

We use $\|\mb A\|$, $\|\mb A\|_F$, and $\|\mb A\|_{*}$ to denote the spectral, Frobenius, and nuclear norm of the matrix $\mb A$ , and $\mb A^*$ its (conjugate) transpose. For convenience, we let $\mb X_0 = (\mb L_0,\mb S_0)$ be the concatenation of $\mb L_0$ and $\mb S_0$. We assume that $\mb L_0$ is a low-rank matrix of rank $r$ whose compact SVD is $\mb L_0 = \mb U \mb \Sigma \mb V^*$, $\mb U\in \R^{n_1\times r}$, $\mb V\in \R^{n_2\times r}$, and without loss of generality, $n_1\geq n_2$. Let $T=\{\mb U\mb Q^*+\mb R\mb V^*| \mb Q\in \R^{n_2\times r},\mb R\in \R^{n_1\times r}\}$ denote the tangent space of rank $r$ matrices at $\mb L_0$. In addition, we assume that $\mb S_0$ is sparse with support in $\Omega$. 

Since it is impossible to disentangle $\mb L_0$ and $\mb S_0$ if the low-rank matrix $\mb L_0$ is sparse, or if the sparse matrix $\mb S_0$ is low-rank, we make the following two assumptions:
\begin{assumption}\label{assumption-incoherent} The low-rank matrix $\mb L_0$ satisfies the incoherence property with parameter $\nu$\footnote{$\nu\geq 1$ since $\|\mb U\|_F^2 = r$, and so $\max_i \|\mb U^* \mb e_i\|^2 \geq \frac{r}{n_1}$.}, i.e.
\begin{equation*}
    \max_i \|\mb U^* \mb e_i\|^2\leq \frac{\nu r}{n_1}, \quad \max_i \|\mb V^* \mb e_i\|^2\leq \frac{\nu r}{n_2}, \quad \|\mb U\mb V^*\|_{\infty}\leq \sqrt{\frac{\nu r}{n_1n_2}}.
\end{equation*}
\end{assumption}
\begin{assumption}
The support $\Omega$ is chosen uniformly among all sets of cardinality $m$, and the signs of supports are random, i.e. $P[(\mb S_0)_{i,j}>0|(i,j)\in \Omega] =P[(\mb S_0)_{i,j}<0|(i,j)\in \Omega] = 0.5 $.
\label{assumption-sparsity}
\end{assumption}
These assumptions follow \cite{candes2011robust}; indeed, our proof makes use of a dual certificate constructed for noiseless low-rank and sparse recovery in that paper.

\subsection{Problem Formulation and Main Results}
Inspired by square root Lasso, we propose to solve the robust noisy matrix recovery problem through the following optimization problem:
\begin{equation} \label{eqn:root-pcp}
\text{\rootpcp}:  \min_{\mb L,\mb S} \; \left\| \mb L \right\|_* + \lambda \| \mb S \|_1 + \mu \left\| \mb L + \mb S - \mb D \right\|_F.
\end{equation}
The parameter $\lambda$ that balances the low-rank and the sparse regularizers is studied in \cite{candes2011robust}, where it is shown that $\lambda = 1/\sqrt{n_1}$ gives exact recovery when $\mb Z_0 = \mb 0$ and the $\mu \left\| \mb L + \mb S - \mb D \right\|_F$ penalty term in \eqref{eqn:root-pcp} is replaced with the constraint $\mb L + \mb S = \mb D$. In this work, we build on this result and focus on the parameter $\mu$. Our main result is that under the aforementioned (standard) hypotheses on $\mb L_0$ and $\mb S_0$, using a single, universal choice $\mu = \sqrt{n_2 / 2}$, \rootpcp recovers $\mb L_0$ and $\mb S_0$, with an estimation error that is proportional to the norm of the noise:

\begin{theorem}\label{thm:main_thm}
Under Assumptions \ref{assumption-incoherent} and \ref{assumption-sparsity}, provided that $r,m$ satisfies
\begin{equation}\label{eqn:conditions-r-s-new}
    r\leq \rho_r n_{2}\nu^{-1}(\log n_{1})^{-2}, \quad m\leq \rho_s n_1n_2,
\end{equation}
where $\rho_r\leq 1 /10,\rho_s$ are some positive constants. Then there is a numerical constant $c$ such that with probability at least $1-cn_{1}^{-10}$, the \rootpcp problem \eqref{eqn:root-pcp} with $\lambda = 1/\sqrt{n_{1}}$ and $\mu = \sqrt{n_{2}/2}$ produces a solution $\wh{\mb X} = (\wh{\mb L},\wh{\mb S})$ such that 
\begin{equation}
    \|\wh{\mb X}-\mb X_0\|_F\leq 560\sqrt{n_1n_2}\|\mb Z_0\|_F.
\end{equation}
\end{theorem}

Why is it possible to achieve accurate estimation with a single choice of $\mu$? We draw intuition from a connection to the {\em stable principal component pursuit} formulations studied in \cite{zhou2010stable}. This work studies both constrained and unconstrained formulations: 
\begin{equation} \label{eqn:spcp}
\text{\stableC}:  \min_{\mb L, \mb S} \; \| \mb L \|_* + \lambda \| \mb S \|_1 ~ s.t.~\| \mb L + \mb S - \mb D \|_F\leq \delta.
\end{equation}
\begin{equation} \label{eqn:spcp-dual}
\text{\stableU}:  \min_{\mb L, \mb S} \; \| \mb L \|_* + \lambda \| \mb S \|_1 +\frac{\bar{\mu}}{2}\| \mb L + \mb S - \mb D \|_F^2.
\end{equation}
These formulations are equivalent, and equivalent to \rootpcp in the following sense: for each problem instance, there is a calibration of parameters $\delta \leftrightarrow \bar{\mu} \leftrightarrow \mu$ such that \rootpcp, \stableU\ and \stableC\ have exactly the same set of optimal solutions. However, \eqref{eqn:spcp}-\eqref{eqn:spcp-dual} require that the parameters $\delta$ and $\bar{\mu}$ be determined on an instance-by-instance basis, based on the noise level. Choosing these parameters correctly is essential: in \eqref{eqn:spcp}, $\delta$ should be chosen to be larger than $\| \mb Z_0 \|_F$. For square ($n_1 = n_2 = n$) matrices, in a stochastic setting in which $(\mb Z_0)_{ij}$ are iid $\mc N(0,\sigma^2)$, following \cite{zhou2010stable}, $\bar{\mu}$ can be chosen as $\tfrac{1}{2 \sigma \sqrt{n}}$. This follows from the fact that in this setting $n^{-1/2} \| \mb Z_0\| \to 2 \sigma$ almost surely; setting $\bar{\mu}$ in this fashion ensures that the singular value shrinkage induced by the nuclear norm regularizer $\| \cdot \|_*$ is greater than the largest singular value of $\mb Z_0$. 

In contrast to this $\sigma$-dependent penalty parameter, fixing $\mb S=\mb S_0$, \rootpcp formulation \eqref{eqn:root-pcp} requires that $\mb 0\in \partial (\|\wh{\mb L}\|_*+\mu \|\wh{\mb L}-\mb Z_0-\mb L_0\|_F)$, which translates into $-\mu\frac{\wh{\mb L}-\mb Z_0-\mb L_0}{\|\wh{\mb L}-\mb Z_0-\mb L_0\|_F} \in \partial \|\wh{\mb L}\|_*$. With the hope that $\wh{\mb L}\approx \mb L_0$, and by the subdifferential formulation\footnote{The subdifferential of a norm satisfies $\partial \|x\| = \{z \mid \left<x,z\right>=\|x\|,  \|z\|^*\le 1 \}$.}, we have $\mu\frac{\|\mb Z_0\|}{\|\mb Z_0\|_F}\approx 1$. The concentration stated above then gives an intuitive choice for $\mu \approx \frac{n\sigma}{2\sigma\sqrt{n}}$, or $ \mu= c_0\sqrt{n}$ for some $c_0>0$. The magic of \rootpcp is that by using the Frobenius norm instead of its square, the objective function becomes homogeneous, i.e. the gradient of the penalty term at the ground truth $\mb X_0$ becomes $\sigma$ independent, making a universal penalty parameter possible.\vspace{-.1in}

\subsection{Relationship to the Literature}

The problem of low-rank matrix recovery from gross sparse corruption can be considered a form of robust PCA,  \cite{candes2011robust,chandrasekaran2011rank}, and has been studied extensively in the literature. Algorithmic theory has been developed for both convex \cite{candes2011robust,chandrasekaran2011rank,hsu2010robust,li2013compressed,chen2013low}, and nonconvex optimization methods \cite{netrapalli2014non, chen2020bridging, gu2016low, ge2017no, you2020robust}. While many of the aforementioned works pertain to noiseless data, a line of work has studied extensions to noisy data. \cite{zhou2010stable} studied the problem of robust matrix recovery with bounded noise under the incoherence assumption, and proved a bound on the recovery error, with linear dependence on the noise level but suboptimal dependence on the matrix size. \cite{agarwal2012noisy} studied the problem with a weaker assumption about spikiness using decomposable regularizers and restricted strong convexity (RSC), and obtained essentially optimal bounds on the reconstruction error when the noise level is large. These weaker assumptions are not sufficient to ensure exact recovery, and so when the noise standard deviation $\sigma$ is small, this approach does not yield a reconstruction error proportional to the noise level. \cite{chen2015fast} formulated robust PCA as a semidefinite programming problem, which requires strong assumptions about square matrices and positive semidefiniteness of the low-rank matrix. Some other works \cite{klopp2017robust,wong2017matrix} further assumed partial observation of the matrix, and also derived tighter bounds on the recovery error. The recent work of \cite{chen2020bridging} achieves optimal error bounds for both large and small noise, using a novel analysis that leverages an auxiliary nonconvex program. Taken together, these results give efficient and provably effective methods, whose statistical performance is nearly optimal. Compared to e.g., \cite{chen2020bridging, agarwal2012noisy}, the stability guarantees provided by our theory are worse by a dimension-dependent factor. Nevertheless, all of the above works regarding robust PCA with noise, the optimization involves  parameters that must be set based on the noise level distribution, and therefore challenging in actual applications. 

On the other hand, there has been existing work in the literature on structured signal recovery without needing to know the noise level. Our proposed square root PCP is directly inspired by the square root Lasso \cite{belloni2011square}, which proposed the idea of replacing the squared loss with its ``square root'' version. This allows for a choice of the parameter independent of noise level, while maintaining near-oracle performance. Later works have extended this idea to other scenarios,
such as group lasso \cite{bunea2013group}, SLOPE variable selection \cite{derumigny2018improved}, elastic net \cite{raninen2017scaled} and matrix completion \cite{klopp2014noisy}, etc. Notably, the work of \cite{klopp2014noisy} studied the matrix completion where one aims to recover a low-rank matrix from noisy linear observations, aka matrix completion. Compared with that paper, this work aims to solve a different problem where the observation also contains a sparse outlier matrix. To the best of our knowledge, this paper is the first to propose a provable algorithm for Robust PCA with noisy observation that does not require knowledge of the noise level beforehand. On the algorithmic side, interior point method and first order method are used to solve the square root Lasso in \cite{belloni2011square}, while later works apply ADMM to the problem \cite{LassoADMM}\cite{LassoFlare}. In our problem, the objective function can be transformed into a separable form, making ADMM a reasonable choice. 

We note that for large $\sigma$ the error bound established in this paper is suboptimal compared with \cite{belloni2011square} and \cite{klopp2014noisy}. The problem of square root lasso enjoys benign properties (lower bounds on restricted eigenvalues) which we do not have in square root PCP. The paper \cite{klopp2014noisy} on matrix completion makes a spikiness assumption, and proves that a square root lasso-inspired formulation achieves essentially optimal estimation when the noise is large. As with robust matrix recovery, the spikiness assumption is not strong enough to imply exact recovery in the noiseless case. Compared to these works, the principal differences in this paper are (i) the problem formulation: we consider robust PCA with sparse errors, (ii) the analysis, which proceeds down different lines, and (iii) that our bounds are linear in the noise level, for both large and small noise. However, in contrast to \cite{klopp2014noisy}, our analysis does not yield minimax optimal estimation errors; it is worse by a dimension-dependent factor. Improving this dependence is an important direction for future work. \vspace{-.1in}

\section{Analysis}\label{section:analysis}

The proof of the main Theorem \ref{thm:main_thm} is different from the standard approach in \cite{belloni2011square} due to a lack of the Restricted Strong Convexity property for the map $(\mb L,\mb S)\to \|\mb L+\mb S-\mb D\|_F$. Instead, our approach has three key ingredients:
\begin{itemize}[noitemsep,nolistsep,leftmargin=3mm]
    \item The result from \stableC\  (Theorem \ref{thm:stable-pcp}) shows a recovery error $\|(\wh{\mb L},\wh{\mb S}) -(\mb L_0,\mb S_0)\|_F$ which depends linearly on the parameter $\delta$.

    
    \item The intimate connection between \rootpcp formulation \eqref{eqn:root-pcp} and \stableC \ formulation \eqref{eqn:spcp} can help translate the above solution property to \rootpcp (Lemma \ref{lemma:root_and_stable}).
    \item The powerful dual certificate construction proposed in \cite{candes2011robust} (restated in Lemma \ref{lemma:dual-existence}) can be used as an approximate subgradient to bound the regularizer at $\wh{\mb X}$.  
\end{itemize}
The proof of the main theorem has two steps. First, it uses the optimality condition and the subgradient to provide an upper and an lower bound for the regularization terms at $\wh{\mb X}$. Second, the result in Theorem \ref{thm:stable-pcp} is translated into the \rootpcp setting, and together with the bounds obtained above, we get the desired result. The proof is given in the supplementary material, and below we provide three ingredients.

First, we state the main theorem for \stableC \ problem:
\begin{theorem}[Theorem 2 in \cite{zhou2010stable}]\label{thm:stable-pcp}
Under Assumptions \ref{assumption-incoherent} and \ref{assumption-sparsity}, assuming further that $r\leq \rho_r' n_{2}\nu^{-1}(\log n_{1})^{-2}$ and $m\leq \rho_s' n_1n_2$ where $\rho_r',\rho_s'$ are some positive constants, there is a numerical constant $c'$ such that with probability at least $1-c'n_{1}^{-10}$, for any $\mb Z_0$ with $\|\mb Z_0\|_F\leq \delta$, the solution $\wh{\mb X} = (\wh{\mb L},\wh{\mb S})$ to the \stableC\ problem \ref{eqn:spcp} with $\lambda = 1/\sqrt{n_{1}}$ satisfies
\begin{equation}
    \|\wh{\mb X} - \mb X_0\|_F\leq \sqrt{320n_1n_2+4}\cdot \delta.
\end{equation}
\end{theorem}

Note that choosing $\delta = \|\mb Z_0\|_F$ allows a reconstruction error that is $O(\sqrt{n_1n_2}\|\mb Z_0\|_F)$. In the case when $\mb Z_0 = \mb 0$, \stableC\ recovers the matrices exactly: $\wh{\mb X} = \mb X_0$. This is in agreement with the result in \cite{candes2011robust}. The next lemma connects the two formulations \rootpcp and \stableC\ and the proof is provided in the supplementary material: 
\begin{lemma}\label{lemma:root_and_stable}
Consider the \rootpcp problem parameterized by $\mu$ and denote the result as $\wh {\mb L}_{\mathrm{root}}(\mu), \wh {\mb S}_{\mathrm{root}}(\mu)$, 
as well as the \stableC\ formulation parameterized by $\delta$ and denote the result as $\wh {\mb L}_{\mathrm{stable}}(\delta), \wh {\mb S}_{\mathrm{stable}}(\delta)$. Define $\delta(\mu) = \| \mb D -  \wh {\mb L}_{\mathrm{root}}(\mu)-\wh {\mb S}_{\mathrm{root}}(\mu) \|_F$, then 
\begin{equation}
    \wh {\mb L}_{\mathrm{stable}}(\delta(\mu)), \wh {\mb S}_{\mathrm{stable}}(\delta(\mu)) = \wh {\mb L}_{\mathrm{root}}(\mu), \wh {\mb S}_{\mathrm{root}}(\mu).
\end{equation}
\end{lemma}

Lastly, we show an adapted dual certificate construction: 
\begin{lemma}[Adapted from \cite{candes2011robust}]
Under Assumptions \ref{assumption-incoherent} and \ref{assumption-sparsity}, assume that $r,m$ satisfies
\begin{equation}
    r\leq \rho_r n_{2}\nu^{-1}(\log n_{1})^{-2}, \quad m\leq \rho_s n_1n_2,
\end{equation}
where $\rho_r\leq 1 /10,\rho_s$ are some positive constants. Then there is a numerical constant $c$ such that with probability at least $1-cn_{1}^{-10}$, there exists $\mb W, \mb F, \mb H$ such that 
\begin{equation}
    \mb U\mb V^*+\mb W = \lambda (\sign(\mb S_0)+\mb F+ P_{\Omega}\mb H),
\end{equation}
where $\mb W \in T^{\perp}$, $\|\mb W\|\leq \frac{1}{2}$, $P_{\Omega}\mb F=\mb 0$, $\|\mb F\|_{\infty}\leq \frac{1}{2}$, and $\|P_{\Omega}\mb H\|_F\leq \frac{1}{260\sqrt{2}}$.
\label{lemma:dual-existence}
\end{lemma}

The dual construction in \cite{candes2011robust} satisfies $\|P_{\Omega}\mb H\|_F\leq \frac{1}{4}$. However, the proof for Lemma 2.8(b) indicates that $\|P_{\Omega}\mb H\|_F\leq \frac{\sqrt{r}}{n_1^2}\leq \frac{\sqrt{r/n_2}}{n_1^{1.5}}$ and we only need to make sure that $\frac{\sqrt{r/n_2}}{n_1^{1.5}}\leq \frac{1}{260\sqrt{2}}$. This is a very mild condition, especially when it comes to the high dimensional real data (such as video). If we require that $r\leq n_2/10$, then problems of reasonably large dimension suffice, say, $n_1\geq 120$. And in the extreme case, we can set $\rho_r \leq 1/(260\sqrt{2})^2$.

\section{Solving \rootpcp with ADMM} 
Different from \cite{zhou2010stable} where \stableU\ \eqref{eqn:spcp-dual} is solved via \textit{Accelerated Proximal Gradient} method, we solve \rootpcp (and \stableU) via ADMM-splitting since the objective is not differentiable. To avoid multi-block ADMM which is not guaranteed to converge \cite{admmConvergence}, we define variables $\mb X_1^* = (\mb L_1^*,\mb S_1^*,\mb Z^*), \ \mb X_2^* = (\mb L_2^*,\mb S_2^*)$, and reformulate problem \eqref{eqn:root-pcp} as: 
\begin{align}\label{eqn:admm_split_separable}
    \min_{\mb X_1,\mb X_2} \; &f(\mb X_1):= \left\| \mb L_1 \right\|_* + \lambda \| \mb S_1 \|_1 + \mu \left\| \mb Z \right\|_F \\
    \mathrm{s.t.} \quad &\mb X_1 + \begin{bmatrix}-\mb I &\mb 0\\\mb 0&-\mb I\\\mb I&\mb I \end{bmatrix}\mb X_2 = \begin{bmatrix}\mb 0\\ \mb 0\\ \mb D\end{bmatrix}. \nonumber
\end{align}

The problem \eqref{eqn:admm_split_separable} can be separated into 2 blocks nicely ($\mb X_1$ and $\mb X_2$), which guarantees convergence of ADMM (under additional mild conditions)\cite{boydADMM}. 

Define dual variables $\mb Y^* = (\mb Y_1^*,\mb Y_2^*,\mb Y_3^*)$, the Lagrangian can be written as
\begin{align*}
    \mc L_\rho(\mb X_1,\mb X_2,\mb Y) &= \left\| \mb L_1 \right\|_* + \lambda \| \mb S_1 \|_1 + \mu \left\| \mb Z \right\|_F+ \innerprod{\mb L_1-\mb L_2}{\mb Y_1} +\frac{\rho}{2}\|\mb L_1-\mb L_2\|_F^2+ \innerprod{\mb S_1-\mb S_2}{\mb Y_2} \nonumber\\
    &\quad+ \frac{\rho}{2} \|\mb S_1-\mb S_2\|_F^2+\innerprod{\mb L_2+\mb S_2+\mb Z-\mb D}{\mb Y_3}  + \frac{\rho}{2}\|\mb L_2+\mb S_2+\mb Z-\mb D\|_F^2.
\end{align*}
We present the update rules\footnote{Recall that $\prox_{\gamma \|\cdot\|_*} \paren{\mb Z} = \sum_i \max(\lambda_i-\gamma,0 )\mb u_i \mb v_i^*$, where $\mb Z = \sum_i \lambda_i \mb u_i\mb v_i^*$ is the SVD, $[\prox_{\gamma \|\cdot\|_1} \paren{\mb Z}]_{i,j} = \max(|\mb Z_{i,j}|-\gamma ,0)\cdot \sign(\mb Z_{i,j})$, and $\prox_{\gamma \|\cdot\|_F} \paren{\mb Z} = \max(\|\mb Z\|_F-\gamma ,0)\frac{\mb Z}{\|\mb Z\|_F}$.} as well as the stopping criteria adapted from \cite{boydADMM} in Algorithm \ref{alg:root-pcp} and \texttt{helper()} function in the supplementary material. The stopping criteria takes into account the primal and the dual feasibility conditions, and the algorithm stops when the tolerances set using an absolute and relative criterion are reached. 

If we modify the update of $\mb Z$ in Algorithm \ref{alg:root-pcp} to $
\mb Z \leftarrow \paren{\mb D - \mb L_2 - \mb S_2 - \frac{1}{\rho}\mb Y_3} / (1+\mu/\rho)$, we get ADMM for \stableU\ \eqref{eqn:spcp-dual}. \vspace{-1em}

\begin{algorithm*}[h!]
\begin{multicols}{2}
\begin{algorithmic}
\item \textbf{Input:} $\mb D\in \R^{n_1\times n_2}, \lambda,\mu$.
\item \textbf{Output:} $\mb L, \mb S\in \R^{n_1\times n_2}$.
\hrule
\vspace{.1in}
\LineComment{Tolerance levels, max iterations}
\State $\epsilon_{\mathrm{abs}} \leftarrow 10^{-6}, \epsilon_{\mathrm{rel}} \leftarrow 10^{-6}, N \leftarrow 5000$ 
\LineComment{Initialization}
\State $\bm L_1,\bm L_2,\bm S_1,\bm S_2,\bm Z,\bm Y_1,\bm Y_2,\bm Y_3  \leftarrow  \bm 0_{n_1\times n_2}$ 
\State $\rho \leftarrow 0.1$
\For{$i=1, i\le N, i++$}
	\LineComment{Save old values temporarily}
	\State $(\mb L'_2,\mb S'_2) \leftarrow (\mb L_2,\mb S_2)$ 
	\LineComment{ADMM updates}
	\State $\mb L_1 \leftarrow \prox_{\frac{1}{\rho}\|\cdot\|_*} \paren{\mb L_2 - \frac{1}{\rho} \mb Y_1}$ 
	\State $\mb S_1 \leftarrow \prox_{\frac{\lambda}{\rho}\|\cdot\|_1} \paren{\mb S_2 - \frac{1}{\rho} \mb Y_2}$
	\State $\mb Z \leftarrow \prox_{\frac{\mu}{\rho}\|\cdot\|_F} \paren{\mb D - \mb L_2 - \mb S_2 - \frac{1}{\rho}\mb Y_3}$
	\State $\mb L_2 \leftarrow \frac{ \paren{ \mb D - \mb Z + 2\mb L_1 - \mb S_1 + \frac{1}{\rho} \paren{2\mb Y_1 - \mb Y_2 - \mb Y_3} }}{3}$
	\State $\mb S_2 \leftarrow \frac{ \paren{ \mb D - \mb Z + 2\mb S_1 - \mb L_1 + \frac{1}{\rho} \paren{2\mb Y_2 - \mb Y_1 - \mb Y_3} }}{3}$
	\State $\mb Y_1 \leftarrow  \mb Y_1 + \rho (\mb L_1 - \mb L_2)$
	\State $\mb Y_2 \leftarrow  \mb Y_2 + \rho (\mb S_1 - \mb S_2)$
	\State $\mb Y_3 \leftarrow  \mb Y_3 + \rho (\mb L_2 + \mb S_2 + \mb Z - \mb D)$
	\LineComment{Update $\rho$ and check convergence}
	\State $\rho, \mathrm{ifConverge} \leftarrow \texttt{helper()}$
	\If{ $\mathrm{ifConverge}$} 
		\State \textbf{break}
	\EndIf
\EndFor 
\State $(\mb L,\mb S) \leftarrow ((\mb L_1+\mb L_2)/2,(\mb S_1+\mb S_2)/2)$
\item \textbf{return} $\mb L,\mb S$

\end{algorithmic}
\end{multicols}
\caption{Algorithm for \rootpcp} \label{alg:root-pcp}
\vspace{-1em}
\end{algorithm*}

\section{Experiments}\label{section:experiments}
To show the effectiveness of our new formulation, we test \rootpcp on simulated data as well as real-world video datasets. The experiments suggest that our error bound in Theorem \ref{thm:main_thm} has a correct dependency on the noise level of $\mb Z_0$, but loses a factor of $n$ (the dimension of the problem). In addition, the solutions produced by \rootpcp with our proposed noise-independent $\mu$ and \stableU\ with the noise-dependent $\mu$ often look very similar to each other. Moreover, experiments on real-world datasets with natural noise also show the denoising effect of \rootpcp, making \rootpcp a practical approach with good performance in this robust noisy low-rank matrix recovery setting.

Additional experiments of \rootpcp on simulated data with varying $\mu$ also suggest that $\sqrt{n_2/2}$ can provide performance (recovery error) close to the optimal $\mu$, justifying our proposed choice of $\mu = \sqrt{n_2/2}$.

\subsection{Simulations with Varying Noise Levels and Dimension}\label{sec:simulation_exp_comparison}
In this set of experiments, we are interested in how our error bound in Theorem \ref{thm:main_thm} compare with the actual reconstruction error. We simulate $(\mb L_0,\mb S_0,\mb Z_0)$ with varying noise levels of $\mb Z_0$ and problem dimension $n_1,n_2$. To simulate $\mb L_0\in \R^{n_1\times n_2}$ of rank $r$, we generate $\mb U\in \R^{n_1\times r},\mb V\in \R^{n_2\times r}$ as the unnormalized singular vectors such that $\mb U,\mb V$ are entrywise i.i.d. $\mc N(0,1/n_1)$ and $\mc N(0,1/n_2)$ respectively and let $\mb L_0 = \mb U\mb V^*$. For $\mb S_0$, we let $P[(i,j)\in \Omega]=\rho_S$ and for $(i,j)$ in support $\Omega$, $(\mb S_0)_{(i,j)} \in \{0.05,-0.05\}$ with equal probability. For the noise $\mb Z_0$, we generate it as entrywise i.i.d. $\mc N(0,\sigma^2)$. 

In addition, in the experiments we take $n_1=n_2=n$, so we choose $\lambda = 1/\sqrt{n}$, $\mu_{\mathrm{stable}} = 1/(2\sigma\sqrt{n})$ (the noise level $\sigma$ is known), and $\mu_{\mathrm{root}} = \sqrt{n/2}$. Theoretical analysis in Theorem \ref{thm:main_thm} and \ref{thm:stable-pcp} shows that with these parameters, $\|\wh{\mb X}-\mb X_0\|_F=O(n\|\mb Z_0\|_F)$.

\begin{figure}[h!]
\centering
\vspace{-1em}
\begin{subfigure}[RMSE, vary $\sigma$]{
  \includegraphics[width = 0.19\textwidth]{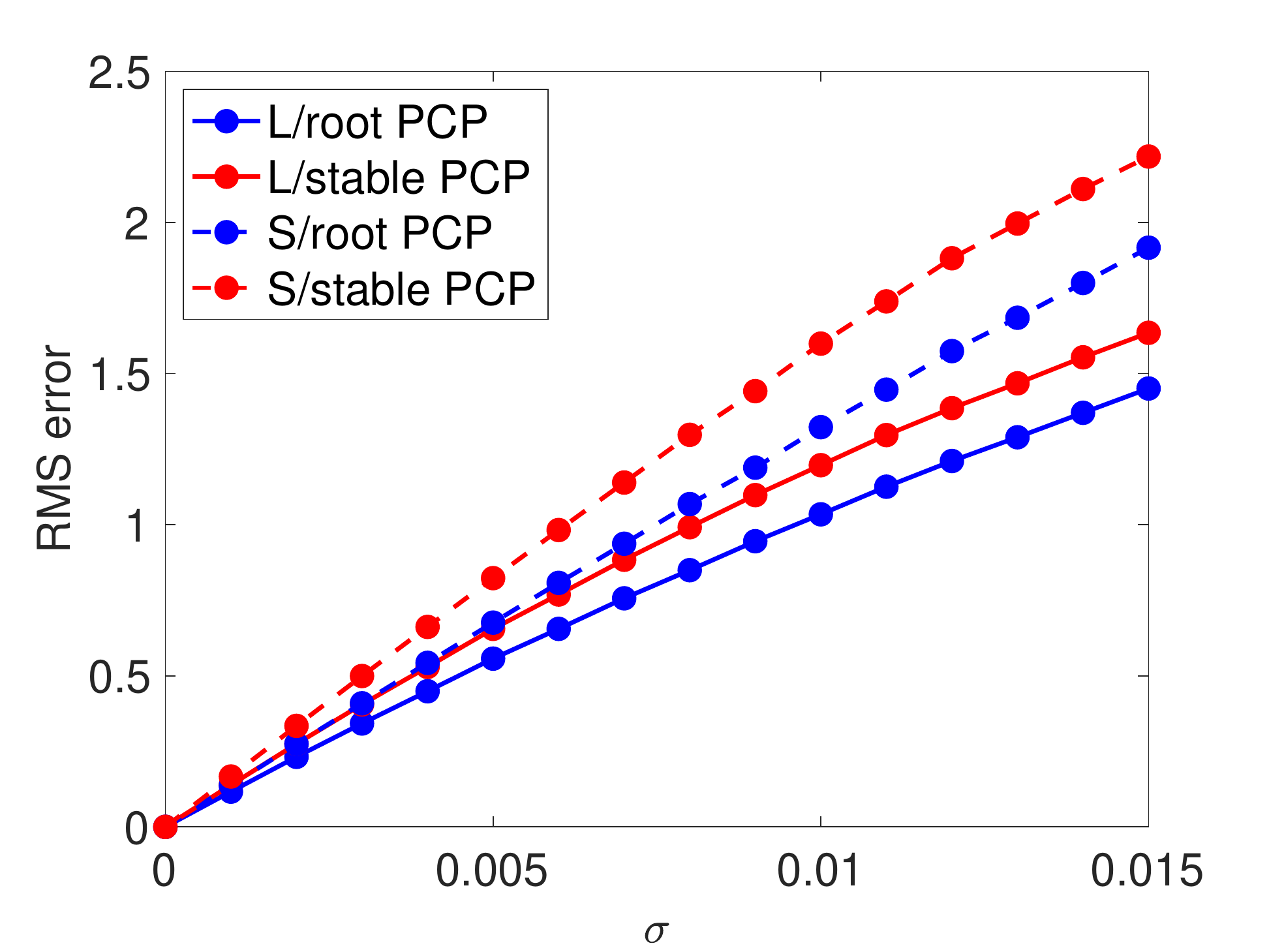}
\label{fig:rms_sigma}}
\end{subfigure}
\hspace{-0.9em}
\begin{subfigure}[RMSE, vary $n$]{
  \includegraphics[width = 0.19\textwidth]{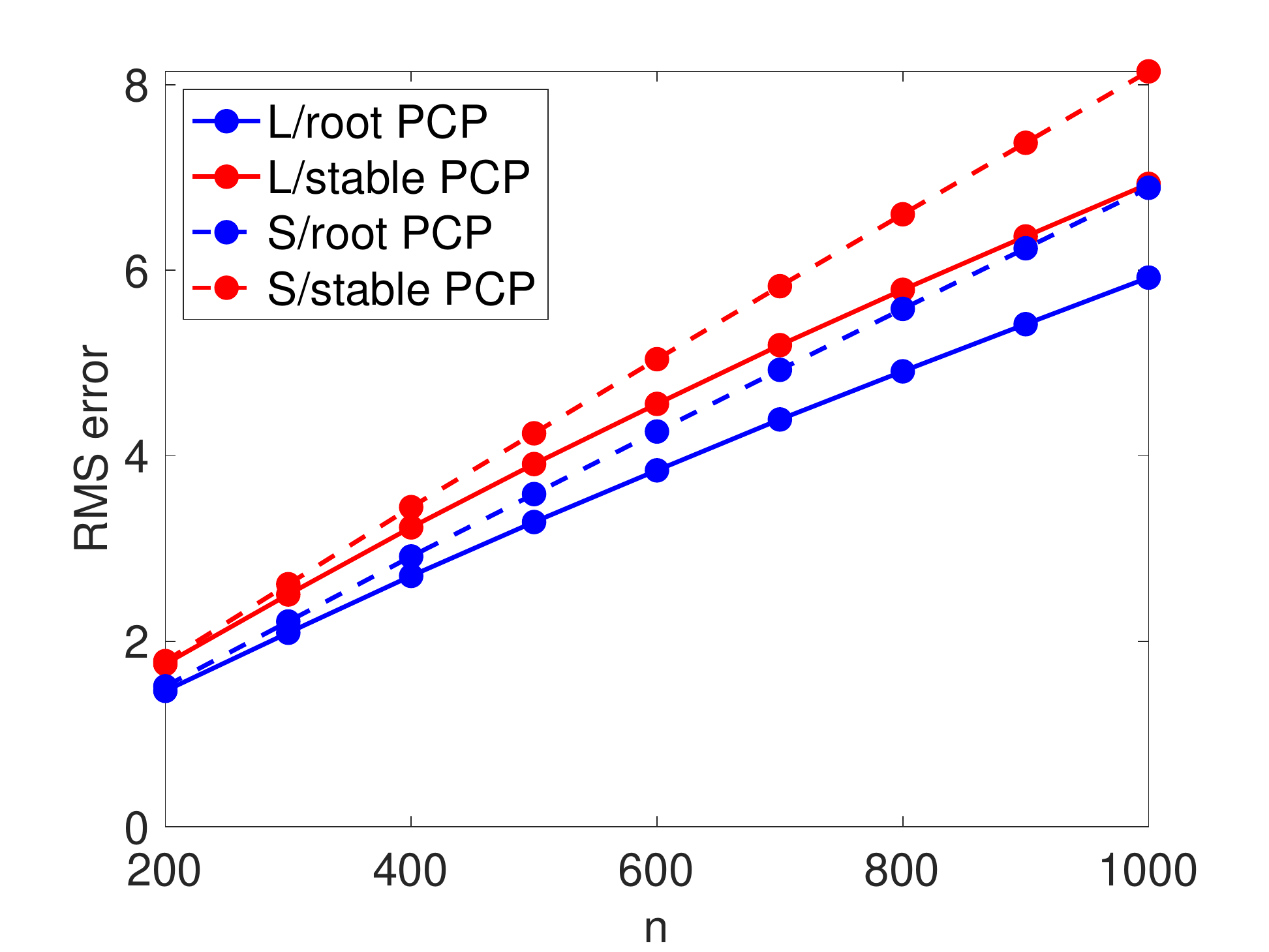}
\label{fig:rms_n}}
\end{subfigure}
\hspace{-0.9em}
\begin{subfigure}[``hall'' + noise]{
  \includegraphics[width = 0.19\textwidth]{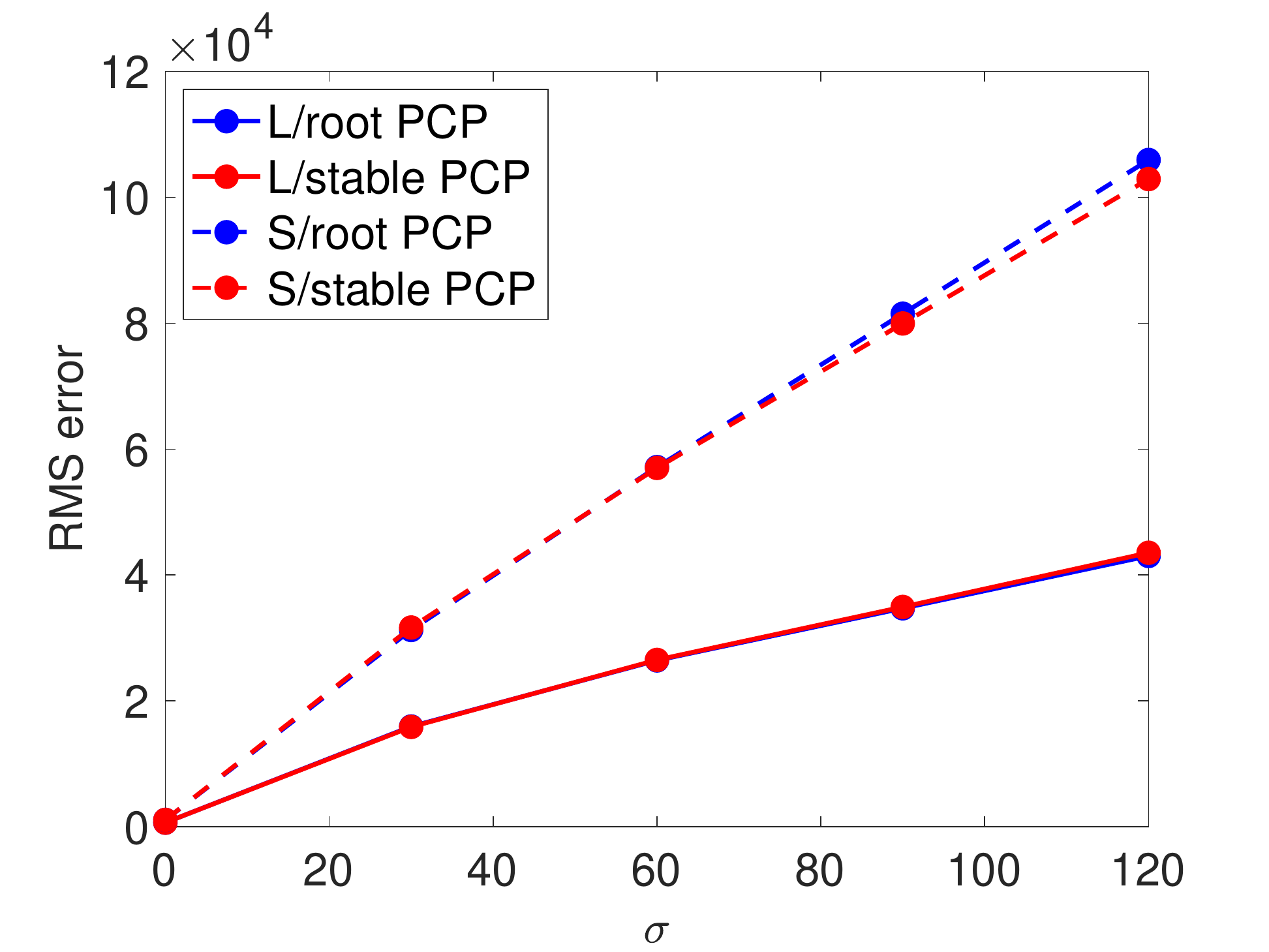}
\label{fig:error_video}}
\end{subfigure}
\vspace{-1em}
\caption{\stableU\ vs \rootpcp: a,b): simulated, c): hall}
\vspace{-1em}
\end{figure}

To test the dependency of the error on $\sigma$, we take $n=200$, $r = 10$, and so $\|\mb L_0\|_F^2 \approx r=10$. For the outlier $\mb S_0$, we take $\rho_S = 0.1$, so $\|\mb S_0\|_F^2 \approx 0.05^2n^2\rho_S=10 $. For the noise $\mb Z_0$, we take $\sigma\in \{0,0.001,\ldots,0.015\}$\footnote{When $\sigma = 0$ and $\mu_{\mathrm{stable}} = +\infty$, \stableU\ is equivalent to \stableC\ with $\delta = 0$.}, so $\|\mb Z_0\|_F^2\approx \sigma^2n^2 \in [0,9]$. For each $\sigma$ in the given set, we randomly generate 20 ground truth $(\mb L_0,\mb S_0,\mb Z_0)$ triplets and run \rootpcp and \stableU\ on them. We use the root-mean-squared (RMS) error defined as $(\frac{1}{20}\sum_{k=1}^{20}\|\wh{\mb L}^{(k)}-\mb L_0\|_F^2)^{1/2}$ and $(\frac{1}{20}\sum_{k=1}^{20}\|\wh{\mb S}^{(k)}-\mb S_0\|_F^2)^{1/2}$ for evaluation. In Figure \ref{fig:rms_sigma} we show the RMS error over 20 trials for the low-rank and the sparse. It is clear from the plot that $\|\wh{\mb L}-\mb L_0\|_F$ and $\|\wh{\mb S}-\mb S_0\|_F$ are $O(\sigma)$ for both \rootpcp and \stableU, which confirms that the reconstruction error is linear in the noise level $\sigma$.

We also notice that the recovery error in Figures \ref{fig:rms_sigma} and \ref{fig:error_video} is linear in the noise level for small $\sigma$, but exhibits a sublinear behavior for larger $\sigma$. This behavior reflects a general phenomenon in recovery/denoising using structured models (sparse, low-rank, etc.): the minimax noise sensitivity $\eta = \sup_{\sigma>0} \frac{1}{\sigma}E[\|\wh{\mb x}-\mb x_0\|]$ is obtained as $\sigma\to 0$. This means that for small $\sigma$, we expect a linear trend with slope $\eta$, while for larger $\sigma$, the dependence can be sublinear. This behavior has a general geometric explanation. For simplicity we sketch how this plays out in a simpler norm denoising problem, in which the target is to recover a structured signal $\mb x_0$, and we observe $\mb y = \mb x_0 +\sigma \mb z$. For simplicity, assume that we know that $\|\mb x_0\|_1\leq \tau$, and solve $\min_{\|\mb x\|_1\leq \tau} \|\mb x-\mb y\|_2$. For small $\sigma$, the estimation error $\wh{\mb x}-\mb x_0$ is simply the projection of the noise $\sigma \mb z$ onto the descent cone of the norm ball $\{\|\mb x\|_1\leq \tau\}$ at $\mb x_0$; its size is linear in $\sigma$. For larger $\sigma$, there is additional denoising due to the fact that the L1 ball is smaller than the descent cone at $\mb x_0$ — this leads to the behavior observed here.

To test the dependency of the error on the problem dimension,  we vary $n\in \{200,300,\ldots,1000\}$ and take $r = 0.1n$. We keep the setting for $\mb S_0$, and take $\sigma = 0.01$ as the noise level for $\mb Z_0$. Figure \ref{fig:rms_n} shows the RMS error. Note that for fixed $\sigma$, $\|\mb Z_0\|_F \sim n\sigma$, so the results in Theorem \ref{thm:main_thm} and Theorem \ref{thm:stable-pcp} bound the reconstruction error as $O(n^2)$. However, the analysis provides only a loose error bound. As can be seen from this set of experiment, the error is closer to $O(n)$. We provide experiments with different distributions of the noise in the appendix. 

\subsection{Real Data with Added Noise: Surveillance Video}\label{section:surveillance_video}
Many imaging datasets can be modeled as the sum of a low-rank matrix $\mb L_0$, a sparse outlier $\mb S_0$, and noise $\mb Z_0$. For instance, video data often consists of an almost fixed background which can be seen as low-rank, and a foreground (such as people) that only occupies a small fraction of the image pixels for a short amount of time, which can be considered as sparse. Thus, videos can naturally fit into our robust PCA framework. 

In this set of experiments, we use \rootpcp and \stableU\ to separate the background and the foreground for surveillance video data. We assume that the original video is noiseless, and manually add noise $\mb Z_0$ that is entrywise i.i.d.\ $\mc N(0,\sigma^2)$ to test the dependency of the reconstruction error on the noise level $\sigma$. 

We use the ``hall dataset'' in \cite{videoData}, a $200$-frame video of a hall that has people walking around. Each frame has resolution $144\times 176$, and is flattened as one column of the noiseless observation matrix $\mb D$, so we have $n_1 = 144\times 176$ and $n_2 = 200$. Each pixel is represented by a number in $[0,255]$, and the mean value among all pixels is $150.3295$, with standard deviation $45.7438$, and median $155.0000$.

For the added noise, we choose $\sigma\in \{0,30,60,90,120\}$, and denote the recovered matrices as $\wh{\mb X}_{\mathrm{root/stable}}^{(\sigma)}$. In addition, we let $\mb X_0 = \frac{1}{2}(\wh{\mb X}_{\mathrm{root}}^{(0)}+\wh{\mb X}_{\mathrm{stable}}^{(0)})$ be the ground truth, and evaluate the error using $\|\wh{\mb L}_{\mathrm{root/stable}}^{(\sigma)}-\mb L_0\|_F$ and $\|\wh{\mb S}_{\mathrm{root/stable}}^{(\sigma)}-\mb S_0\|_F$. We take $\lambda = 1/\sqrt{n_1}$, $\mu_{\mathrm{root}} = \sqrt{n_2/2}$ and $\mu_{\mathrm{stable}} =\frac{1}{\sigma(\sqrt{n_1}+\sqrt{n_2})} $ following the same intuition as in Section \ref{sec:introduction}.\footnote{Recall that $E[\|\mb Z_0\|]\leq \sigma(\sqrt{n_1}+\sqrt{n_2})$ for rectangular matrices, e.g. from \cite{vershynin2012matrices}} We run the experiments on a laptop with 2.3 GHz Dual-Core Intel Core i5, and set the maximal iteration of ADMM to be 5000. All of these experiments on real datasets end within 1 hour. For full details, please see the supplementary material. 

\begin{figure}[h!]
\centering
\vspace{-1em}
\begin{subfigure}[video]{
  \includegraphics[width = 0.14\textwidth]{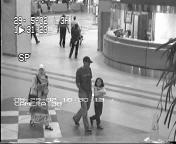}
\label{fig:hall_original_s1f1}}
\end{subfigure}
\hspace{-0.9em}
\begin{subfigure}[$\wh{\mb L}_{\mathrm{root}}^{(0)}$]{
  \includegraphics[width = 0.14\textwidth]{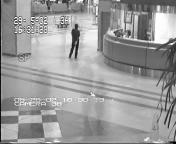}
\label{fig:hall_Lroot_s1f1}}
\end{subfigure}
\hspace{-0.9em}
\begin{subfigure}[$\wh{\mb S}_{\mathrm{root}}^{(0)}$]{
  \includegraphics[width = 0.14\textwidth]{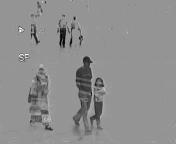}
\label{fig:hall_Sroot_s1f1}}
\end{subfigure}
\hspace{-0.9em}
\begin{subfigure}[ video$+\mb Z_0$]{
  \includegraphics[width = 0.14\textwidth]{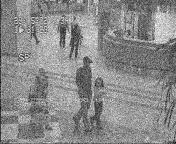}
\label{fig:hall_original_s2f1}}
\end{subfigure}
\hspace{-0.9em}
\begin{subfigure}[$\wh{\mb L}_{\mathrm{root}}^{(30)}$]{
  \includegraphics[width = 0.14\textwidth]{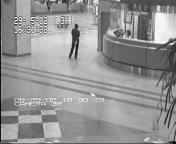}
\label{fig:hall_Lroot_s2f1}}
\end{subfigure}
\hspace{-0.9em}
\begin{subfigure}[$\wh{\mb S}_{\mathrm{root}}^{(30)}$]{
  \includegraphics[width = 0.14\textwidth]{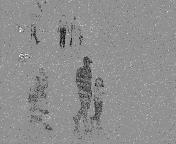}
\label{fig:hall_Sroot_s2f1}}
\end{subfigure}
\vspace{-1em}
\caption{hall frame 1: \rootpcp for $\sigma = 0,30$}
\vspace{-1em}
\label{fig:hall_frame1}
\end{figure}

In Figure \ref{fig:error_video}, we show the reconstruction error with varying noise levels. It can be seen that the error is indeed linear in $\sigma$, as predicted by our analysis. In Figures \ref{fig:hall_frame1}, we present the first frame (i.e. the first column) of the original video (with noise $\sigma = 0,30$), and the \rootpcp recovered low-rank and sparse matrices. Although the added noise blurs the videos, our \rootpcp is still stable and successfully decompose the background and the foreground.

\subsection{Real Data with Natural Noise: Low Light Video}\label{section:low_light_video}
Low light videos are known to have very large observation noise due to limited photon counts. In this experiment, we apply our \rootpcp to the Dark Raw Video (DRV) dataset in \cite{seeingMotion} (under MIT License) for foreground background separation and denoising. This dataset of RGB videos, approximately 110 frames each, $3672\times5496$ in resolution, was collected at low light settings, so the signal-to-noise ratio (SNR) is extremely low (negative if measured in dB)\cite{seeingMotion}. 

For the experiments, we choose video M0001 (basketball player), M0004 (toy windmill), and M0009 (billiard table) from DRV. As preprocessing, we convert the RGB videos to grayscale using \texttt{rgb2gray()} in Matlab, and crop and downsample each frame to reduce data size. The final resolution is $322\times440$ for M0001, $294\times440$ for M0004, and $306\times 458$ for M0009. 

\begin{figure}[h!]
\centering
\vspace{-2em}
\begin{subfigure}{
\makebox[20pt]{\raisebox{20pt}{\rotatebox[origin=c]{90}{}}}%
\makebox[0.16\textwidth]{$\mb D$}%
\makebox[0.16\textwidth]{$\wh{\mb L}+\wh{\mb S}$}%
\makebox[0.16\textwidth]{$\wh{\mb L}$}%
\makebox[0.16\textwidth]{$\wh{\mb S}$}%
\makebox[0.16\textwidth]{$\wh{\mb Z}$}%
}
\end{subfigure}
\vspace{-0.5em}
\begin{subfigure}{
\makebox[20pt]{\raisebox{20pt}{\rotatebox[origin=c]{90}{M0001}}}%
\includegraphics[width = 0.15\textwidth]{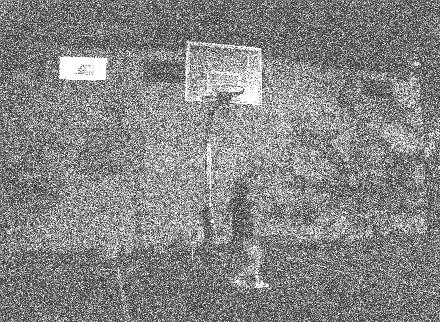}}
\end{subfigure}
\hspace{-0.5em}
\begin{subfigure}{
  \includegraphics[width = 0.15\textwidth]{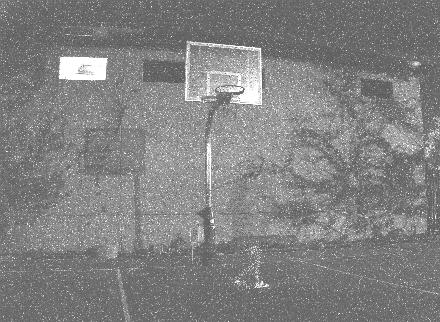}}
\end{subfigure}
\hspace{-0.5em}
\begin{subfigure}{
  \includegraphics[width = 0.15\textwidth]{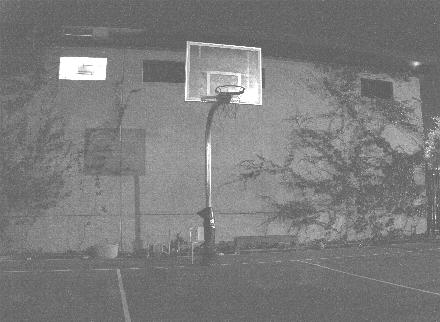}}
\end{subfigure}
\hspace{-0.5em}
\begin{subfigure}{
  \includegraphics[width = 0.15\textwidth]{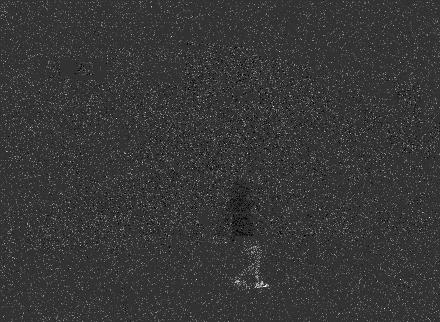}}
\end{subfigure}
\hspace{-0.5em}
\begin{subfigure}{
  \includegraphics[width = 0.15\textwidth]{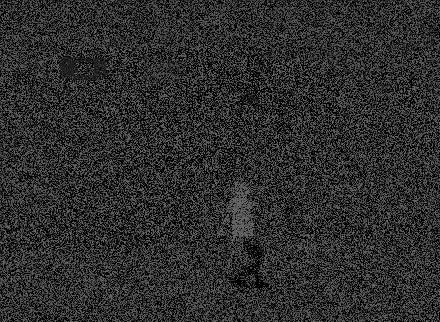}}
\end{subfigure}
\vspace{-0.5em}
\begin{subfigure}{
\makebox[20pt]{\raisebox{20pt}{\rotatebox[origin=c]{90}{M0004}}}%
  \includegraphics[width = 0.15\textwidth]{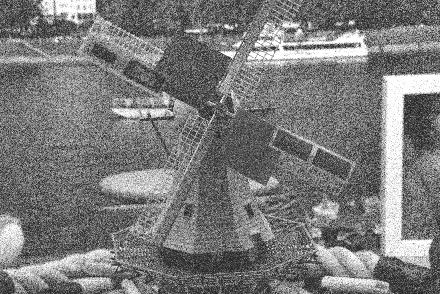}}
\end{subfigure}
\hspace{-0.5em}
\begin{subfigure}{
  \includegraphics[width = 0.15\textwidth]{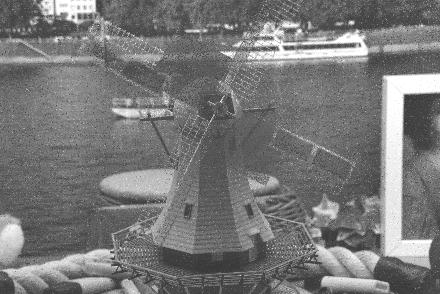}}
\end{subfigure}
\hspace{-0.5em}
\begin{subfigure}{
  \includegraphics[width = 0.15\textwidth]{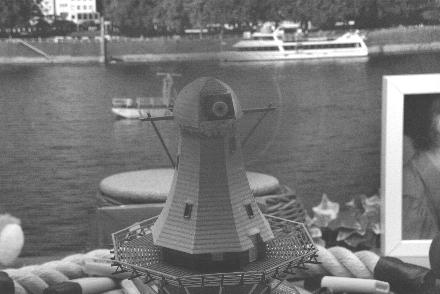}}
\end{subfigure}
\hspace{-0.5em}
\begin{subfigure}{
  \includegraphics[width = 0.15\textwidth]{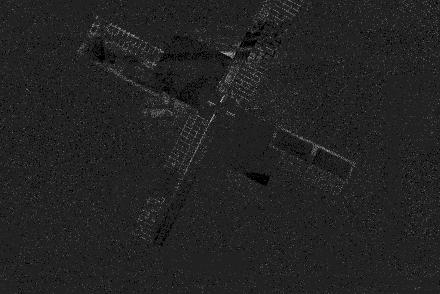}}
\end{subfigure}
\hspace{-0.5em}
\begin{subfigure}{
  \includegraphics[width = 0.15\textwidth]{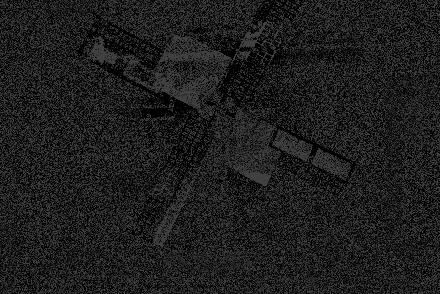}}
\end{subfigure}
\begin{subfigure}{
\hspace{.05cm}
\makebox[20pt]{\raisebox{20pt}{\rotatebox[origin=c]{90}{M0009}}}%
  \includegraphics[width = 0.15\textwidth]{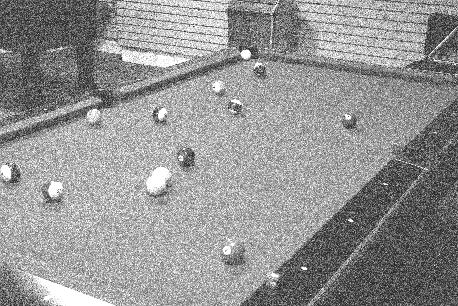}}
\end{subfigure}
\hspace{-0.5em}
\begin{subfigure}{
  \includegraphics[width = 0.15\textwidth]{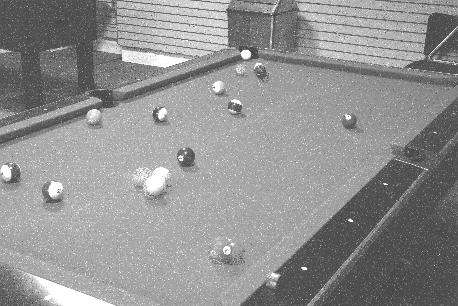}}
\end{subfigure}
\hspace{-0.5em}
\begin{subfigure}{
  \includegraphics[width = 0.15\textwidth]{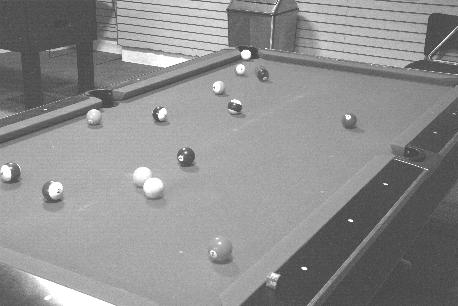}}
\end{subfigure}
\hspace{-0.5em}
\begin{subfigure}{
  \includegraphics[width = 0.15\textwidth]{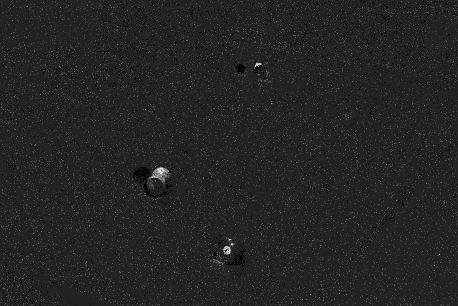}}
\end{subfigure}
\hspace{-0.5em}
\begin{subfigure}{
  \includegraphics[width = 0.15\textwidth]{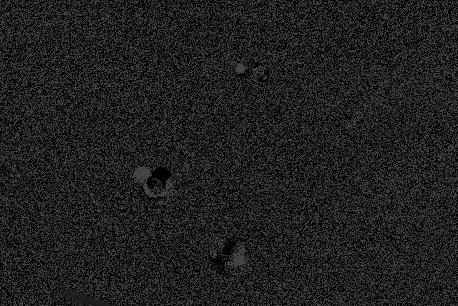}}
\end{subfigure}
\vspace{-1em}
\caption{Low light video frame 30 for M0001, M0004, and M0009 ($\wh{\mb Z} = \wh{\mb L}+\wh{\mb S}-\wh{\mb D}$). Image contrast is enhanced using \texttt{imadjustn()} in Matlab.}
\vspace{-1em}
\label{fig:denoising_video}
\end{figure}

We apply \rootpcp with $\lambda = 1/\sqrt{n_1}$, and $\mu = \sqrt{n_2/2}$ to these 3 videos, and present the results for frame 30 in Figure \ref{fig:denoising_video}. The denoising effect of \rootpcp can be seen by comparing $\mb D$ with $\wh{\mb L}+\wh{\mb S}$. In addition, $\wh {\mb L}$ recovers the background pretty well, $\wh {\mb S}$ captures the moving foreground but is still mixed with noise, which we believe is due to the extremely low SNR. 

\subsection{Real Data with Natural Noise: Optical Coherence Tomography}\label{section:oct}
In medical imaging, Optical Coherence Tomography can be used for micro-scale resolution, quick scanning of biological phenomenon \cite{octData}.  These scans of the same scene over time, called time-lapse B-scan, are often noisy, but fit into our low rank/sparse model. 

\begin{figure}[h!]
\centering
\vspace{-1em}
\begin{subfigure}[$\mb D$]{
  \includegraphics[width = 0.09\textwidth]{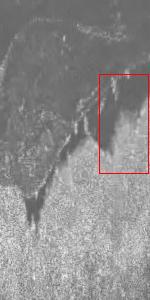}
  \label{fig:oct_raw50}}
\end{subfigure}
\hspace{-0.9em}
\begin{subfigure}[$\wh{\mb L}+\wh{\mb S}$]{
  \includegraphics[width = 0.09\textwidth]{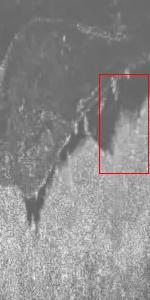}
\label{fig:oct_LS50}}
\end{subfigure}
\hspace{-0.9em}
\begin{subfigure}[$\wh{\mb L}$]{
  \includegraphics[width = 0.09\textwidth]{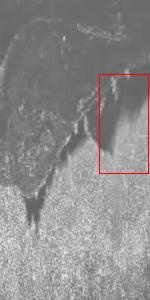}
\label{fig:oct_L50}}
\end{subfigure}
\hspace{-0.9em}
\begin{subfigure}[$\wh{\mb S}$]{
  \includegraphics[width = 0.09\textwidth]{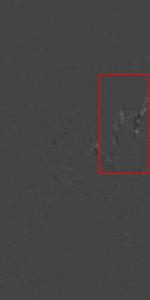}
\label{fig:oct_S50}}
\end{subfigure}
\hspace{-0.9em}
\begin{subfigure}[$\wh{\mb Z}$]{
  \includegraphics[width = 0.09\textwidth]{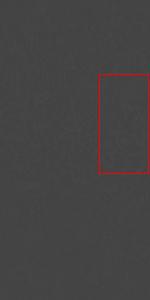}
\label{fig:oct_noise50}}
\end{subfigure}
\hspace{-0.9em}
\begin{subfigure}[$\mb D$]{
  \includegraphics[width = 0.09\textwidth]{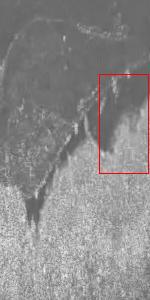}
  \label{fig:oct_raw100}}
\end{subfigure}
\hspace{-0.9em}
\begin{subfigure}[$\wh{\mb L}+\wh{\mb S}$]{
  \includegraphics[width = 0.09\textwidth]{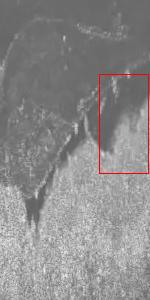}
\label{fig:oct_LS100}}
\end{subfigure}
\hspace{-0.9em}
\begin{subfigure}[$\wh{\mb L}$]{
  \includegraphics[width = 0.09\textwidth]{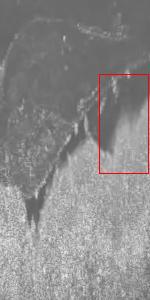}
\label{fig:oct_L100}}
\end{subfigure}
\hspace{-0.9em}
\begin{subfigure}[$\wh{\mb S}$]{
  \includegraphics[width = 0.09\textwidth]{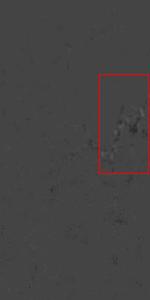}
\label{fig:oct_S100}}
\end{subfigure}
\hspace{-0.9em}
\begin{subfigure}[$\wh{\mb Z}$]{
  \includegraphics[width = 0.09\textwidth]{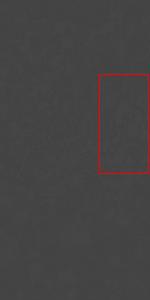}
\label{fig:oct_noise100}}
\end{subfigure}
\vspace{-1em}
\caption{OCT, a-e): frame 50, f-j): frame 100 ($\wh{\mb Z} = \wh{\mb L}+\wh{\mb S}-\wh{\mb D}$).}
\vspace{-1em}
\label{fig:oct}
\end{figure}

In this experiment, we apply \rootpcp to the time-lapse B-scans (250 frames of resolution $300 \times 150$) of human trachea samples containing motile cilia (demo dataset of \cite{octData} under CC0 License). We present the recovered frame 50 and 100 in Figure \ref{fig:oct}. As expected, $\wh{\mb L}$ captures the static background, and $\wh{\mb S}$ captures the motion of cilia. 

\subsection{Optimal Choice of $\mu$}\label{section:optimal_mu}
Our main result Theorem \ref{thm:main_thm} suggests a tuning-free $\mu = \sqrt{n_2/2}$. Here, we investigate experimentally if this choice of $\mu$ is optimal. We vary the problem dimensions $n_1$ and $n_2$, the rank-dimension ratio $\rho_L := r/n$ ($n=n_1=n_2$), and the noise standard deviation $\sigma$. For each choice of parameters $(n_1,n_2, \rho_L,\sigma)$, we generate 10 pairs of $(\mb L_0,\mb S_0,\mb Z_0)$ using the same method as in Section \ref{sec:simulation_exp_comparison}, run \rootpcp with $\lambda = 1/\sqrt{n_1}$ and $\mu = c\mu_0$, where $\mu_0 = \sqrt{n_2}$ and $c$ is a varying coefficient ($c = 1/\sqrt{2}\approx 0.71$ corresponds to our proposed value). Settings of all parameters are included in the supplementary material. 

\begin{figure}[h!]
\centering
\vspace{-1em}
\begin{subfigure}[vary $n=n_1=n_2$]{
  \includegraphics[width = 0.22\textwidth]{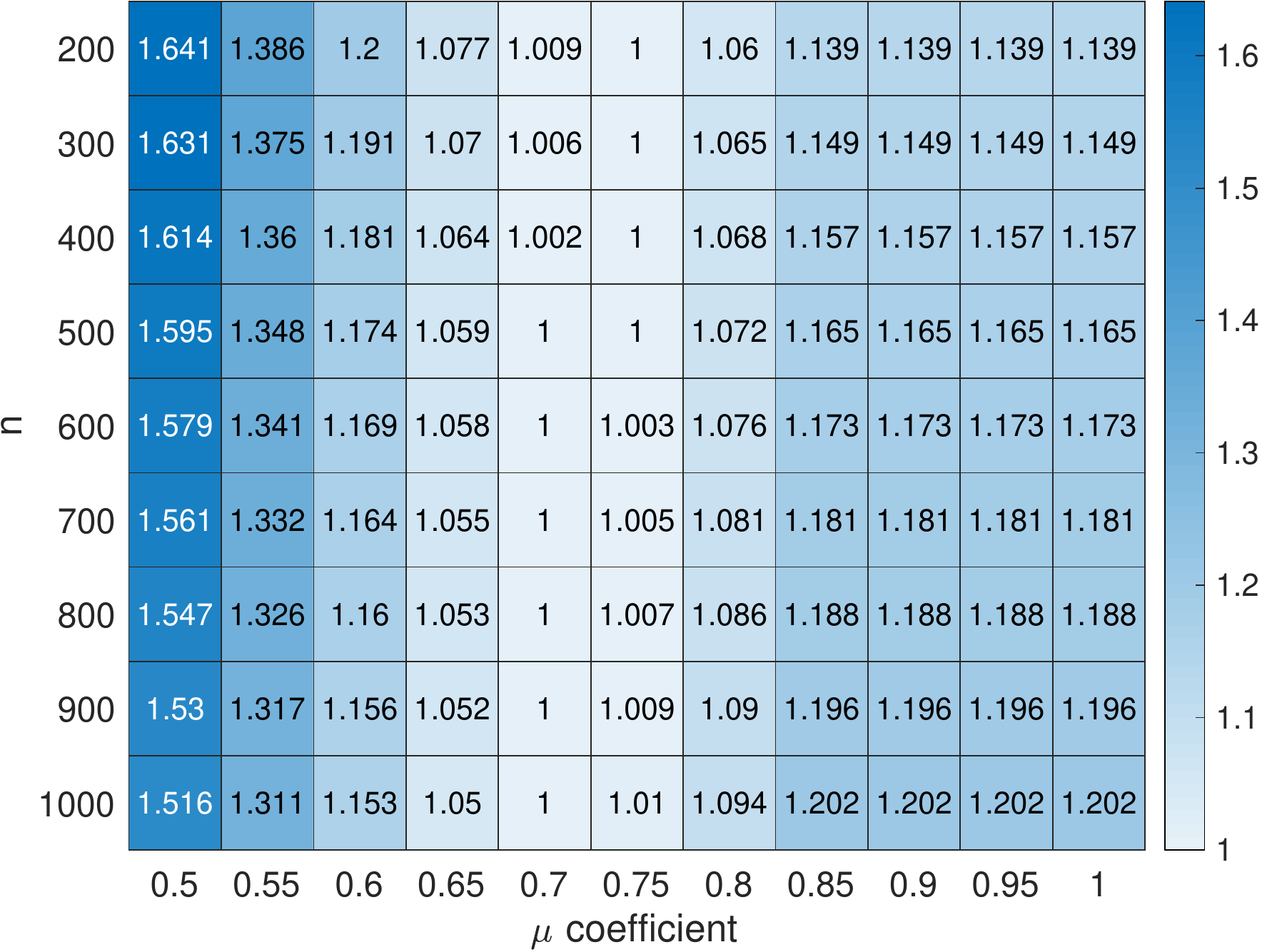}
\label{fig:vary_mu_n}}
\end{subfigure}
\begin{subfigure}[vary $n_1$]{
  \includegraphics[width = 0.22\textwidth]{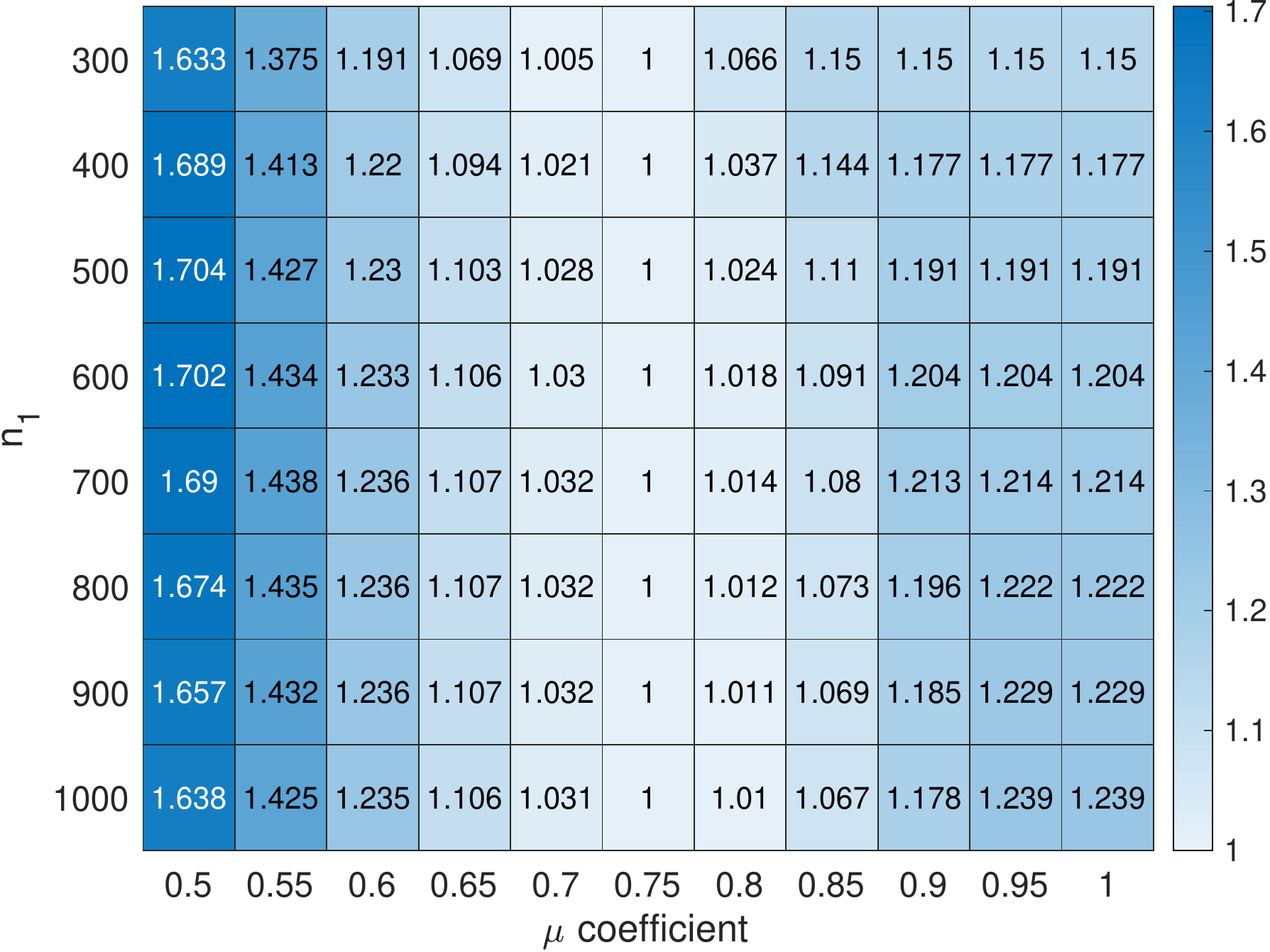}
\label{fig:vary_mu_p}}
\end{subfigure}
\begin{subfigure}[vary $\rho_L$]{
  \includegraphics[width = 0.22\textwidth]{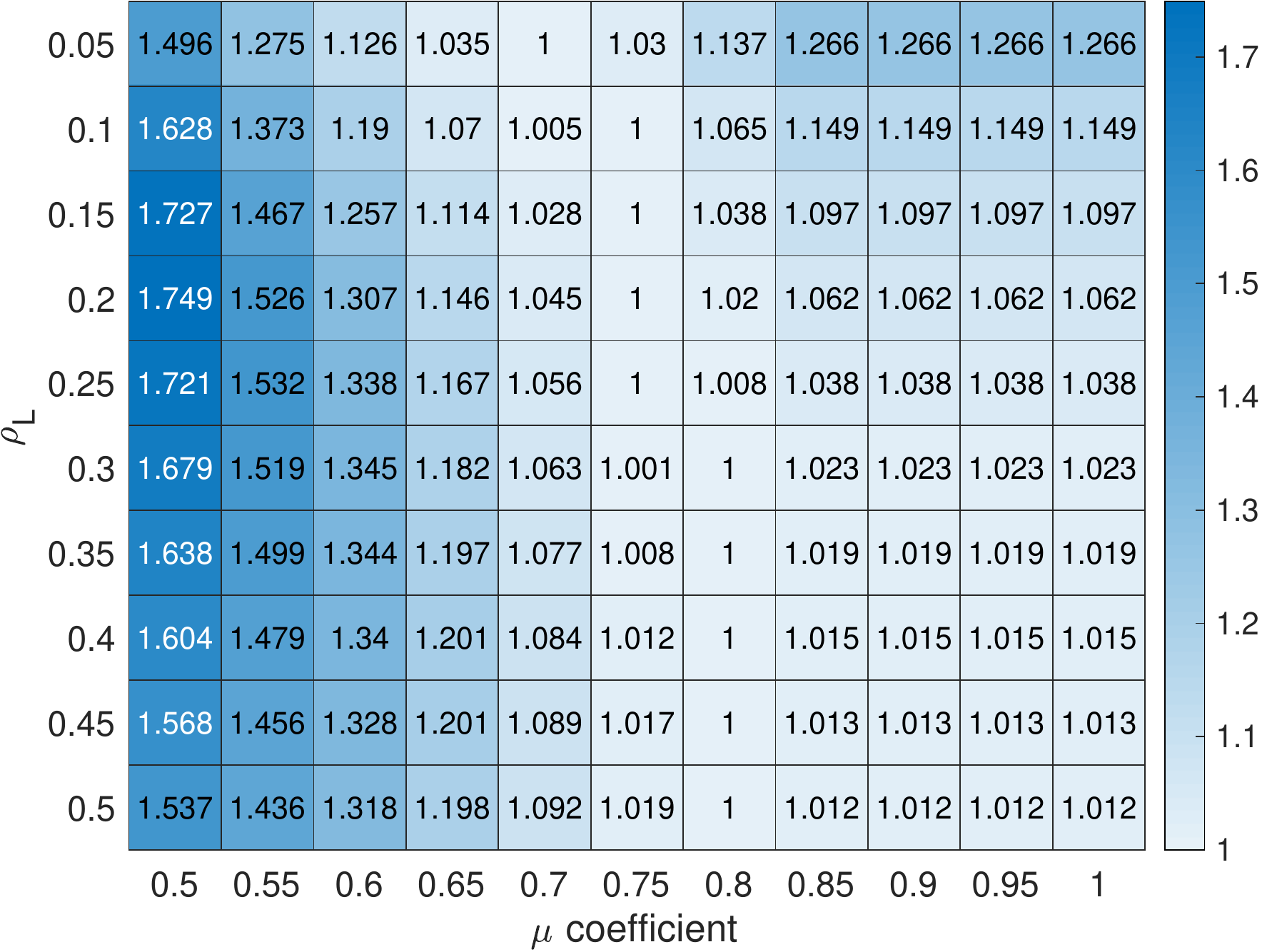}
\label{fig:vary_mu_rho}}
\end{subfigure}
\begin{subfigure}[vary $\sigma$]{
  \includegraphics[width = 0.22\textwidth]{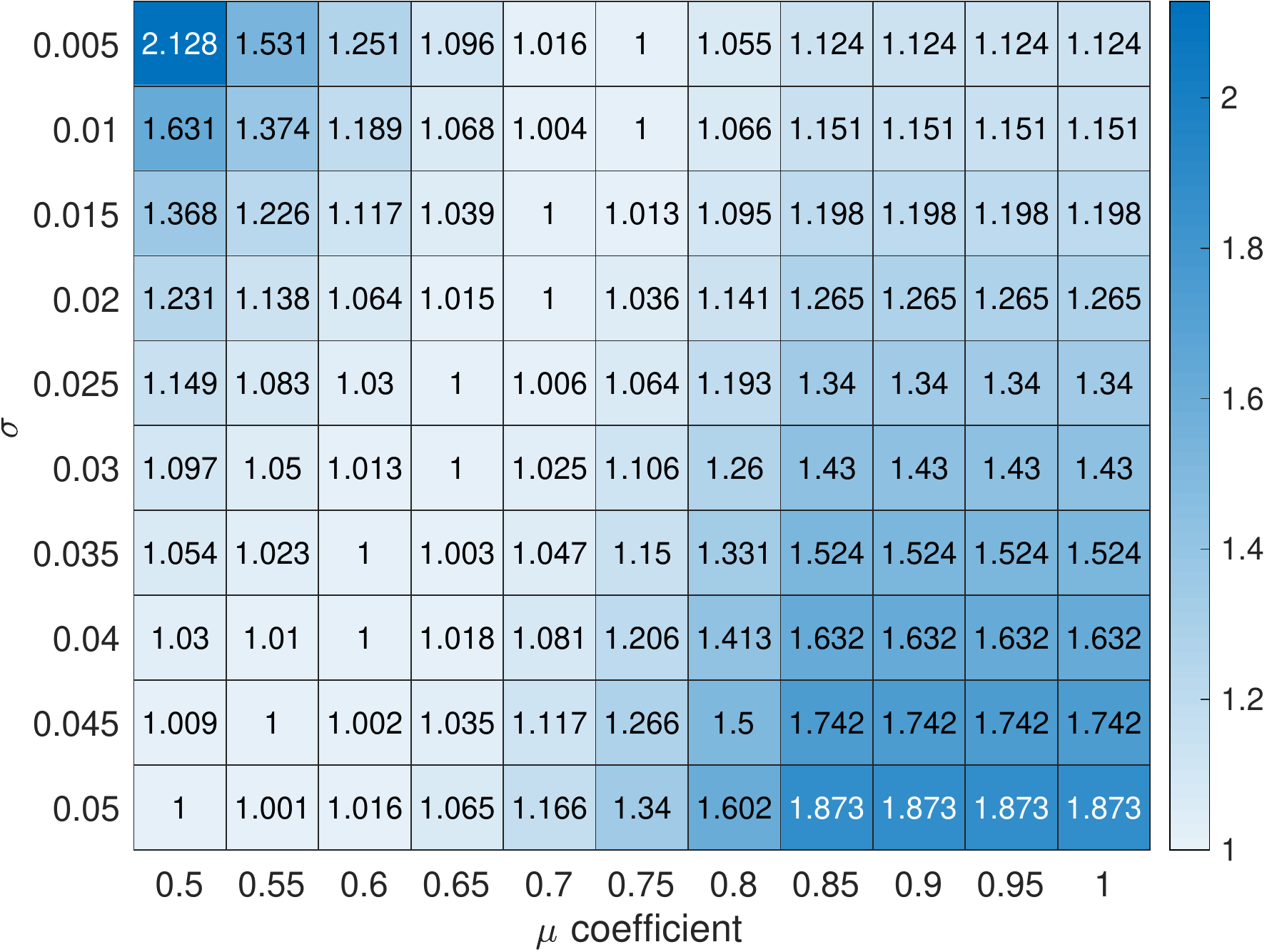}
\label{fig:vary_mu_sigma}}
\end{subfigure}
\vspace{-1em}

\caption{$\eta_{\mathrm{rel}} (\mu)$ under different varying parameters}
\vspace{-1em}
\label{fig:optimal_mu}
\end{figure}

We use the 10-average of $\|(\wh{\mb L},\wh{\mb S}) - (\mb L_0,\mb S_0)\|_F $ as the evaluation metric. In Figure \ref{fig:optimal_mu}, we show the heatmaps of this metric relative to the optimal $\mu = c\mu_0$ among all tested $c$, i.e. $\eta_{\mathrm{rel}} (\mu) = \frac{\| (\wh {\mb L}(\mu), \wh {\mb S}(\mu)) - (\mb L_0,\mb S_0)\|_F}{\min_{\mu'=c\mu_0}\| (\wh {\mb L}(\mu'), \wh {\mb S}(\mu')) - (\mb L_0,\mb S_0)\|_F}$, so the optimal $\mu$ has value $1$ in each row of the heatmaps. 
From Figure \ref{fig:optimal_mu}, we see that varying $n_1$, $n_2$ has little effect on the optimal choice of $\mu$, which is approximately between $ 0.7 \sqrt{n_2}$ and $ 0.75 \sqrt{n_2}$, close to our $\sqrt{n_2/2}$. However, decreasing $\rho_L$ or increasing $\sigma$ suggests a smaller value of optimal $\mu$. This makes sense because with higher level of noise or smaller rank (thus smaller norm $\|\mb L_0\|_F$), the SNR is smaller, so we should put smaller penalty on $\|\mb L+\mb S-\mb D\|_F$. Nevertheless, in all these settings, choosing $\mu = \sqrt{n_2/2}$ still gives satisfying results, as the recovery errors for $\mu = 0.7\sqrt{n_2}$ are close to the optimal performance: the error ratios are below $1.2$. From these results, we believe that while $\mu = \sqrt{n_2/2}$ may not be the optimal choice with respect to the recovery error, it can provide performance close to the optimal, and is therefore a very effective choice. 

\section{Conclusion}\label{section:conclusion}
In this work, we propose \rootpcp, a convex optimization approach for noisy robust low-rank matrix recovery. The benefit of our approach as compared to previous methods such as stable PCP is that it enables tuning-free recovery of low-rank matrices: theoretical analysis and simulations show that a single universal penalty parameter yields stable recovery at any noise standard deviation. Real video data experiments show suggest that many real life models fit into this low-rank plus sparse setting, and \rootpcp (as well as stable PCP) does a good job in denosing and recovering the patterns of interest. 

The presented experiments suggest the potential for both positive and negative societal impacts: visual surveillance can be abused, leading to significant negative impacts; at the same time, the denoising and foreground/background separation ability of \rootpcp can help improve the quality of noisy data in biomedical and scientific research (e.g. medical imaging), and people's life (e.g. low light video). 

\section*{Acknowledgement}

We would like to thank Marianthi-Anna Kioumourtzoglou, Jeff Goldsmith, Elizabeth Gibson, Rachel Tao and Lawrence Chillrud for many helpful discussions around matrix modeling of environmental data and the need for tuning-free solutions. This work was partially funded by the National Institute of Environmental Health Sciences (NIEHS) grant R01 ES028805. We also thank Christine Hendon for helpful pointers regarding optical coherence tomography data. Jingkai Yan also gratefully acknowledges support from the Wei Family Foundation. 

\bibliography{rootpcp}
\bibliographystyle{unsrt}

\newpage
\appendix

\section{Proof of Lemma \ref{lemma:root_and_stable} and Main Theorem \ref{thm:main_thm}}
\begin{proof}[Lemma \ref{lemma:root_and_stable}]
Clearly, $\wh {\mb L}_{\mathrm{root}}(\mu), \wh {\mb S}_{\mathrm{root}}(\mu)$ satisfies the constraint for the \stableC\ with $\delta = \delta(\mu)$. And by optimality of $\wh {\mb L}_{\mathrm{root}}(\mu), \wh {\mb S}_{\mathrm{root}}(\mu)$ for the \rootpcp problem, we get $\forall \mb L,\mb S~s.t. \| \mb D -\mb L - \mb S \|_F \leq \delta(\mu)$, 
\begin{equation}
\| \mb L \|_* + \lambda \|\mb S \|_1 \geq \|\wh {\mb L}_{\mathrm{root}}(\mu) \|_* + \lambda \| \wh {\mb S}_{\mathrm{root}}(\mu) \|_1.
\end{equation}
This shows the optimality of $\wh {\mb L}_{\mathrm{root}}(\mu), \wh {\mb S}_{\mathrm{root}}(\mu)$ for the \stableC\ problem with $\delta = \delta(\mu)$.
\end{proof}

\begin{lemma}[Adapted from Theorem 2 in \cite{zhou2010stable}]\label{lemma:stable_pcp_with_small_delta}
Under the same conditions as Theorem \ref{thm:stable-pcp}, with probability at least $1-c'n_{1}^{-10}$, for any $\mb Z_0$ with $\|\mb Z_0\|_F\geq \delta$, the solution $\wh{\mb X} = (\wh{\mb L},\wh{\mb S})$ to the \stableC\ problem \ref{eqn:spcp} with $\lambda = 1/\sqrt{n_{1}}$ satisfies
\begin{equation*}
    \|\wh{\mb X} - \mb X_0\|_F\leq \sqrt{720n_1n_2+2}\cdot \|\mb Z_0\|_F.
\end{equation*}
\end{lemma}
The proof of this lemma is essentially the same as the proof of Theorem \ref{thm:stable-pcp} in \cite{zhou2010stable}. The main differences are: we need to replace every occurrence of $\delta$ with $\|\mb Z_0\|_F$, and now since $(\wh{\mb L_0}, \wh{\mb S_0})$ is not a feasible solution, we need to change the bound of the objective to $\|\wh{\mb L}\|_* + \lambda \|\wh{\mb S}\|_1\leq \|\wh{\mb L}+\wh{\mb Z_0}\|_* + \lambda \|\wh{\mb S}\|_1\leq \|\wh{\mb L}\|_* +\sqrt{n_2}\|\mb Z_0\|_F+ \lambda \|\wh{\mb S}\|_1$.

\begin{proof}[Theorem \ref{thm:main_thm}]
Define $\wh {\mb \delta_L }=\wh{\mb L}- \mb L_0$ and $\wh{\mb \delta_S} = \wh{\mb S}-\mb S_0$. By the optimality of $\wh{\mb L},\wh{\mb S} $ and triangle inequality,
\begin{align}\label{eqn:lower_bound} 
    &\quad(\|\mb L_0\|_*+\lambda \|\mb S_0\|_1)- (\| \wh{\mb L}\|_*+\lambda \|\wh{\mb S}\|_1)\nonumber\\
    & \geq \mu(\|\mb D -\wh{\mb L}-\wh{\mb S} \|_F - \|\mb Z_0\|_F)\nonumber\\
    & \geq  \mu(\|\wh{\mb \delta_L} + \wh{\mb \delta_S} \|_F-2\|\mb Z_0\|_F).
\end{align}
Treating the dual certificate in Lemma \ref{lemma:dual-existence} as an approximate subgradient for the norm $\|\cdot\|_{*}$ and $\|\cdot\|_{1}$,
\begin{align}\label{eqn:upper_bound}
    &\quad(\| \wh{\mb L}\|_* - \|\mb L_0\|_*)+\lambda (\| \wh{\mb S}\|_1 - \|\mb S_0\|_1)\nonumber \\
    &\geq \langle \wh{\mb \delta_L}, \mb U\mb V^*+\mb W\rangle +\langle \wh{\mb \delta_S}, \lambda (\sign(\mb S_0)+\mb F)\rangle\nonumber\\
    &=\underbrace{\left\langle \wh{\mb \delta_L} + \wh{\mb \delta_S},\mb U\mb V^* +\mb W-\frac{\lambda P_{\Omega}\mb H}{2}\right\rangle}_{\Theta_1}+\underbrace{\lambda \left\langle \wh{\mb \delta_L} - \wh{\mb \delta_S},\frac{P_{\Omega}\mb H}{2}\right\rangle}_{\Theta_2}.
\end{align}
Next, we bound $\Theta_1,\Theta_2$ in \eqref{eqn:upper_bound}.
\begin{align*}\label{eqn:theta1}
    \Theta_1&\geq -\|\wh{\mb \delta_L} + \wh{\mb \delta_S} \|_F\cdot \|\mb U\mb V^* +\mb W-\lambda P_{\Omega}\mb H/2 \|_F\nonumber\\
    &\geq-\|\wh{\mb \delta_L} + \wh{\mb \delta_S} \|_F\cdot ( \sqrt{r+n_{2}/4} +\lambda\|P_{\Omega}\mb H\|_F/2)\nonumber\\
    &\geq-\|\wh{\mb \delta_L} + \wh{\mb \delta_S} \|_F\cdot(\sqrt{7/10}+1/520)\sqrt{n_{2}/2}\nonumber\\
    &\geq-0.85\mu \|\wh{\mb \delta_L} + \wh{\mb \delta_S} \|_F.
\end{align*}
where the second inequality follows from triangle inequality and properties of $\mb W$ from Lemma \ref{lemma:dual-existence}. The third inequality follows from the condition \eqref{eqn:conditions-r-s-new} $r\leq n_2/10$. For $\Theta_2$, by the dual construction,
\begin{equation*}\label{eqn:theta2}
    \Theta_2 \geq -(\lambda \|P_{\Omega}\mb H\|_F/2)\cdot (\|\wh{\mb \delta_L}  \|_F+\|\wh{\mb \delta_S} \|_F)\geq -(\|\wh{\mb \delta_L}  \|_F+\|\wh{\mb \delta_S} \|_F)/(520\sqrt{2n_1}).
\end{equation*}
Combining \eqref{eqn:lower_bound} and \eqref{eqn:upper_bound} with the bounds on $\Theta_1$ and $\Theta_2$,
\begin{equation}\label{eqn:bound_both_simple}
    \frac{1}{520\sqrt{n_1n_2}}(\|\wh{\mb \delta_L}  \|_F+\|\wh{\mb \delta_S} \|_F)+2\|\mb Z_0\|_F\geq 0.15\|\wh{\mb \delta_L} + \wh{\mb \delta_S} \|_F.
\end{equation}
From Lemma \ref{lemma:root_and_stable}, $(\wh{\mb L},\wh{\mb S})$ is also the solution to the \stableC\ problem parameterized by $\delta = \|\wh{\mb \delta_L} + \wh{\mb \delta_S}-\mb Z_0\|_F$. If $\delta\leq \|\mb Z_0\|_F$, Lemma \ref{lemma:stable_pcp_with_small_delta} gives the bound that 
\begin{equation*}
    \|(\wh{\mb \delta_L} ,\wh{\mb \delta_S})\|_F \leq 27\sqrt{n_1n_2}\|\mb Z_0\|_F.
\end{equation*}

If $\delta\geq \|\mb Z_0\|_F$, from Theorem \ref{thm:stable-pcp}, together with the trivial inequality that $\sqrt{320n_1n_2+4} \leq 26\sqrt{n_1n_2/2}$,
\begin{equation}\label{eqn:bound-sum}
    \|(\wh{\mb \delta_L} ,\wh{\mb \delta_S})\|_F \leq 26\sqrt{n_1n_2/2}\cdot\|\wh{\mb \delta_L} + \wh{\mb \delta_S}-\mb Z_0\|_F.
\end{equation}
Combining \eqref{eqn:bound_both_simple} and \eqref{eqn:bound-sum}, we get 
\begin{equation*}\label{eqn:bound-not-simplified}
    \|\wh{\mb \delta_L}  \|_F+\|\wh{\mb \delta_S} \|_F\leq26\sqrt{n_1n_2}(\|\wh{\mb \delta_L} + \wh{\mb \delta_S}\|_F+\|\mb Z_0\|_F)\leq\frac{1}{3}(\|\wh{\mb \delta_L}  \|_F+\|\wh{\mb \delta_S} \|_F)+ 373\sqrt{n_1n_2}\|\mb Z_0\|_F.
\end{equation*}
which proves the claim that
\begin{equation*}
    \|\wh{\mb X}-\mb X_0\|_F\leq \|\wh{\mb \delta_L}  \|_F+\|\wh{\mb \delta_S} \|_F\leq 560\sqrt{n_1n_2}\|\mb Z_0\|_F.
\end{equation*}
\end{proof}

\section{Stopping Criteria in Algorithm \ref{alg:root-pcp}}
The function \texttt{helper()} containing the stopping criteria and updates for $\rho$, adapted from \cite{boydADMM}, is presented in Algorithm \ref{alg:rho_convergence}. 

\begin{algorithm*}[h!]
\begin{algorithmic}
\item \textbf{Input:} $\mb D,\bm L_1,\bm L_2,\bm L_2',\bm S_1,\bm S_2, \bm S_2',\bm Z,\bm Y_1,\bm Y_2,\bm Y_3 \in \R^{n_1\times n_2}, \rho, \epsilon_{\mathrm{abs}}, \epsilon_{\mathrm{rel}}$.
\item \textbf{Output:} $\rho_+, \mathrm{ifConverge}$.
\hrule
\vspace{.1in}
    
    \LineComment{Calculate residuals}
	\State $r_\mathrm{primal} \leftarrow \|(\mb L_1-\mb L_2, \mb S_1-\mb S_2, \mb Z+\mb L_2+\mb S_2-\mb D)\|_F$ 
	\State $r_\mathrm{dual} \leftarrow \rho\cdot \|(\mb L_2-\mb L'_2, \mb S_2-\mb S'_2, \mb L_2+\mb S_2-\mb L'_2-\mb S'_2)\|_F$
	\LineComment{Calculate thresholds}
	\State $\theta_\mathrm{primal} \leftarrow  \epsilon_\mathrm{rel} \max\{ \|(\mb L_1, \mb S_1, \mb Z)\|_F,\|(\mb L_2, \mb S_2, \mb L_2+\mb S_2)\|_F,\|\mb D\|_F \}+\epsilon_\mathrm{abs} \sqrt{3n_1n_2}$ 
	\State $\theta_\mathrm{dual} \leftarrow  \epsilon_\mathrm{rel} \|(\mb Y_1,\mb Y_2,\mb Y_3)\|_F+\epsilon_\mathrm{abs} \sqrt{3n_1n_2}$
	\LineComment{Update $\rho$}
	\State $\rho_+\leftarrow \rho$
	\If {$r_\mathrm{primal}  > 10\cdot r_\mathrm{dual}$} 
		\State $\rho_+ \leftarrow 2\rho$
	\ElsIf{$r_\mathrm{dual}  > 10\cdot r_\mathrm{primal}$}
		\State $\rho_+ \leftarrow \rho/2$
	\EndIf
	\LineComment{Check convergence}
	\State $\mathrm{ifConverge} \leftarrow \mathrm{False}$
	\If{$r_\mathrm{primal} < \theta_\mathrm{primal}$ and $r_\mathrm{dual} < \theta_\mathrm{dual}$} 
		\State $\mathrm{ifConverge}\leftarrow \mathrm{True}$
	\EndIf
	
\item \textbf{return} $\rho_+,\mathrm{ifConverge}$
\end{algorithmic}
\caption{Function \texttt{helper()}: update $\rho$ and check convergence} \label{alg:rho_convergence}
\end{algorithm*}

\section{Experiments and Settings}
\subsection{Experiments with Different Distributions of Noise for Section \ref{sec:simulation_exp_comparison}}

\begin{figure}[h!]
\centering
\vspace{-1em}
\begin{subfigure}[Gaussian]{
  \includegraphics[width = 0.19\textwidth]{plots/simulation_varying_sigma/error_L_S-eps-converted-to.pdf}
\label{fig:rms_sigma_Gaussian}}
\end{subfigure}
\hspace{-0.9em}
\begin{subfigure}[Poisson $\lambda_P = 1$]{
  \includegraphics[width = 0.19\textwidth]{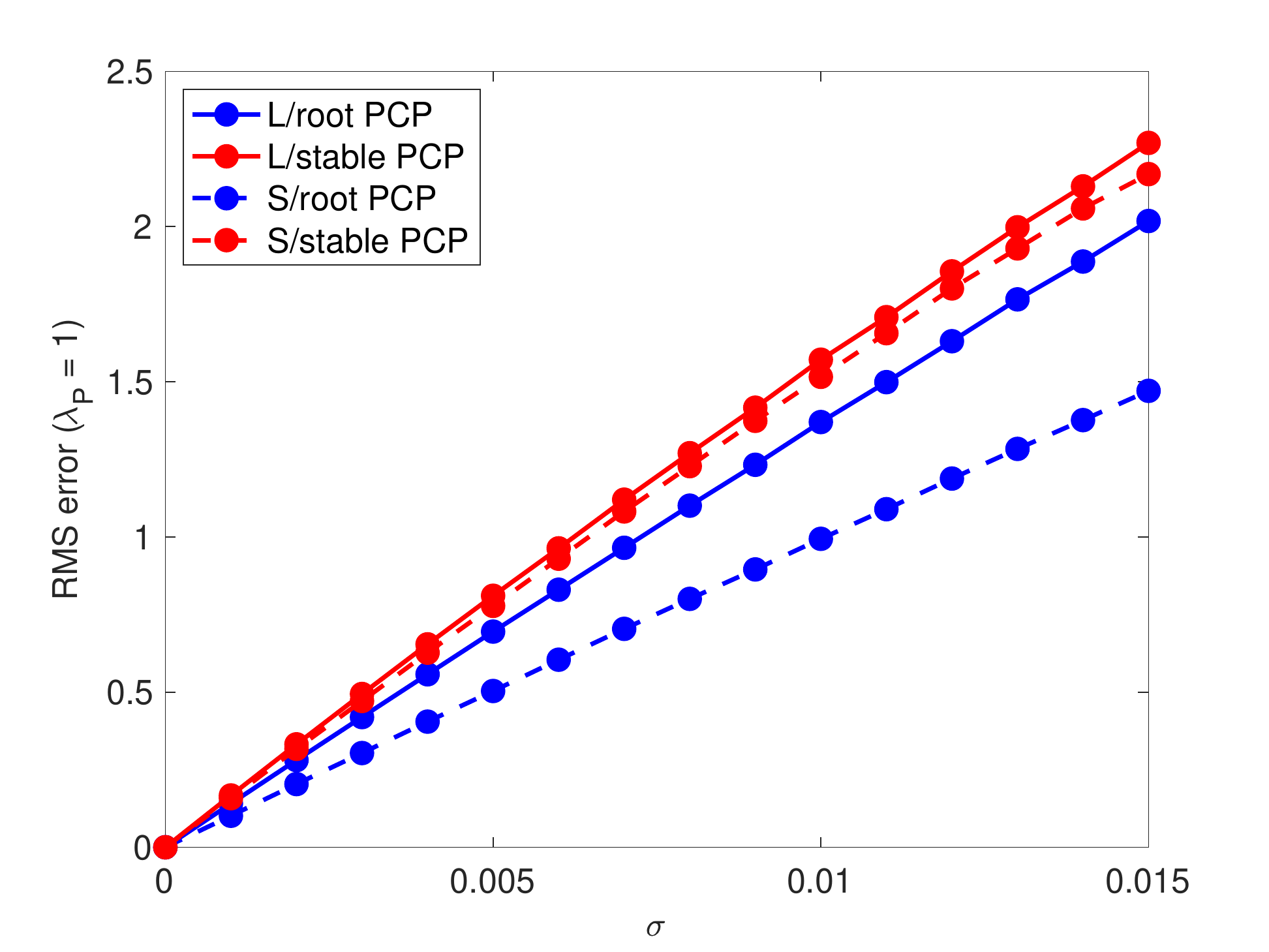}
\label{fig:rms_sigma_Poisson1}}
\end{subfigure}
\hspace{-0.9em}
\begin{subfigure}[Poisson $\lambda_P = 3$]{
  \includegraphics[width = 0.19\textwidth]{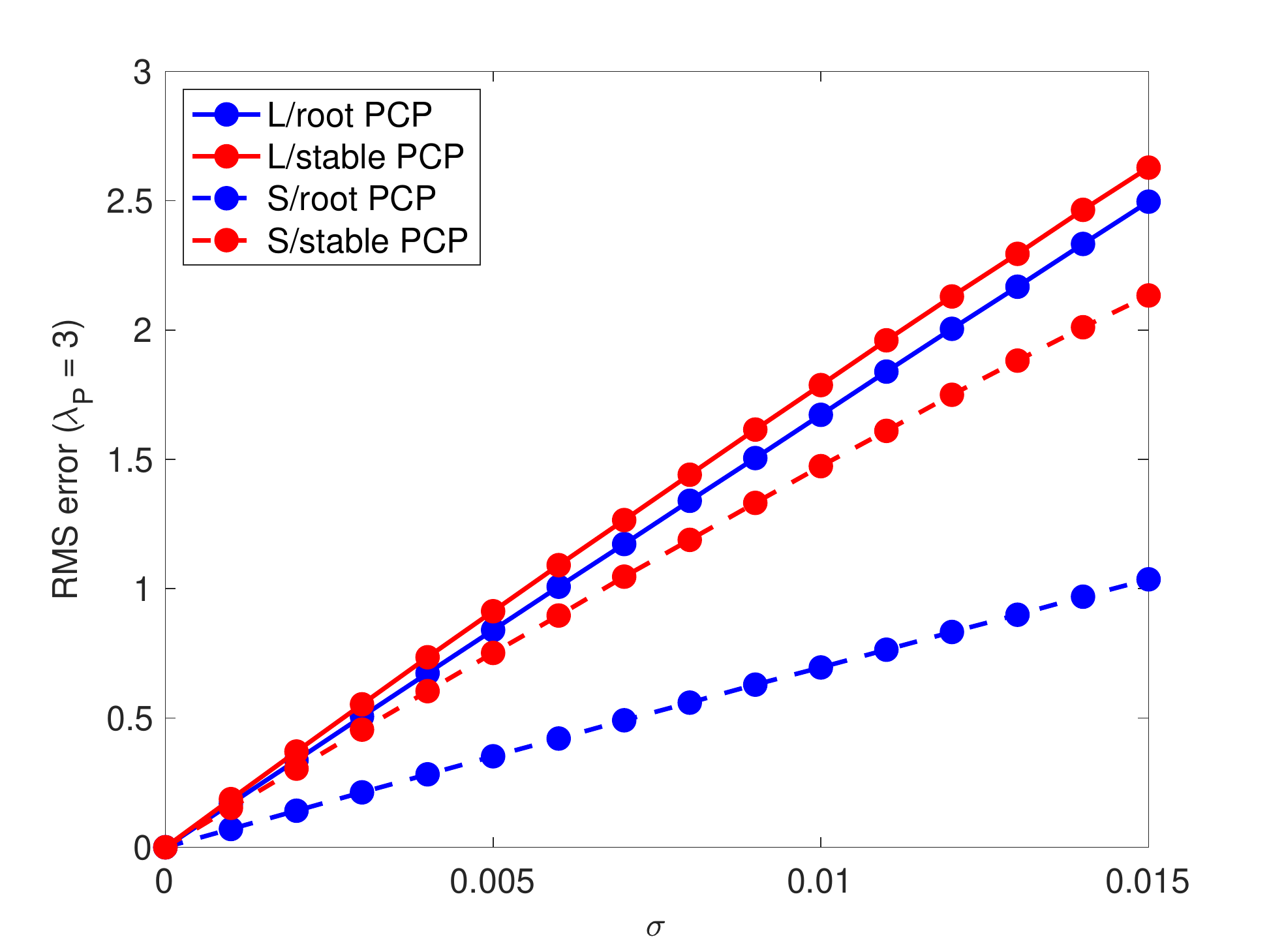}
\label{fig:rms_sigma_Poisson3}}
\end{subfigure}
\hspace{-0.9em}
\begin{subfigure}[Poisson $\lambda_P = 5$]{
  \includegraphics[width = 0.19\textwidth]{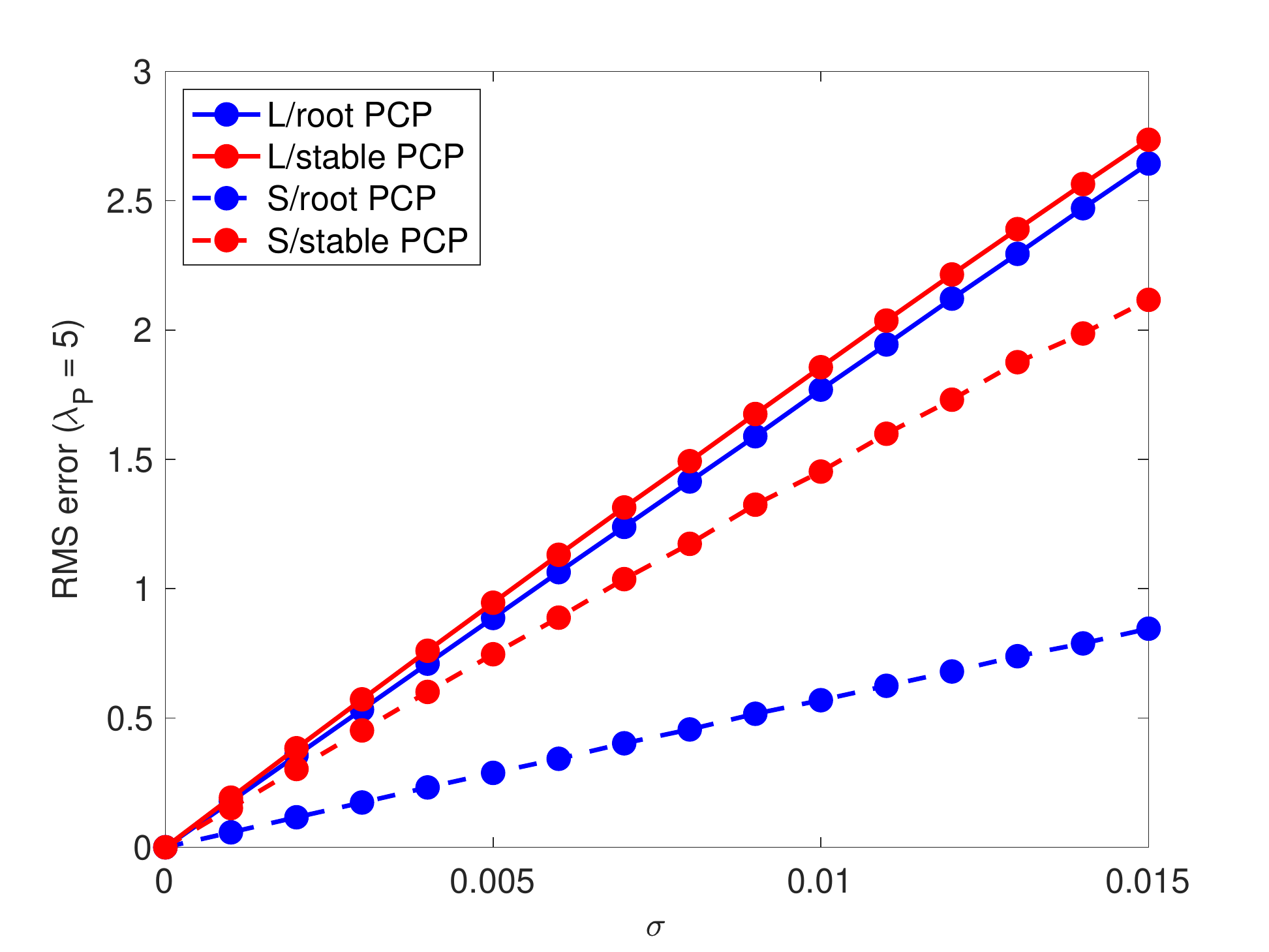}
\label{fig:rms_sigma_Poisson5}}
\end{subfigure}
\hspace{-0.9em}
\begin{subfigure}[Uniform]{
  \includegraphics[width = 0.19\textwidth]{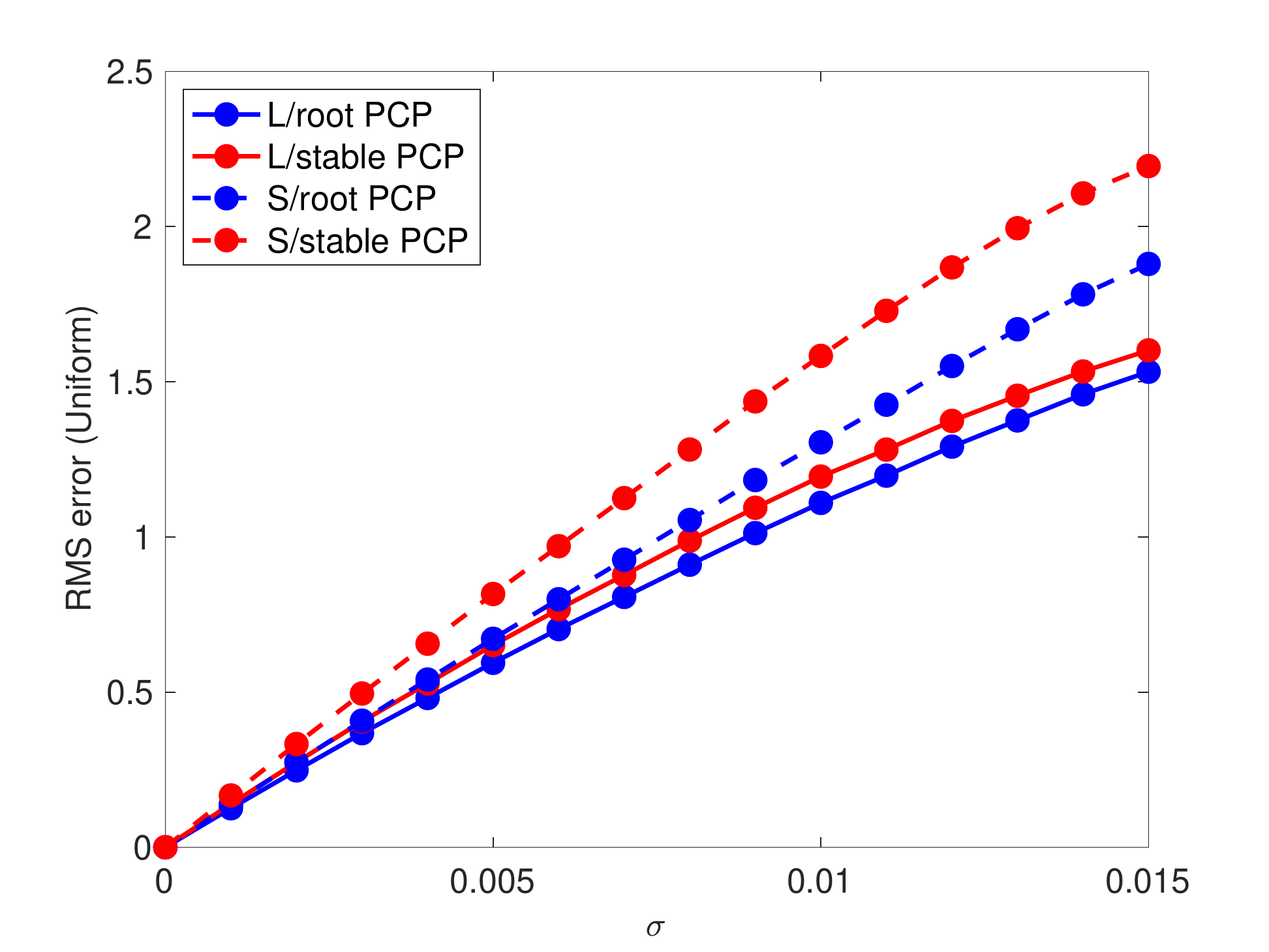}
\label{fig:rms_sigma_Uniform}}
\end{subfigure}
\vspace{-1em}
\caption{\stableU\ vs \rootpcp: effect of varying $\sigma$ for different noise distributions}
\label{fig:other_noise_sigma}
\end{figure}

\begin{figure}[h!]
\centering
\vspace{-1em}
\begin{subfigure}[Gaussian]{
  \includegraphics[width = 0.19\textwidth]{plots/simulation_varying_n/error_L_S-eps-converted-to.pdf}
\label{fig:rms_n_Gaussian}}
\end{subfigure}
\hspace{-0.9em}
\begin{subfigure}[Poisson $\lambda_P = 1$]{
  \includegraphics[width = 0.19\textwidth]{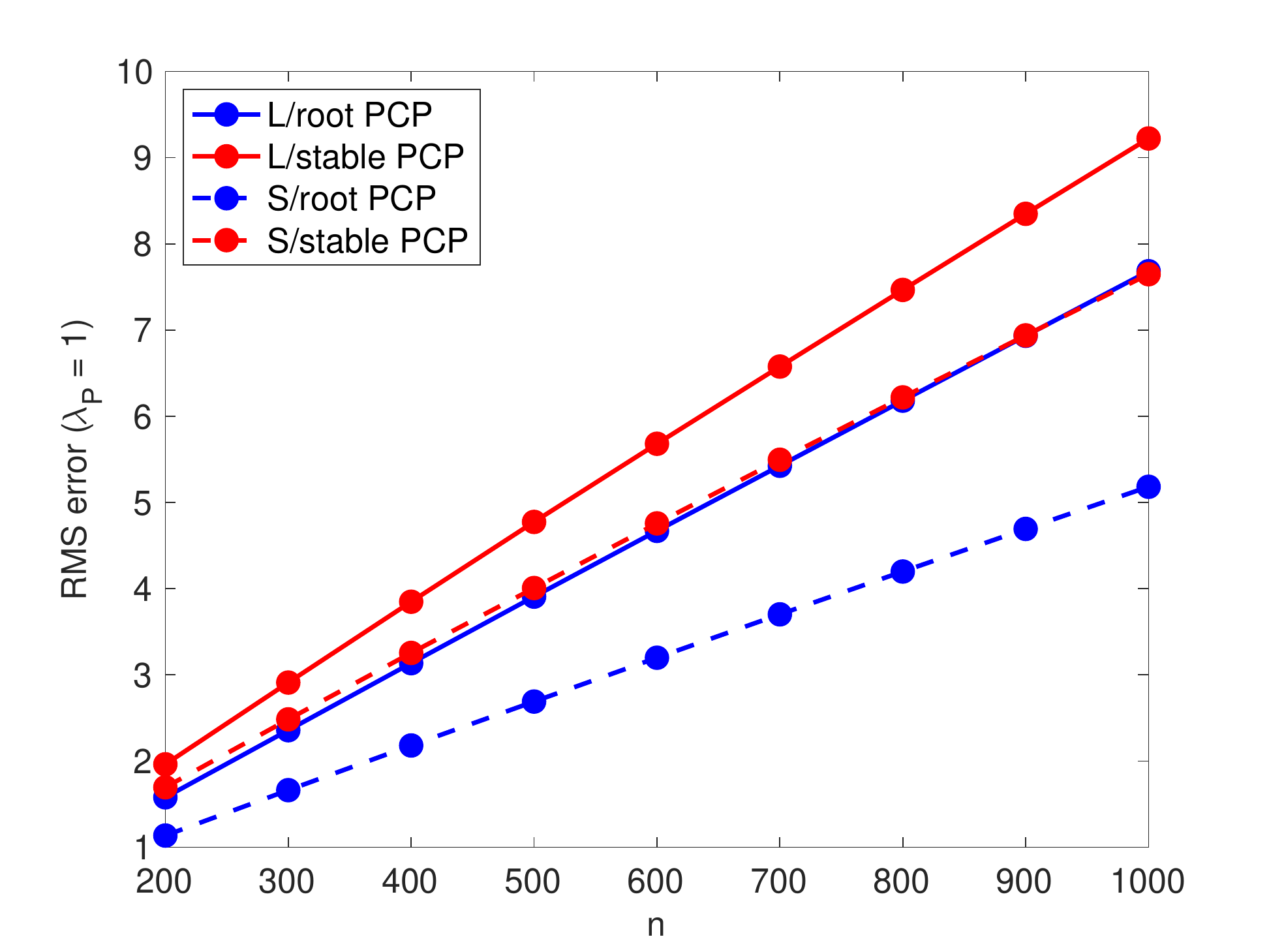}
\label{fig:rms_n_Poisson1}}
\end{subfigure}
\hspace{-0.9em}
\begin{subfigure}[Poisson $\lambda_P = 3$]{
  \includegraphics[width = 0.19\textwidth]{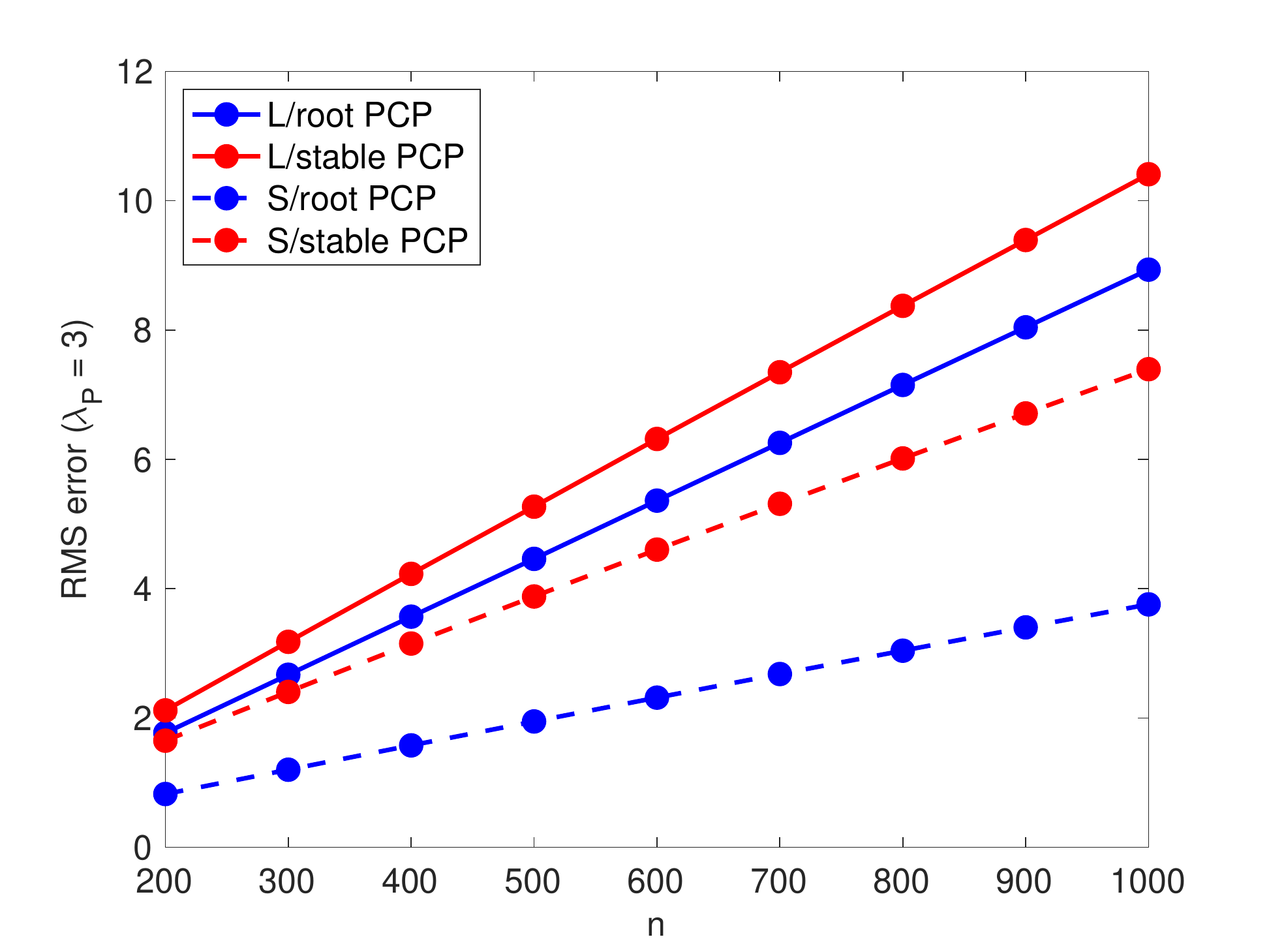}
\label{fig:rms_n_Poisson3}}
\end{subfigure}
\hspace{-0.9em}
\begin{subfigure}[Poisson $\lambda_P = 5$]{
  \includegraphics[width = 0.19\textwidth]{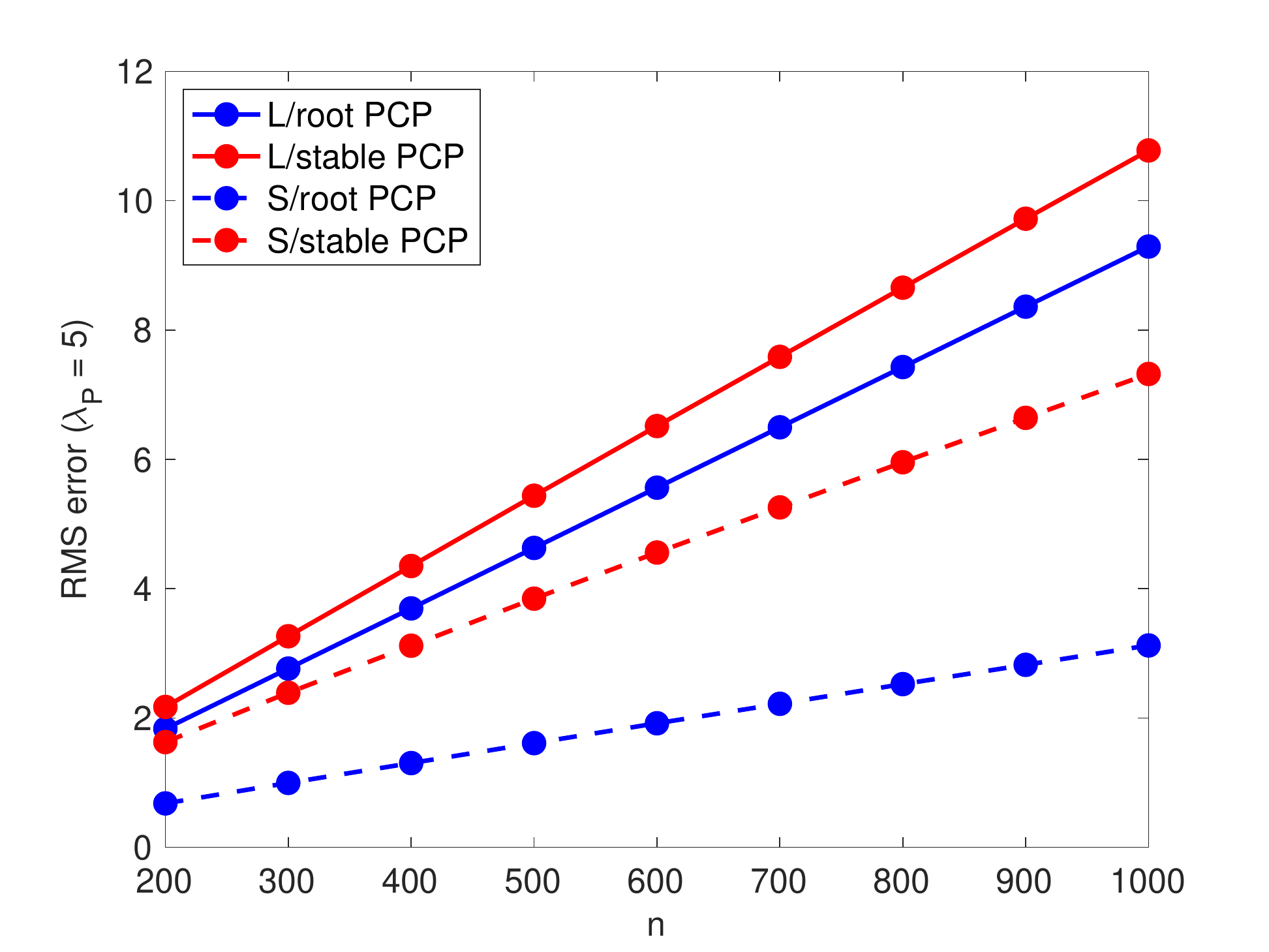}
\label{fig:rms_n_Poisson5}}
\end{subfigure}
\hspace{-0.9em}
\begin{subfigure}[Uniform]{
  \includegraphics[width = 0.19\textwidth]{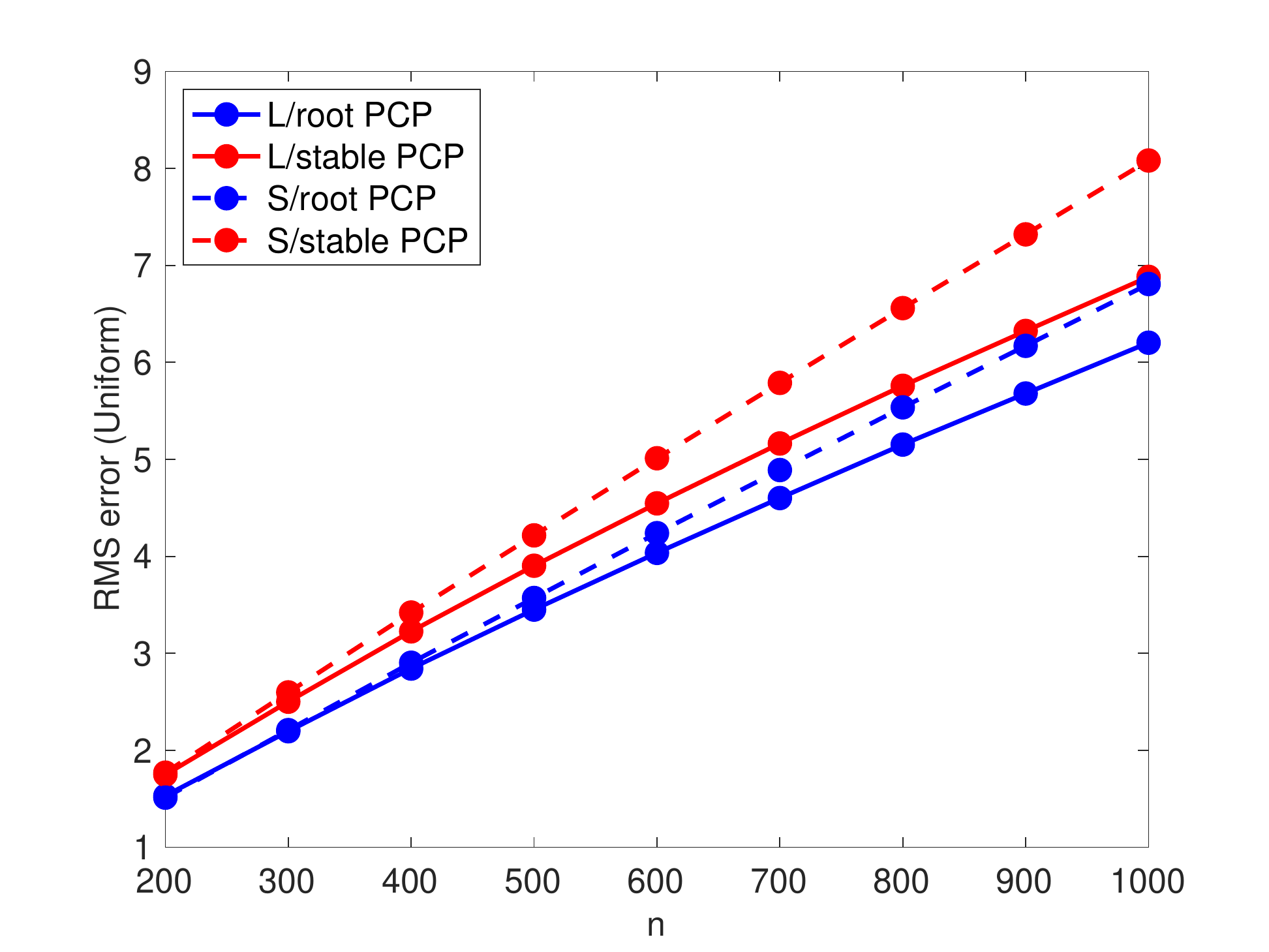}
\label{fig:rms_n_Uniform}}
\end{subfigure}
\vspace{-1em}
\caption{\stableU\ vs \rootpcp: effect of varying $n$ for different noise distributions}
\label{fig:other_noise_n}
\end{figure}

We test our \rootpcp and \stableU on simulation experiments with different noise distributions. All the setup and parameters are the same as in Section \ref{sec:simulation_exp_comparison} except that now, instead of adding Gaussian noise which follows $\mathcal{N}(0,\sigma^2)$, we add (scaled) Poisson noise, $l\cdot Poisson(\lambda_P)$ where we choose $\lambda_P \in \{1,3,5\}$\footnote{Since a Poisson variable with parameter $\lambda_P$ equals 0 with probability $e^{-\lambda_{P}}$, our choices of $\lambda_P$ give $\mb Z_0$'s which are approximately $36.79\%$, $4.98\%$, and $0.67\%$ sparse.} with scale $l = \frac{\sigma}{\sqrt{\lambda_P+\lambda_P^2}}$, and Uniform noise $Uniform(-\sqrt{3}\sigma,\sqrt{3}\sigma)$. We choose $(\lambda_P,l)$ and the range of the Uniform distribution in this way such that $E[(\mb Z_0)_{ij}^2]=\sigma^2$. Results are presented in Figures \ref{fig:other_noise_sigma} and \ref{fig:other_noise_n}. 

\subsection{Additional Results for Section \ref{section:surveillance_video}}
We run the experiments on a laptop with 2.3 GHz Dual-Core Intel Core i5, and we set the maximal iteration number of our ADMM to be 5000. For experiments that don't converge in 5000 steps, the number of iterations is represented as $5000+$.

For the hall dataset, we present the relative error for $\mb L,\mb S$, the running time, and the number of iteration in Table \ref{table:hall_root} and Table \ref{table:hall_stable}. 
\begin{table}[h!]
\begin{minipage}{.5\linewidth}
\caption{\rootpcp: Hall dataset}
\label{table:hall_root}
\vskip 0.15in
\begin{center}
\begin{small}
\begin{sc}
\begin{tabular}{l||c|c|c|c}
$\sigma$ & $\frac{\|\wh{\mb L}-\mb L_0\|_F}{\|\mb L_0\|_F} $& $\frac{\|\wh{\mb S}-\mb S_0\|_F}{\|\mb S_0\|_F}$& Time ($\times 10^3$ s)  & Iter\\
\hline
0    & 0.0019& 0.0266& 0.6894&  1688\\
30   & 0.0445& 0.7403& 0.5381&  1293\\
60   & 0.0737& 1.3525& 0.9470&  2425\\
90   & 0.0968& 1.9288& 1.4040&  3258\\
120  & 0.1200& 2.5067& 1.6230&  4113\\
\end{tabular}
\end{sc}
\end{small}
\end{center}
\vskip -0.1in
\end{minipage}
\qquad\quad
\begin{minipage}{.5\linewidth}
\caption{\stableU: Hall dataset}
\label{table:hall_stable}
\vskip 0.15in
\begin{center}
\begin{small}
\begin{sc}
\begin{tabular}{l||c|c|c|c}
$\sigma$ & $\frac{\|\wh{\mb L}-\mb L_0\|_F}{\|\mb L_0\|_F} $& $\frac{\|\wh{\mb S}-\mb S_0\|_F}{\|\mb S_0\|_F}$& Time ($\times 10^3$ s)  & Iter\\
\hline
0    & 0.0019& 0.0266& 1.7340 &  4423\\                      
30   & 0.0443& 0.7495& 1.8936 &  4918\\
60   & 0.0740& 1.3494& 1.3006 &  3362\\
90   & 0.0974& 1.8922& 1.8050 &  4563\\
120  & 0.1214& 2.4349& 1.9886 &  5000+\\
\end{tabular}
\end{sc}
\end{small}
\end{center}
\vskip -0.1in
\end{minipage}
\end{table}

\begin{figure}[h!]
\centering
\vspace{-2em}
\begin{subfigure}{
\makebox[20pt]{\raisebox{20pt}{\rotatebox[origin=c]{90}{}}}%
\makebox[0.16\textwidth]{video $ + \mb Z_0$}%
\makebox[0.16\textwidth]{$\wh{\mb L}_{\mathrm{root}}^{(\sigma)}$}%
\makebox[0.16\textwidth]{$\wh{\mb S}_{\mathrm{root}}^{(\sigma)}$}%
\makebox[0.16\textwidth]{$\wh{\mb L}_{\mathrm{stable}}^{(\sigma)}$}%
\makebox[0.16\textwidth]{$\wh{\mb S}_{\mathrm{stable}}^{(\sigma)}$}%
}
\end{subfigure}
\vspace{-0.5em}
\begin{subfigure}{
\makebox[20pt]{\raisebox{15pt}{\rotatebox[origin=c]{90}{$\sigma = 0$}}}%
\includegraphics[width = 0.15\textwidth]{plots/real_data_hall/original_sigma_1_frame_1.jpeg}}
\end{subfigure}
\hspace{-0.5em}
\begin{subfigure}{
  \includegraphics[width = 0.15\textwidth]{plots/real_data_hall/L_root_sigma_1_frame_1.jpeg}}
\end{subfigure}
\hspace{-0.5em}
\begin{subfigure}{
  \includegraphics[width = 0.15\textwidth]{plots/real_data_hall/S_root_sigma_1_frame_1.jpeg}}
\end{subfigure}
\hspace{-0.5em}
\begin{subfigure}{
  \includegraphics[width = 0.15\textwidth]{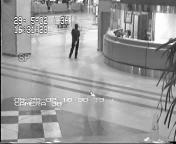}}
\end{subfigure}
\hspace{-0.5em}
\begin{subfigure}{
  \includegraphics[width = 0.15\textwidth]{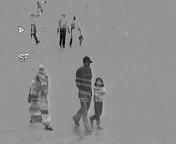}}
\end{subfigure}
\vspace{-0.5em}
\begin{subfigure}{
\makebox[20pt]{\raisebox{15pt}{\rotatebox[origin=c]{90}{$\sigma = 30$}}}%
\includegraphics[width = 0.15\textwidth]{plots/real_data_hall/original_sigma_2_frame_1.jpeg}}
\end{subfigure}
\hspace{-0.5em}
\begin{subfigure}{
  \includegraphics[width = 0.15\textwidth]{plots/real_data_hall/L_root_sigma_2_frame_1.jpeg}}
\end{subfigure}
\hspace{-0.5em}
\begin{subfigure}{
  \includegraphics[width = 0.15\textwidth]{plots/real_data_hall/S_root_sigma_2_frame_1.jpeg}}
\end{subfigure}
\hspace{-0.5em}
\begin{subfigure}{
  \includegraphics[width = 0.15\textwidth]{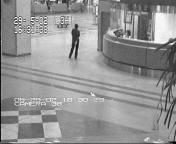}}
\end{subfigure}
\hspace{-0.5em}
\begin{subfigure}{
  \includegraphics[width = 0.15\textwidth]{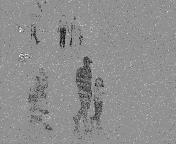}}
\end{subfigure}
\vspace{-0.5em}
\begin{subfigure}{
\makebox[20pt]{\raisebox{15pt}{\rotatebox[origin=c]{90}{$\sigma = 60$}}}%
\includegraphics[width = 0.15\textwidth]{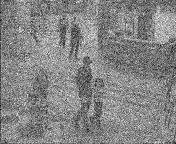}}
\end{subfigure}
\hspace{-0.5em}
\begin{subfigure}{
  \includegraphics[width = 0.15\textwidth]{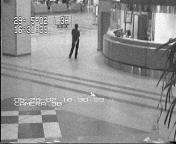}}
\end{subfigure}
\hspace{-0.5em}
\begin{subfigure}{
  \includegraphics[width = 0.15\textwidth]{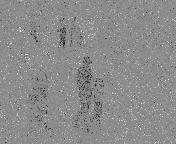}}
\end{subfigure}
\hspace{-0.5em}
\begin{subfigure}{
  \includegraphics[width = 0.15\textwidth]{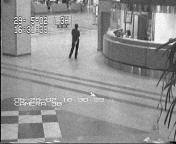}}
\end{subfigure}
\hspace{-0.5em}
\begin{subfigure}{
  \includegraphics[width = 0.15\textwidth]{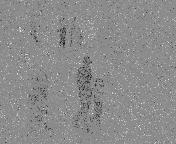}}
\end{subfigure}
\vspace{-0.5em}
\begin{subfigure}{
\makebox[20pt]{\raisebox{15pt}{\rotatebox[origin=c]{90}{$\sigma = 90$}}}%
\includegraphics[width = 0.15\textwidth]{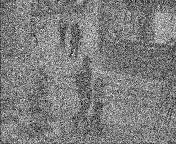}}
\end{subfigure}
\hspace{-0.5em}
\begin{subfigure}{
  \includegraphics[width = 0.15\textwidth]{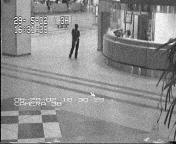}}
\end{subfigure}
\hspace{-0.5em}
\begin{subfigure}{
  \includegraphics[width = 0.15\textwidth]{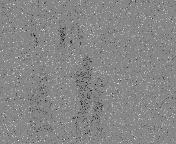}}
\end{subfigure}
\hspace{-0.5em}
\begin{subfigure}{
  \includegraphics[width = 0.15\textwidth]{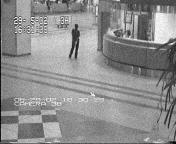}}
\end{subfigure}
\hspace{-0.5em}
\begin{subfigure}{
  \includegraphics[width = 0.15\textwidth]{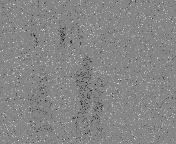}}
\end{subfigure}
\vspace{-0.5em}
\begin{subfigure}{
\makebox[20pt]{\raisebox{15pt}{\rotatebox[origin=c]{90}{$\sigma = 120$}}}%
\includegraphics[width = 0.15\textwidth]{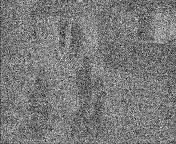}}
\end{subfigure}
\hspace{-0.5em}
\begin{subfigure}{
  \includegraphics[width = 0.15\textwidth]{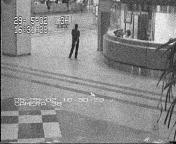}}
\end{subfigure}
\hspace{-0.5em}
\begin{subfigure}{
  \includegraphics[width = 0.15\textwidth]{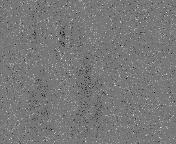}}
\end{subfigure}
\hspace{-0.5em}
\begin{subfigure}{
  \includegraphics[width = 0.15\textwidth]{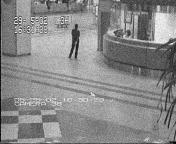}}
\end{subfigure}
\hspace{-0.5em}
\begin{subfigure}{
  \includegraphics[width = 0.15\textwidth]{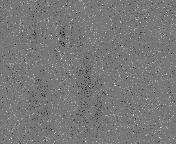}}
\end{subfigure}
\caption{hall: recovered $\wh{\mb L},\wh{\mb S}$ for frame 1}
\label{fig:hall_frame1_all}
\end{figure}

\begin{figure}[h!]
\centering
\vspace{-2em}
\begin{subfigure}{
\makebox[20pt]{\raisebox{20pt}{\rotatebox[origin=c]{90}{}}}%
\makebox[0.16\textwidth]{video $ + \mb Z_0$}%
\makebox[0.16\textwidth]{$\wh{\mb L}_{\mathrm{root}}^{(\sigma)}$}%
\makebox[0.16\textwidth]{$\wh{\mb S}_{\mathrm{root}}^{(\sigma)}$}%
\makebox[0.16\textwidth]{$\wh{\mb L}_{\mathrm{stable}}^{(\sigma)}$}%
\makebox[0.16\textwidth]{$\wh{\mb S}_{\mathrm{stable}}^{(\sigma)}$}%
}
\end{subfigure}
\vspace{-0.5em}
\begin{subfigure}{
\makebox[20pt]{\raisebox{15pt}{\rotatebox[origin=c]{90}{$\sigma = 0$}}}%
\includegraphics[width = 0.15\textwidth]{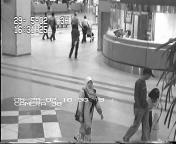}}
\end{subfigure}
\hspace{-0.5em}
\begin{subfigure}{
  \includegraphics[width = 0.15\textwidth]{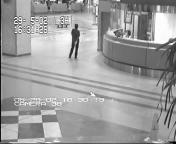}}
\end{subfigure}
\hspace{-0.5em}
\begin{subfigure}{
  \includegraphics[width = 0.15\textwidth]{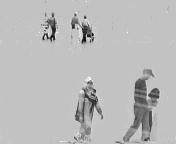}}
\end{subfigure}
\hspace{-0.5em}
\begin{subfigure}{
  \includegraphics[width = 0.15\textwidth]{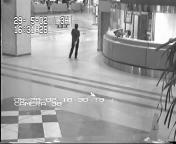}}
\end{subfigure}
\hspace{-0.5em}
\begin{subfigure}{
  \includegraphics[width = 0.15\textwidth]{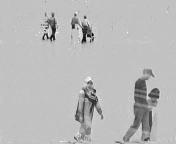}}
\end{subfigure}
\vspace{-0.5em}
\begin{subfigure}{
\makebox[20pt]{\raisebox{15pt}{\rotatebox[origin=c]{90}{$\sigma = 30$}}}%
\includegraphics[width = 0.15\textwidth]{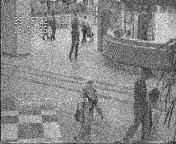}}
\end{subfigure}
\hspace{-0.5em}
\begin{subfigure}{
  \includegraphics[width = 0.15\textwidth]{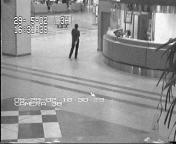}}
\end{subfigure}
\hspace{-0.5em}
\begin{subfigure}{
  \includegraphics[width = 0.15\textwidth]{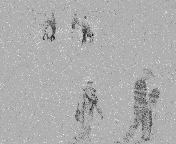}}
\end{subfigure}
\hspace{-0.5em}
\begin{subfigure}{
  \includegraphics[width = 0.15\textwidth]{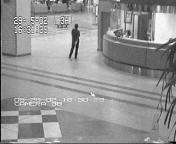}}
\end{subfigure}
\hspace{-0.5em}
\begin{subfigure}{
  \includegraphics[width = 0.15\textwidth]{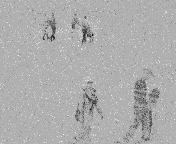}}
\end{subfigure}
\vspace{-0.5em}
\begin{subfigure}{
\makebox[20pt]{\raisebox{15pt}{\rotatebox[origin=c]{90}{$\sigma = 60$}}}%
\includegraphics[width = 0.15\textwidth]{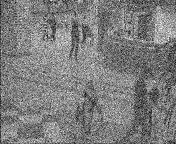}}
\end{subfigure}
\hspace{-0.5em}
\begin{subfigure}{
  \includegraphics[width = 0.15\textwidth]{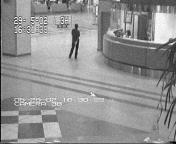}}
\end{subfigure}
\hspace{-0.5em}
\begin{subfigure}{
  \includegraphics[width = 0.15\textwidth]{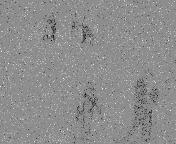}}
\end{subfigure}
\hspace{-0.5em}
\begin{subfigure}{
  \includegraphics[width = 0.15\textwidth]{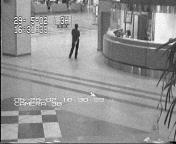}}
\end{subfigure}
\hspace{-0.5em}
\begin{subfigure}{
  \includegraphics[width = 0.15\textwidth]{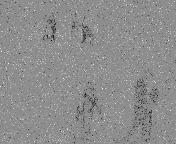}}
\end{subfigure}
\vspace{-0.5em}
\begin{subfigure}{
\makebox[20pt]{\raisebox{15pt}{\rotatebox[origin=c]{90}{$\sigma = 90$}}}%
\includegraphics[width = 0.15\textwidth]{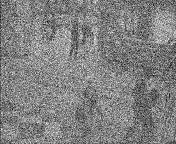}}
\end{subfigure}
\hspace{-0.5em}
\begin{subfigure}{
  \includegraphics[width = 0.15\textwidth]{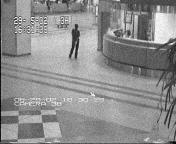}}
\end{subfigure}
\hspace{-0.5em}
\begin{subfigure}{
  \includegraphics[width = 0.15\textwidth]{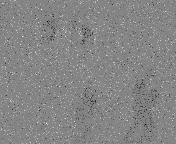}}
\end{subfigure}
\hspace{-0.5em}
\begin{subfigure}{
  \includegraphics[width = 0.15\textwidth]{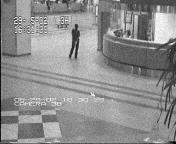}}
\end{subfigure}
\hspace{-0.5em}
\begin{subfigure}{
  \includegraphics[width = 0.15\textwidth]{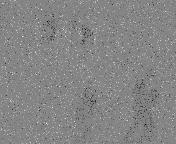}}
\end{subfigure}
\vspace{-0.5em}
\begin{subfigure}{
\makebox[20pt]{\raisebox{15pt}{\rotatebox[origin=c]{90}{$\sigma = 120$}}}%
\includegraphics[width = 0.15\textwidth]{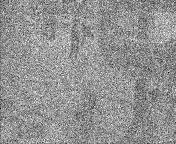}}
\end{subfigure}
\hspace{-0.5em}
\begin{subfigure}{
  \includegraphics[width = 0.15\textwidth]{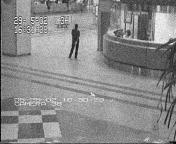}}
\end{subfigure}
\hspace{-0.5em}
\begin{subfigure}{
  \includegraphics[width = 0.15\textwidth]{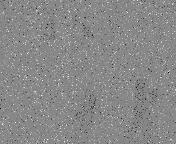}}
\end{subfigure}
\hspace{-0.5em}
\begin{subfigure}{
  \includegraphics[width = 0.15\textwidth]{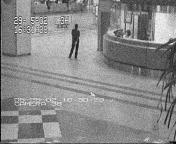}}
\end{subfigure}
\hspace{-0.5em}
\begin{subfigure}{
  \includegraphics[width = 0.15\textwidth]{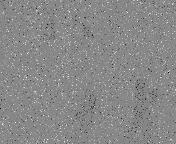}}
\end{subfigure}
\caption{hall: recovered $\wh{\mb L},\wh{\mb S}$ for frame 20}
\label{fig:hall_frame20_all}
\end{figure}

In Figure \ref{fig:hall_frame1_all} and \ref{fig:hall_frame20_all}, we present more results for frame 1, 20 for varying $\sigma$ for this hall dataset.

We also apply our algorithms to the dataset lights. We present the relative error for $\mb L,\mb S$, the running time, and the number of iteration in Table \ref{table:light_root} and Table \ref{table:light_stable}. 

\begin{table}[h!]
\begin{minipage}{.5\linewidth}
\caption{\rootpcp: Lights dataset}
\label{table:light_root}
\vskip 0.15in
\begin{center}
\begin{small}
\begin{sc}
\begin{tabular}{l||c|c|c|c}
$\sigma$ & $\frac{\|\wh{\mb L}-\mb L_0\|_F}{\|\mb L_0\|_F} $& $\frac{\|\wh{\mb S}-\mb S_0\|_F}{\|\mb S_0\|_F}$& Time ($\times 10^3$ s)  & Iter\\
\hline
0    & 0.0013& 0.1052& 0.7100&  1605\\
30   & 0.0520& 2.9707& 0.5120&  1155\\
60   & 0.0938& 5.8908& 1.7101&  3880\\
90   & 0.1323& 8.7761& 2.1983&  5000+\\
120  & 0.1689& 11.5848& 2.5344&  5000+\\
\end{tabular}
\end{sc}
\end{small}
\end{center}
\vskip -0.1in
\end{minipage}
\qquad\quad
\begin{minipage}{.5\linewidth}
\caption{\stableU: Lights dataset}
\label{table:light_stable}
\vskip 0.15in
\begin{center}
\begin{small}
\begin{sc}
\begin{tabular}{l||c|c|c|c}
$\sigma$ & $\frac{\|\wh{\mb L}-\mb L_0\|_F}{\|\mb L_0\|_F} $& $\frac{\|\wh{\mb S}-\mb S_0\|_F}{\|\mb S_0\|_F}$& Time ($\times 10^3$ s)  & Iter\\
\hline
0    & 0.0013& 0.1052& 2.2069&  5000+\\ 
30   & 0.0527& 2.7920& 1.4677&  3377\\
60   & 0.0951& 5.5085& 1.2136&  2804\\
90   & 0.1342& 8.2023& 1.6699&  3886\\
120  & 0.1710& 10.8930& 2.0856&  4902\\
\end{tabular}
\end{sc}
\end{small}
\end{center}
\vskip -0.1in
\end{minipage}
\end{table}

\begin{figure}[h!]
\centering
\vspace{-2em}
\begin{subfigure}{
\makebox[20pt]{\raisebox{20pt}{\rotatebox[origin=c]{90}{}}}%
\makebox[0.16\textwidth]{video $ + \mb Z_0$}%
\makebox[0.16\textwidth]{$\wh{\mb L}_{\mathrm{root}}^{(\sigma)}$}%
\makebox[0.16\textwidth]{$\wh{\mb S}_{\mathrm{root}}^{(\sigma)}$}%
\makebox[0.16\textwidth]{$\wh{\mb L}_{\mathrm{stable}}^{(\sigma)}$}%
\makebox[0.16\textwidth]{$\wh{\mb S}_{\mathrm{stable}}^{(\sigma)}$}%
}
\end{subfigure}
\vspace{-0.5em}
\begin{subfigure}{
\makebox[20pt]{\raisebox{15pt}{\rotatebox[origin=c]{90}{$\sigma = 0$}}}%
\includegraphics[width = 0.15\textwidth]{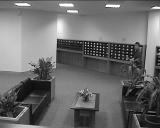}}
\end{subfigure}
\hspace{-0.5em}
\begin{subfigure}{
  \includegraphics[width = 0.15\textwidth]{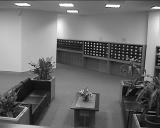}}
\end{subfigure}
\hspace{-0.5em}
\begin{subfigure}{
  \includegraphics[width = 0.15\textwidth]{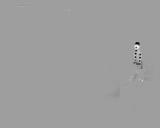}}
\end{subfigure}
\hspace{-0.5em}
\begin{subfigure}{
  \includegraphics[width = 0.15\textwidth]{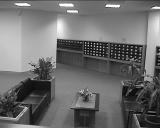}}
\end{subfigure}
\hspace{-0.5em}
\begin{subfigure}{
  \includegraphics[width = 0.15\textwidth]{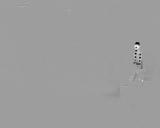}}
\end{subfigure}
\vspace{-0.5em}
\begin{subfigure}{
\makebox[20pt]{\raisebox{15pt}{\rotatebox[origin=c]{90}{$\sigma = 30$}}}%
\includegraphics[width = 0.15\textwidth]{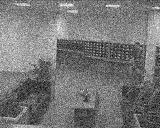}}
\end{subfigure}
\hspace{-0.5em}
\begin{subfigure}{
  \includegraphics[width = 0.15\textwidth]{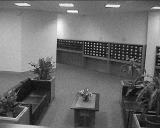}}
\end{subfigure}
\hspace{-0.5em}
\begin{subfigure}{
  \includegraphics[width = 0.15\textwidth]{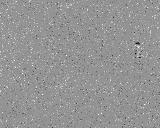}}
\end{subfigure}
\hspace{-0.5em}
\begin{subfigure}{
  \includegraphics[width = 0.15\textwidth]{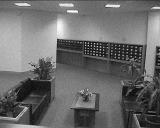}}
\end{subfigure}
\hspace{-0.5em}
\begin{subfigure}{
  \includegraphics[width = 0.15\textwidth]{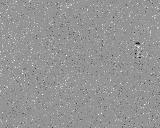}}
\end{subfigure}
\vspace{-0.5em}
\begin{subfigure}{
\makebox[20pt]{\raisebox{15pt}{\rotatebox[origin=c]{90}{$\sigma = 60$}}}%
\includegraphics[width = 0.15\textwidth]{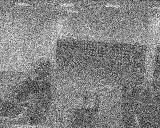}}
\end{subfigure}
\hspace{-0.5em}
\begin{subfigure}{
  \includegraphics[width = 0.15\textwidth]{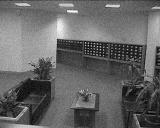}}
\end{subfigure}
\hspace{-0.5em}
\begin{subfigure}{
  \includegraphics[width = 0.15\textwidth]{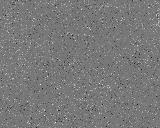}}
\end{subfigure}
\hspace{-0.5em}
\begin{subfigure}{
  \includegraphics[width = 0.15\textwidth]{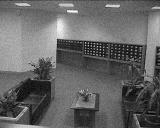}}
\end{subfigure}
\hspace{-0.5em}
\begin{subfigure}{
  \includegraphics[width = 0.15\textwidth]{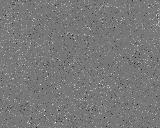}}
\end{subfigure}
\vspace{-0.5em}
\begin{subfigure}{
\makebox[20pt]{\raisebox{15pt}{\rotatebox[origin=c]{90}{$\sigma = 90$}}}%
\includegraphics[width = 0.15\textwidth]{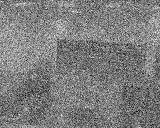}}
\end{subfigure}
\hspace{-0.5em}
\begin{subfigure}{
  \includegraphics[width = 0.15\textwidth]{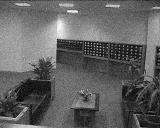}}
\end{subfigure}
\hspace{-0.5em}
\begin{subfigure}{
  \includegraphics[width = 0.15\textwidth]{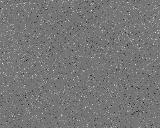}}
\end{subfigure}
\hspace{-0.5em}
\begin{subfigure}{
  \includegraphics[width = 0.15\textwidth]{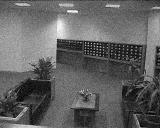}}
\end{subfigure}
\hspace{-0.5em}
\begin{subfigure}{
  \includegraphics[width = 0.15\textwidth]{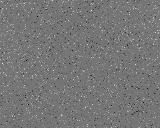}}
\end{subfigure}
\vspace{-0.5em}
\begin{subfigure}{
\makebox[20pt]{\raisebox{15pt}{\rotatebox[origin=c]{90}{$\sigma = 120$}}}%
\includegraphics[width = 0.15\textwidth]{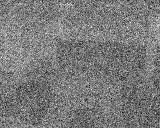}}
\end{subfigure}
\hspace{-0.5em}
\begin{subfigure}{
  \includegraphics[width = 0.15\textwidth]{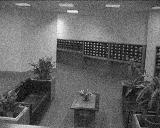}}
\end{subfigure}
\hspace{-0.5em}
\begin{subfigure}{
  \includegraphics[width = 0.15\textwidth]{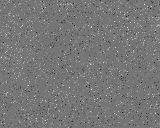}}
\end{subfigure}
\hspace{-0.5em}
\begin{subfigure}{
  \includegraphics[width = 0.15\textwidth]{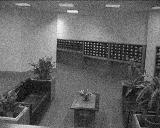}}
\end{subfigure}
\hspace{-0.5em}
\begin{subfigure}{
  \includegraphics[width = 0.15\textwidth]{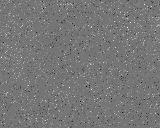}}
\end{subfigure}
\caption{lights: recovered $\wh{\mb L},\wh{\mb S}$ for frame 1}
\label{fig:lights_frame1_all}
\end{figure}

\begin{figure}[h!]
\centering
\vspace{-2em}
\begin{subfigure}{
\makebox[20pt]{\raisebox{20pt}{\rotatebox[origin=c]{90}{}}}%
\makebox[0.16\textwidth]{video $ + \mb Z_0$}%
\makebox[0.16\textwidth]{$\wh{\mb L}_{\mathrm{root}}^{(\sigma)}$}%
\makebox[0.16\textwidth]{$\wh{\mb S}_{\mathrm{root}}^{(\sigma)}$}%
\makebox[0.16\textwidth]{$\wh{\mb L}_{\mathrm{stable}}^{(\sigma)}$}%
\makebox[0.16\textwidth]{$\wh{\mb S}_{\mathrm{stable}}^{(\sigma)}$}%
}
\end{subfigure}
\vspace{-0.5em}
\begin{subfigure}{
\makebox[20pt]{\raisebox{15pt}{\rotatebox[origin=c]{90}{$\sigma = 0$}}}%
\includegraphics[width = 0.15\textwidth]{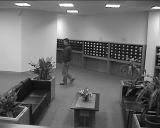}}
\end{subfigure}
\hspace{-0.5em}
\begin{subfigure}{
  \includegraphics[width = 0.15\textwidth]{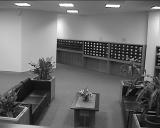}}
\end{subfigure}
\hspace{-0.5em}
\begin{subfigure}{
  \includegraphics[width = 0.15\textwidth]{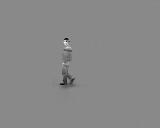}}
\end{subfigure}
\hspace{-0.5em}
\begin{subfigure}{
  \includegraphics[width = 0.15\textwidth]{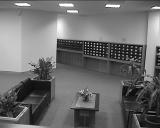}}
\end{subfigure}
\hspace{-0.5em}
\begin{subfigure}{
  \includegraphics[width = 0.15\textwidth]{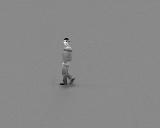}}
\end{subfigure}
\vspace{-0.5em}
\begin{subfigure}{
\makebox[20pt]{\raisebox{15pt}{\rotatebox[origin=c]{90}{$\sigma = 30$}}}%
\includegraphics[width = 0.15\textwidth]{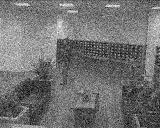}}
\end{subfigure}
\hspace{-0.5em}
\begin{subfigure}{
  \includegraphics[width = 0.15\textwidth]{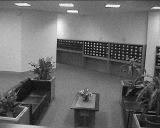}}
\end{subfigure}
\hspace{-0.5em}
\begin{subfigure}{
  \includegraphics[width = 0.15\textwidth]{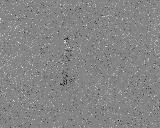}}
\end{subfigure}
\hspace{-0.5em}
\begin{subfigure}{
  \includegraphics[width = 0.15\textwidth]{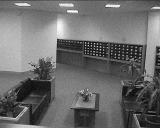}}
\end{subfigure}
\hspace{-0.5em}
\begin{subfigure}{
  \includegraphics[width = 0.15\textwidth]{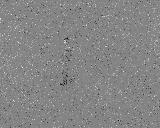}}
\end{subfigure}
\vspace{-0.5em}
\begin{subfigure}{
\makebox[20pt]{\raisebox{15pt}{\rotatebox[origin=c]{90}{$\sigma = 60$}}}%
\includegraphics[width = 0.15\textwidth]{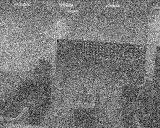}}
\end{subfigure}
\hspace{-0.5em}
\begin{subfigure}{
  \includegraphics[width = 0.15\textwidth]{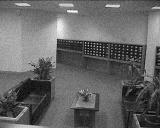}}
\end{subfigure}
\hspace{-0.5em}
\begin{subfigure}{
  \includegraphics[width = 0.15\textwidth]{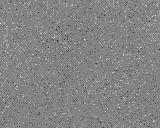}}
\end{subfigure}
\hspace{-0.5em}
\begin{subfigure}{
  \includegraphics[width = 0.15\textwidth]{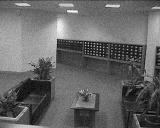}}
\end{subfigure}
\hspace{-0.5em}
\begin{subfigure}{
  \includegraphics[width = 0.15\textwidth]{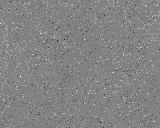}}
\end{subfigure}
\vspace{-0.5em}
\begin{subfigure}{
\makebox[20pt]{\raisebox{15pt}{\rotatebox[origin=c]{90}{$\sigma = 90$}}}%
\includegraphics[width = 0.15\textwidth]{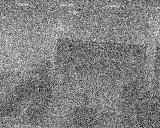}}
\end{subfigure}
\hspace{-0.5em}
\begin{subfigure}{
  \includegraphics[width = 0.15\textwidth]{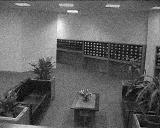}}
\end{subfigure}
\hspace{-0.5em}
\begin{subfigure}{
  \includegraphics[width = 0.15\textwidth]{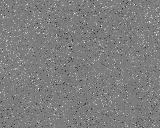}}
\end{subfigure}
\hspace{-0.5em}
\begin{subfigure}{
  \includegraphics[width = 0.15\textwidth]{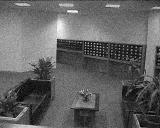}}
\end{subfigure}
\hspace{-0.5em}
\begin{subfigure}{
  \includegraphics[width = 0.15\textwidth]{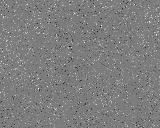}}
\end{subfigure}
\vspace{-0.5em}
\begin{subfigure}{
\makebox[20pt]{\raisebox{15pt}{\rotatebox[origin=c]{90}{$\sigma = 120$}}}%
\includegraphics[width = 0.15\textwidth]{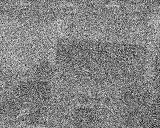}}
\end{subfigure}
\hspace{-0.5em}
\begin{subfigure}{
  \includegraphics[width = 0.15\textwidth]{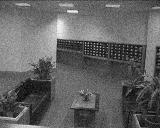}}
\end{subfigure}
\hspace{-0.5em}
\begin{subfigure}{
  \includegraphics[width = 0.15\textwidth]{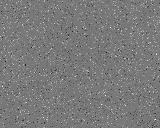}}
\end{subfigure}
\hspace{-0.5em}
\begin{subfigure}{
  \includegraphics[width = 0.15\textwidth]{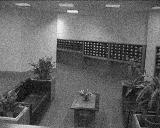}}
\end{subfigure}
\hspace{-0.5em}
\begin{subfigure}{
  \includegraphics[width = 0.15\textwidth]{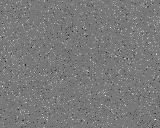}}
\end{subfigure}
\caption{lights: recovered $\wh{\mb L},\wh{\mb S}$ for frame 20}
\label{fig:lights_frame20_all}
\end{figure}

In Figure \ref{fig:lights_frame1_all} and \ref{fig:lights_frame20_all}, we present more results for frame 1, 20 for varying $\sigma$ for this lights dataset.

\begin{figure}[h!]
\centering
\begin{subfigure}[``lights``+noise]{
  \includegraphics[width = 0.19\textwidth]{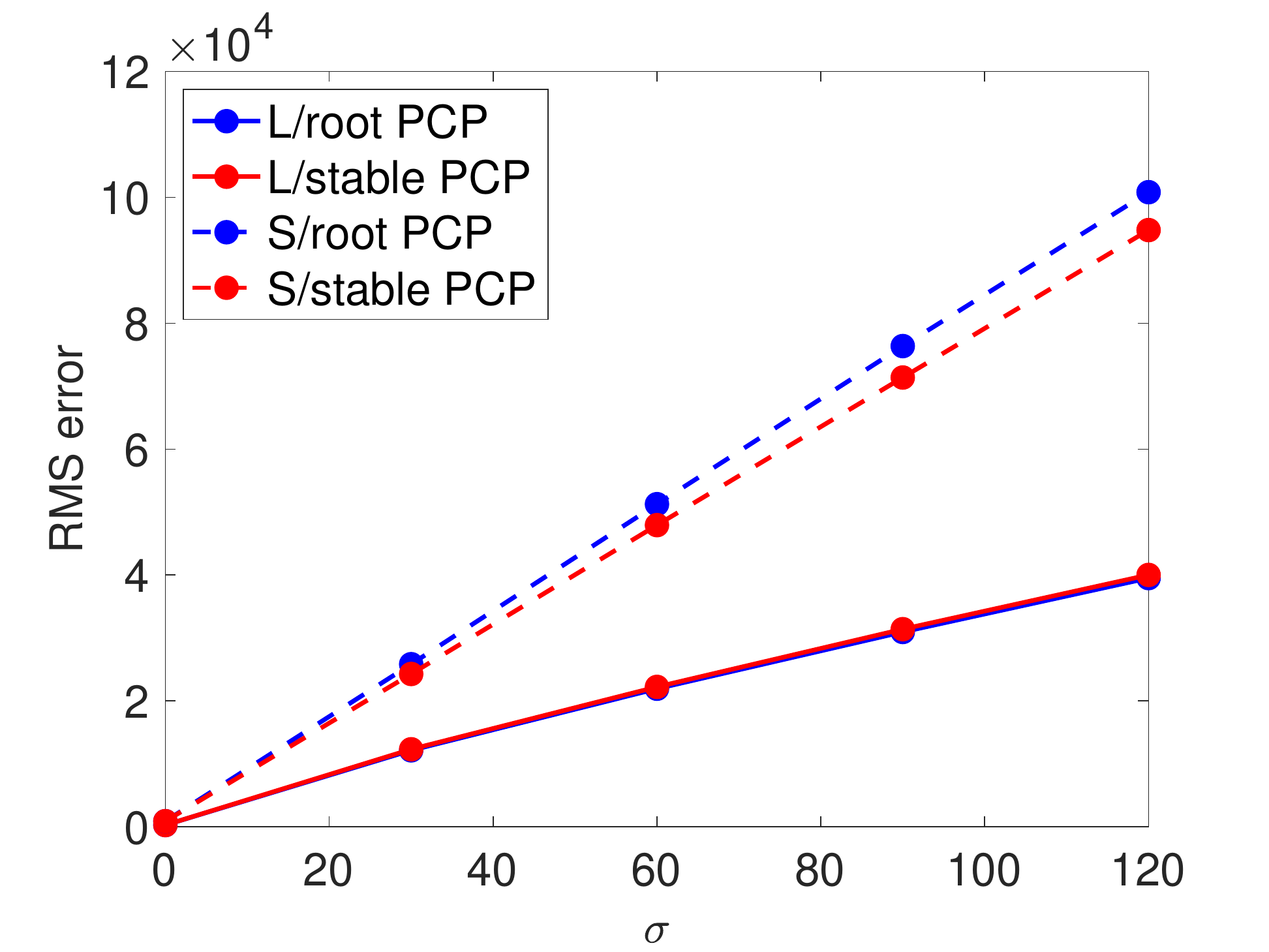}
\label{fig:lights_error}}
\end{subfigure}
\hspace{-0.9em}
\begin{subfigure}[yaleB01 + noise]{
  \includegraphics[width = 0.19\textwidth]{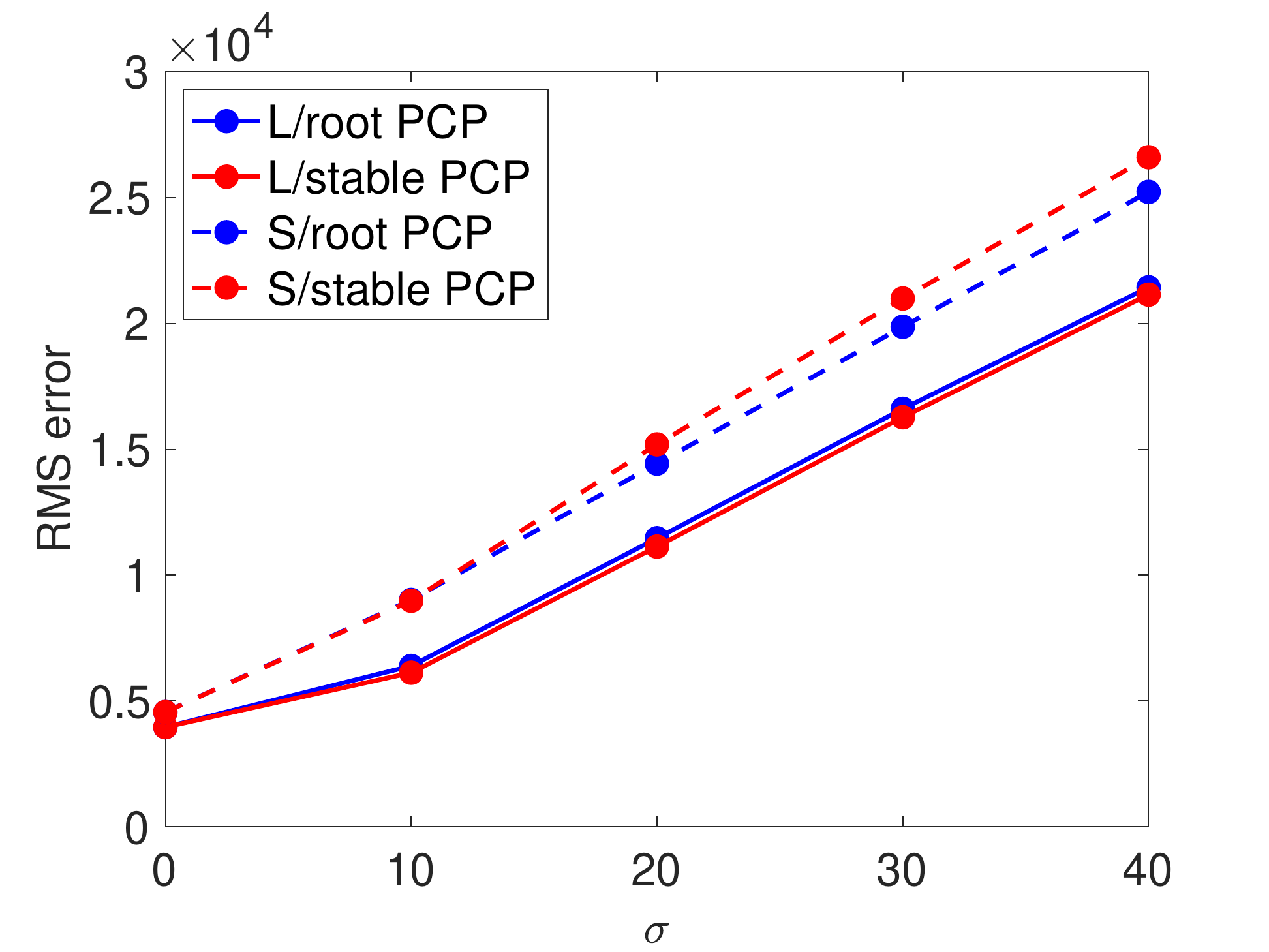}
\label{fig:yaleB01_error}}
\end{subfigure}
\hspace{-0.9em}
\begin{subfigure}[yaleB02 + noise]{
  \includegraphics[width = 0.19\textwidth]{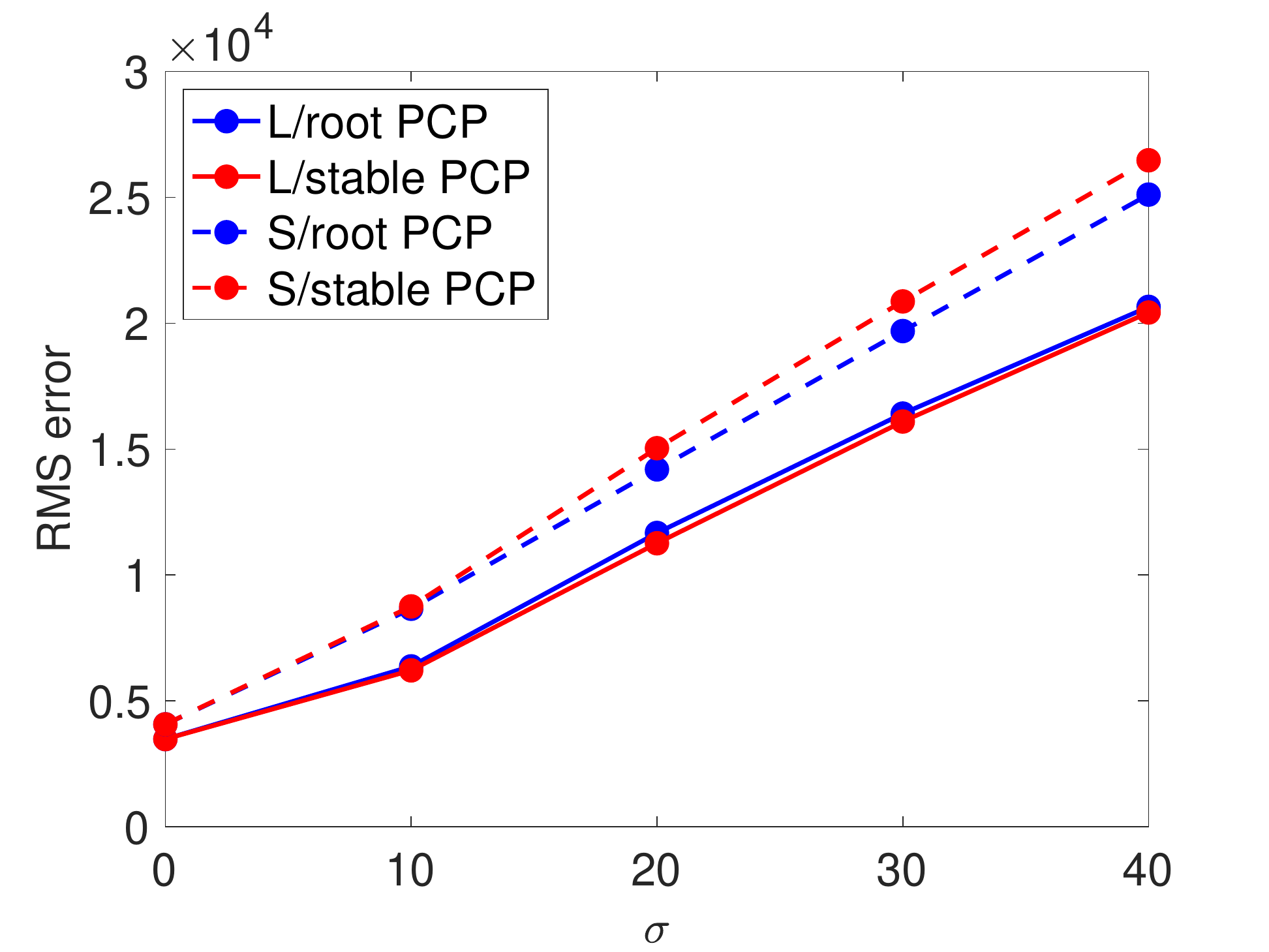}
\label{fig:yaleB02_error}}
\end{subfigure}
\hspace{-0.9em}
\begin{subfigure}[yaleB03 + noise]{
  \includegraphics[width = 0.19\textwidth]{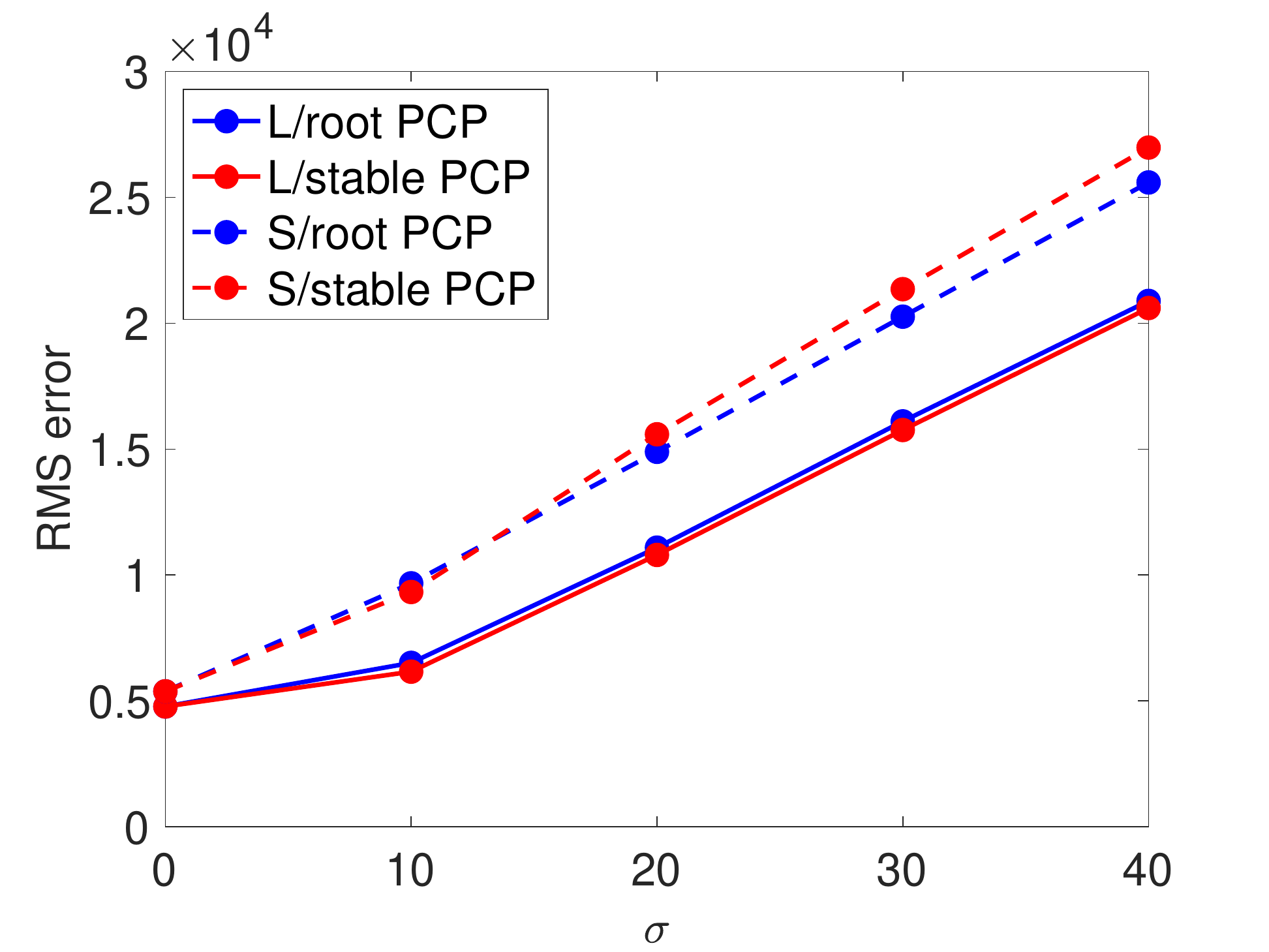}
\label{fig:yaleB03_error}}
\end{subfigure}
\caption{\stableU vs \rootpcp: real datasets}
\end{figure}

In Figure \ref{fig:lights_error}, we present the RMS error. Again, we see that the error is linear in the noise level $\sigma$.

\subsection{Additional Experiments Using Face Dataset}
In addition to the video dataset in the previous section, we also test \rootpcp and \stableU\ on datasets of face images. It has been pointed out in \cite{candes2011robust} that under distant illumination, images of a convex Lambertian object lie near a low dimensional linear subspace called the \textit{harmonic plane}. However, real images of faces are often corrupted by shadows and specularities, which can have large magnitudes but are sparse in the spatial domain. This fits well into our low-rank and sparse model \footnote{Although faces are not convex Lambertian objects and the harmonic plane may not apply here, previous experiments in \cite{candes2011robust} have shown the effectiveness of PCP in this task.}, and our goal is to remove these shadows and specularities from the noisy and corrupted observation. 

To be precise, the dataset we use is from Yale B face database \cite{yaleFace2001}. For each face, there are 65 images of resolution $192\times 168$ under various illuminations, so we have $n_1 = 192\times 168=32256$ and $n_2 = 65$. Similar to the experiments on the video dataset, we assume that there is no noise in these images, and add $\mb Z_0$ with $\sigma \in \{0,10,20,30,40\}$.

\begin{figure}[h!]
\begin{minipage}{0.5\textwidth}
\centering
\vspace{-2em}
\begin{subfigure}{
\makebox[20pt]{\raisebox{20pt}{\rotatebox[origin=c]{90}{}}}%
\makebox[0.18\textwidth]{face$ + \mb Z_0$}%
\makebox[0.18\textwidth]{$\wh{\mb L}_{\mathrm{root}}^{(\sigma)}$}%
\makebox[0.18\textwidth]{$\wh{\mb S}_{\mathrm{root}}^{(\sigma)}$}%
\makebox[0.18\textwidth]{$\wh{\mb L}_{\mathrm{stable}}^{(\sigma)}$}%
\makebox[0.18\textwidth]{$\wh{\mb S}_{\mathrm{stable}}^{(\sigma)}$}%
}
\end{subfigure}

\vspace{-0.5em}
\begin{subfigure}{
\makebox[20pt]{\raisebox{15pt}{\rotatebox[origin=c]{90}{$\sigma = 0$}}}%
\includegraphics[width = 0.17\textwidth]{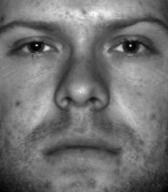}}
\end{subfigure}
\hspace{-0.9em}
\begin{subfigure}{
  \includegraphics[width = 0.17\textwidth]{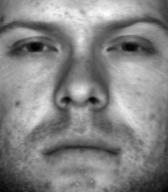}}
\end{subfigure}
\hspace{-0.9em}
\begin{subfigure}{
  \includegraphics[width = 0.17\textwidth]{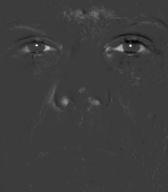}}
\end{subfigure}
\hspace{-0.9em}
\begin{subfigure}{
  \includegraphics[width = 0.17\textwidth]{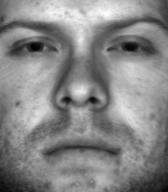}}
\end{subfigure}
\hspace{-0.9em}
\begin{subfigure}{
  \includegraphics[width = 0.17\textwidth]{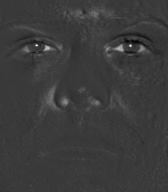}}
\end{subfigure}

\vspace{-0.9em}
\begin{subfigure}{
\makebox[20pt]{\raisebox{15pt}{\rotatebox[origin=c]{90}{$\sigma = 10$}}}%
\includegraphics[width = 0.17\textwidth]{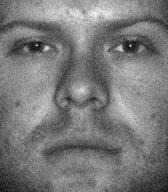}}
\end{subfigure}
\hspace{-0.9em}
\begin{subfigure}{
  \includegraphics[width = 0.17\textwidth]{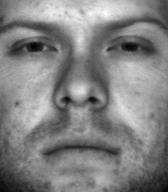}}
\end{subfigure}
\hspace{-0.9em}
\begin{subfigure}{
  \includegraphics[width = 0.17\textwidth]{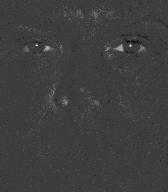}}
\end{subfigure}
\hspace{-0.9em}
\begin{subfigure}{
  \includegraphics[width = 0.17\textwidth]{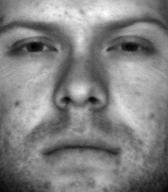}}
\end{subfigure}
\hspace{-0.9em}
\begin{subfigure}{
  \includegraphics[width = 0.17\textwidth]{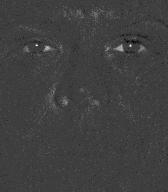}}
\end{subfigure}

\vspace{-0.9em}
\begin{subfigure}{
\makebox[20pt]{\raisebox{15pt}{\rotatebox[origin=c]{90}{$\sigma = 20$}}}%
\includegraphics[width = 0.17\textwidth]{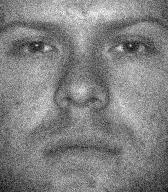}}
\end{subfigure}
\hspace{-0.9em}
\begin{subfigure}{
  \includegraphics[width = 0.17\textwidth]{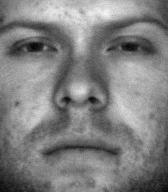}}
\end{subfigure}
\hspace{-0.9em}
\begin{subfigure}{
  \includegraphics[width = 0.17\textwidth]{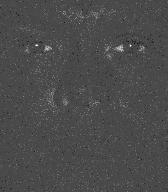}}
\end{subfigure}
\hspace{-0.9em}
\begin{subfigure}{
  \includegraphics[width = 0.17\textwidth]{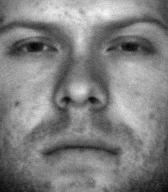}}
\end{subfigure}
\hspace{-0.9em}
\begin{subfigure}{
  \includegraphics[width = 0.17\textwidth]{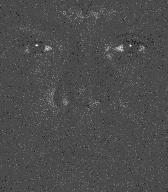}}
\end{subfigure}

\vspace{-0.9em}
\begin{subfigure}{
\makebox[20pt]{\raisebox{15pt}{\rotatebox[origin=c]{90}{$\sigma = 30$}}}%
\includegraphics[width = 0.17\textwidth]{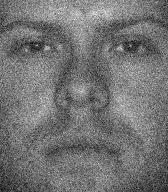}}
\end{subfigure}
\hspace{-0.9em}
\begin{subfigure}{
  \includegraphics[width = 0.17\textwidth]{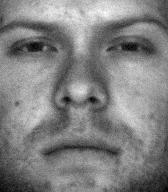}}
\end{subfigure}
\hspace{-0.9em}
\begin{subfigure}{
  \includegraphics[width = 0.17\textwidth]{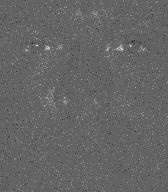}}
\end{subfigure}
\hspace{-0.9em}
\begin{subfigure}{
  \includegraphics[width = 0.17\textwidth]{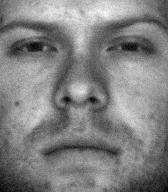}}
\end{subfigure}
\hspace{-0.9em}
\begin{subfigure}{
  \includegraphics[width = 0.17\textwidth]{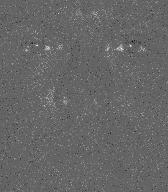}}
\end{subfigure}

\vspace{-0.9em}
\begin{subfigure}{
\makebox[20pt]{\raisebox{15pt}{\rotatebox[origin=c]{90}{$\sigma = 40$}}}%
\includegraphics[width = 0.17\textwidth]{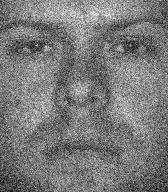}}
\end{subfigure}
\hspace{-0.9em}
\begin{subfigure}{
  \includegraphics[width = 0.17\textwidth]{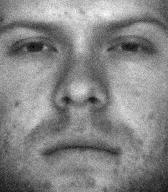}}
\end{subfigure}
\hspace{-0.9em}
\begin{subfigure}{
  \includegraphics[width = 0.17\textwidth]{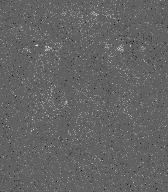}}
\end{subfigure}
\hspace{-0.9em}
\begin{subfigure}{
  \includegraphics[width = 0.17\textwidth]{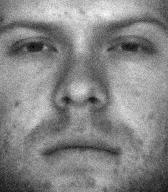}}
\end{subfigure}
\hspace{-0.9em}
\begin{subfigure}{
  \includegraphics[width = 0.17\textwidth]{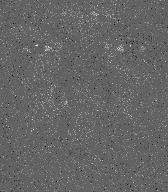}}
\end{subfigure}
\vspace{-1em}
\caption{yaleB01: recovered $\wh{\mb L},\wh{\mb S}$ for frame 1}
\vspace{-1em}
\label{fig:yaleB01_frame1_all}
\end{minipage}
\quad
\begin{minipage}{0.5\textwidth}
\centering
\vspace{-2em}
\begin{subfigure}{
\makebox[20pt]{\raisebox{20pt}{\rotatebox[origin=c]{90}{}}}%
\makebox[0.18\textwidth]{face$ + \mb Z_0$}%
\makebox[0.18\textwidth]{$\wh{\mb L}_{\mathrm{root}}^{(\sigma)}$}%
\makebox[0.18\textwidth]{$\wh{\mb S}_{\mathrm{root}}^{(\sigma)}$}%
\makebox[0.18\textwidth]{$\wh{\mb L}_{\mathrm{stable}}^{(\sigma)}$}%
\makebox[0.18\textwidth]{$\wh{\mb S}_{\mathrm{stable}}^{(\sigma)}$}%
}
\end{subfigure}

\vspace{-0.9em}
\begin{subfigure}{
\makebox[20pt]{\raisebox{15pt}{\rotatebox[origin=c]{90}{$\sigma = 0$}}}%
\includegraphics[width = 0.17\textwidth]{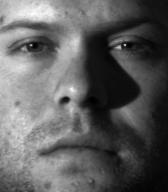}}
\end{subfigure}
\hspace{-0.9em}
\begin{subfigure}{
  \includegraphics[width = 0.17\textwidth]{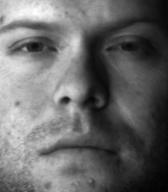}}
\end{subfigure}
\hspace{-0.9em}
\begin{subfigure}{
  \includegraphics[width = 0.17\textwidth]{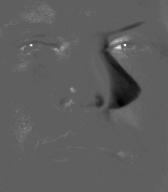}}
\end{subfigure}
\hspace{-0.9em}
\begin{subfigure}{
  \includegraphics[width = 0.17\textwidth]{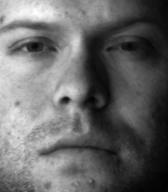}}
\end{subfigure}
\hspace{-0.9em}
\begin{subfigure}{
  \includegraphics[width = 0.17\textwidth]{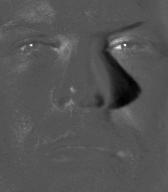}}
\end{subfigure}

\vspace{-0.9em}
\begin{subfigure}{
\makebox[20pt]{\raisebox{15pt}{\rotatebox[origin=c]{90}{$\sigma = 10$}}}%
\includegraphics[width = 0.17\textwidth]{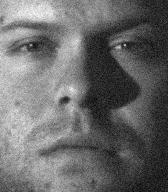}}
\end{subfigure}
\hspace{-0.9em}
\begin{subfigure}{
  \includegraphics[width = 0.17\textwidth]{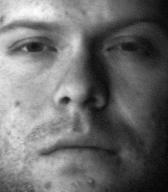}}
\end{subfigure}
\hspace{-0.9em}
\begin{subfigure}{
  \includegraphics[width = 0.17\textwidth]{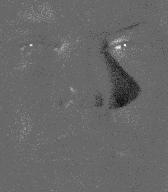}}
\end{subfigure}
\hspace{-0.9em}
\begin{subfigure}{
  \includegraphics[width = 0.17\textwidth]{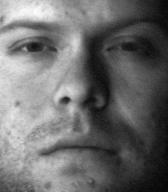}}
\end{subfigure}
\hspace{-0.9em}
\begin{subfigure}{
  \includegraphics[width = 0.17\textwidth]{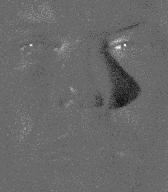}}
\end{subfigure}

\vspace{-0.9em}
\begin{subfigure}{
\makebox[20pt]{\raisebox{15pt}{\rotatebox[origin=c]{90}{$\sigma = 20$}}}%
\includegraphics[width = 0.17\textwidth]{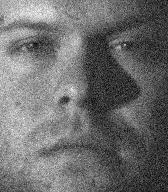}}
\end{subfigure}
\hspace{-0.9em}
\begin{subfigure}{
  \includegraphics[width = 0.17\textwidth]{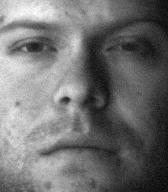}}
\end{subfigure}
\hspace{-0.9em}
\begin{subfigure}{
  \includegraphics[width = 0.17\textwidth]{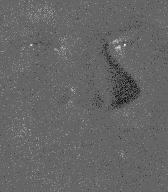}}
\end{subfigure}
\hspace{-0.9em}
\begin{subfigure}{
  \includegraphics[width = 0.17\textwidth]{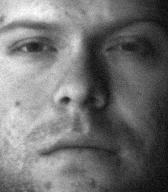}}
\end{subfigure}
\hspace{-0.9em}
\begin{subfigure}{
  \includegraphics[width = 0.17\textwidth]{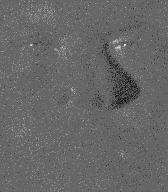}}
\end{subfigure}

\vspace{-0.9em}
\begin{subfigure}{
\makebox[20pt]{\raisebox{15pt}{\rotatebox[origin=c]{90}{$\sigma = 30$}}}%
\includegraphics[width = 0.17\textwidth]{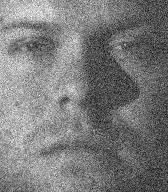}}
\end{subfigure}
\hspace{-0.9em}
\begin{subfigure}{
  \includegraphics[width = 0.17\textwidth]{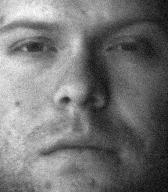}}
\end{subfigure}
\hspace{-0.9em}
\begin{subfigure}{
  \includegraphics[width = 0.17\textwidth]{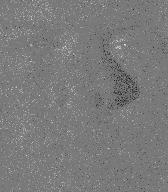}}
\end{subfigure}
\hspace{-0.9em}
\begin{subfigure}{
  \includegraphics[width = 0.17\textwidth]{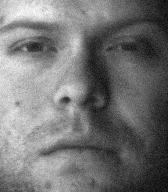}}
\end{subfigure}
\hspace{-0.9em}
\begin{subfigure}{
  \includegraphics[width = 0.17\textwidth]{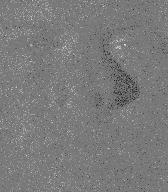}}
\end{subfigure}

\vspace{-0.9em}
\begin{subfigure}{
\makebox[20pt]{\raisebox{15pt}{\rotatebox[origin=c]{90}{$\sigma = 40$}}}%
\includegraphics[width = 0.17\textwidth]{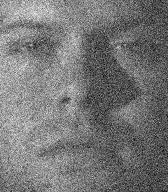}}
\end{subfigure}
\hspace{-0.9em}
\begin{subfigure}{
  \includegraphics[width = 0.17\textwidth]{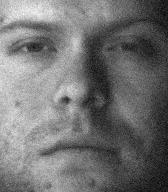}}
\end{subfigure}
\hspace{-0.9em}
\begin{subfigure}{
  \includegraphics[width = 0.17\textwidth]{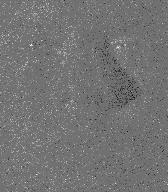}}
\end{subfigure}
\hspace{-0.9em}
\begin{subfigure}{
  \includegraphics[width = 0.17\textwidth]{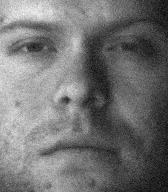}}
\end{subfigure}
\hspace{-0.9em}
\begin{subfigure}{
  \includegraphics[width = 0.17\textwidth]{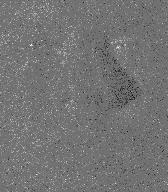}}
\end{subfigure}
\vspace{-1em}
\caption{yaleB01: recovered $\wh{\mb L},\wh{\mb S}$ for frame 20}
\vspace{-1em}
\label{fig:yaleB01_frame20_all}
\end{minipage}
\end{figure}

\begin{figure}[h!]
\begin{minipage}{0.5\textwidth}
\centering
\vspace{-2em}
\begin{subfigure}{
\makebox[20pt]{\raisebox{20pt}{\rotatebox[origin=c]{90}{}}}%
\makebox[0.18\textwidth]{face$ + \mb Z_0$}%
\makebox[0.18\textwidth]{$\wh{\mb L}_{\mathrm{root}}^{(\sigma)}$}%
\makebox[0.18\textwidth]{$\wh{\mb S}_{\mathrm{root}}^{(\sigma)}$}%
\makebox[0.18\textwidth]{$\wh{\mb L}_{\mathrm{stable}}^{(\sigma)}$}%
\makebox[0.18\textwidth]{$\wh{\mb S}_{\mathrm{stable}}^{(\sigma)}$}%
}
\end{subfigure}

\vspace{-0.5em}
\begin{subfigure}{
\makebox[20pt]{\raisebox{15pt}{\rotatebox[origin=c]{90}{$\sigma = 0$}}}%
\includegraphics[width = 0.17\textwidth]{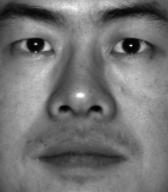}}
\end{subfigure}
\hspace{-0.9em}
\begin{subfigure}{
  \includegraphics[width = 0.17\textwidth]{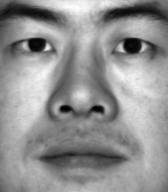}}
\end{subfigure}
\hspace{-0.9em}
\begin{subfigure}{
  \includegraphics[width = 0.17\textwidth]{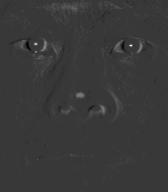}}
\end{subfigure}
\hspace{-0.9em}
\begin{subfigure}{
  \includegraphics[width = 0.17\textwidth]{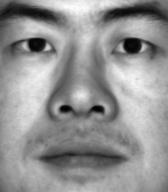}}
\end{subfigure}
\hspace{-0.9em}
\begin{subfigure}{
  \includegraphics[width = 0.17\textwidth]{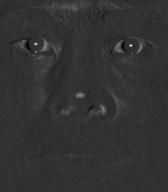}}
\end{subfigure}

\vspace{-0.9em}
\begin{subfigure}{
\makebox[20pt]{\raisebox{15pt}{\rotatebox[origin=c]{90}{$\sigma = 10$}}}%
\includegraphics[width = 0.17\textwidth]{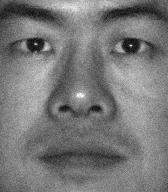}}
\end{subfigure}
\hspace{-0.9em}
\begin{subfigure}{
  \includegraphics[width = 0.17\textwidth]{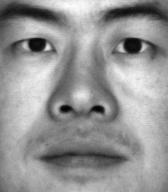}}
\end{subfigure}
\hspace{-0.9em}
\begin{subfigure}{
  \includegraphics[width = 0.17\textwidth]{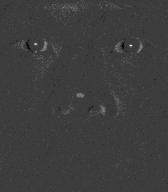}}
\end{subfigure}
\hspace{-0.9em}
\begin{subfigure}{
  \includegraphics[width = 0.17\textwidth]{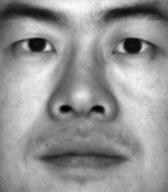}}
\end{subfigure}
\hspace{-0.9em}
\begin{subfigure}{
  \includegraphics[width = 0.17\textwidth]{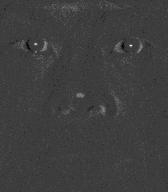}}
\end{subfigure}

\vspace{-0.9em}
\begin{subfigure}{
\makebox[20pt]{\raisebox{15pt}{\rotatebox[origin=c]{90}{$\sigma = 20$}}}%
\includegraphics[width = 0.17\textwidth]{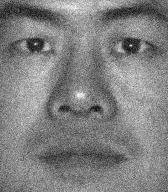}}
\end{subfigure}
\hspace{-0.9em}
\begin{subfigure}{
  \includegraphics[width = 0.17\textwidth]{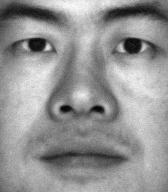}}
\end{subfigure}
\hspace{-0.9em}
\begin{subfigure}{
  \includegraphics[width = 0.17\textwidth]{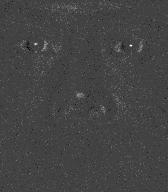}}
\end{subfigure}
\hspace{-0.9em}
\begin{subfigure}{
  \includegraphics[width = 0.17\textwidth]{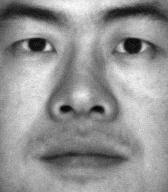}}
\end{subfigure}
\hspace{-0.9em}
\begin{subfigure}{
  \includegraphics[width = 0.17\textwidth]{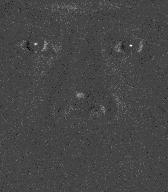}}
\end{subfigure}

\vspace{-0.9em}
\begin{subfigure}{
\makebox[20pt]{\raisebox{15pt}{\rotatebox[origin=c]{90}{$\sigma = 30$}}}%
\includegraphics[width = 0.17\textwidth]{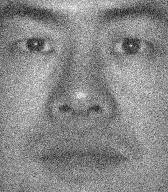}}
\end{subfigure}
\hspace{-0.9em}
\begin{subfigure}{
  \includegraphics[width = 0.17\textwidth]{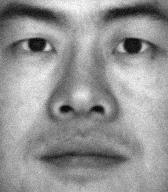}}
\end{subfigure}
\hspace{-0.9em}
\begin{subfigure}{
  \includegraphics[width = 0.17\textwidth]{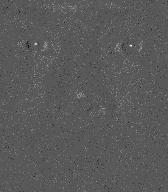}}
\end{subfigure}
\hspace{-0.9em}
\begin{subfigure}{
  \includegraphics[width = 0.17\textwidth]{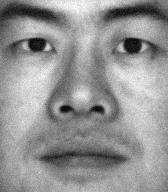}}
\end{subfigure}
\hspace{-0.9em}
\begin{subfigure}{
  \includegraphics[width = 0.17\textwidth]{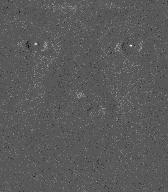}}
\end{subfigure}

\vspace{-0.9em}
\begin{subfigure}{
\makebox[20pt]{\raisebox{15pt}{\rotatebox[origin=c]{90}{$\sigma = 40$}}}%
\includegraphics[width = 0.17\textwidth]{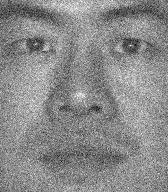}}
\end{subfigure}
\hspace{-0.9em}
\begin{subfigure}{
  \includegraphics[width = 0.17\textwidth]{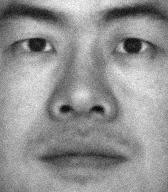}}
\end{subfigure}
\hspace{-0.9em}
\begin{subfigure}{
  \includegraphics[width = 0.17\textwidth]{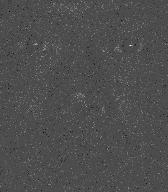}}
\end{subfigure}
\hspace{-0.9em}
\begin{subfigure}{
  \includegraphics[width = 0.17\textwidth]{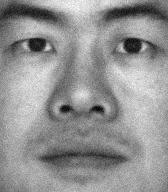}}
\end{subfigure}
\hspace{-0.9em}
\begin{subfigure}{
  \includegraphics[width = 0.17\textwidth]{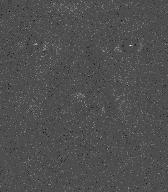}}
\end{subfigure}
\vspace{-1em}
\caption{yaleB02: recovered $\wh{\mb L},\wh{\mb S}$ for frame 1}
\vspace{-1em}
\label{fig:yaleB02_frame1_all}
\end{minipage}
\quad
\begin{minipage}{0.5\textwidth}
\centering
\vspace{-2em}
\begin{subfigure}{
\makebox[20pt]{\raisebox{20pt}{\rotatebox[origin=c]{90}{}}}%
\makebox[0.18\textwidth]{face$ + \mb Z_0$}%
\makebox[0.18\textwidth]{$\wh{\mb L}_{\mathrm{root}}^{(\sigma)}$}%
\makebox[0.18\textwidth]{$\wh{\mb S}_{\mathrm{root}}^{(\sigma)}$}%
\makebox[0.18\textwidth]{$\wh{\mb L}_{\mathrm{stable}}^{(\sigma)}$}%
\makebox[0.18\textwidth]{$\wh{\mb S}_{\mathrm{stable}}^{(\sigma)}$}%
}
\end{subfigure}

\vspace{-0.9em}
\begin{subfigure}{
\makebox[20pt]{\raisebox{15pt}{\rotatebox[origin=c]{90}{$\sigma = 0$}}}%
\includegraphics[width = 0.17\textwidth]{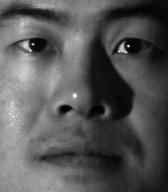}}
\end{subfigure}
\hspace{-0.9em}
\begin{subfigure}{
  \includegraphics[width = 0.17\textwidth]{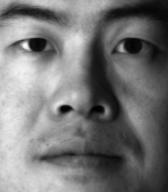}}
\end{subfigure}
\hspace{-0.9em}
\begin{subfigure}{
  \includegraphics[width = 0.17\textwidth]{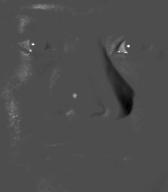}}
\end{subfigure}
\hspace{-0.9em}
\begin{subfigure}{
  \includegraphics[width = 0.17\textwidth]{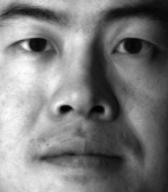}}
\end{subfigure}
\hspace{-0.9em}
\begin{subfigure}{
  \includegraphics[width = 0.17\textwidth]{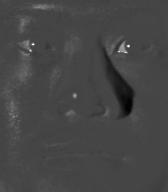}}
\end{subfigure}

\vspace{-0.9em}
\begin{subfigure}{
\makebox[20pt]{\raisebox{15pt}{\rotatebox[origin=c]{90}{$\sigma = 10$}}}%
\includegraphics[width = 0.17\textwidth]{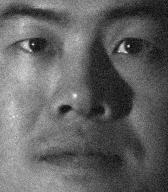}}
\end{subfigure}
\hspace{-0.9em}
\begin{subfigure}{
  \includegraphics[width = 0.17\textwidth]{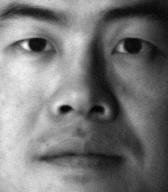}}
\end{subfigure}
\hspace{-0.9em}
\begin{subfigure}{
  \includegraphics[width = 0.17\textwidth]{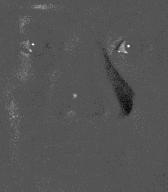}}
\end{subfigure}
\hspace{-0.9em}
\begin{subfigure}{
  \includegraphics[width = 0.17\textwidth]{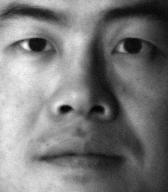}}
\end{subfigure}
\hspace{-0.9em}
\begin{subfigure}{
  \includegraphics[width = 0.17\textwidth]{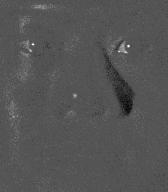}}
\end{subfigure}

\vspace{-0.9em}
\begin{subfigure}{
\makebox[20pt]{\raisebox{15pt}{\rotatebox[origin=c]{90}{$\sigma = 20$}}}%
\includegraphics[width = 0.17\textwidth]{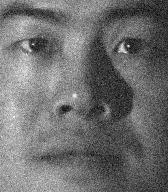}}
\end{subfigure}
\hspace{-0.9em}
\begin{subfigure}{
  \includegraphics[width = 0.17\textwidth]{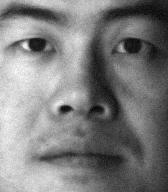}}
\end{subfigure}
\hspace{-0.9em}
\begin{subfigure}{
  \includegraphics[width = 0.17\textwidth]{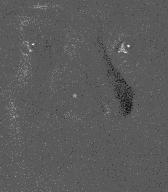}}
\end{subfigure}
\hspace{-0.9em}
\begin{subfigure}{
  \includegraphics[width = 0.17\textwidth]{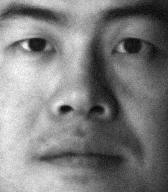}}
\end{subfigure}
\hspace{-0.9em}
\begin{subfigure}{
  \includegraphics[width = 0.17\textwidth]{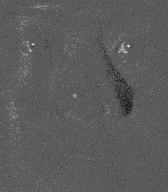}}
\end{subfigure}

\vspace{-0.9em}
\begin{subfigure}{
\makebox[20pt]{\raisebox{15pt}{\rotatebox[origin=c]{90}{$\sigma = 30$}}}%
\includegraphics[width = 0.17\textwidth]{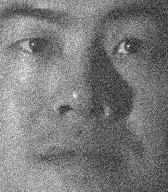}}
\end{subfigure}
\hspace{-0.9em}
\begin{subfigure}{
  \includegraphics[width = 0.17\textwidth]{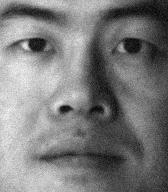}}
\end{subfigure}
\hspace{-0.9em}
\begin{subfigure}{
  \includegraphics[width = 0.17\textwidth]{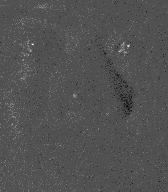}}
\end{subfigure}
\hspace{-0.9em}
\begin{subfigure}{
  \includegraphics[width = 0.17\textwidth]{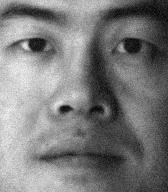}}
\end{subfigure}
\hspace{-0.9em}
\begin{subfigure}{
  \includegraphics[width = 0.17\textwidth]{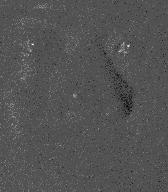}}
\end{subfigure}

\vspace{-0.9em}
\begin{subfigure}{
\makebox[20pt]{\raisebox{15pt}{\rotatebox[origin=c]{90}{$\sigma = 40$}}}%
\includegraphics[width = 0.17\textwidth]{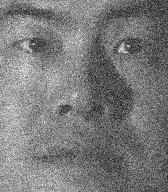}}
\end{subfigure}
\hspace{-0.9em}
\begin{subfigure}{
  \includegraphics[width = 0.17\textwidth]{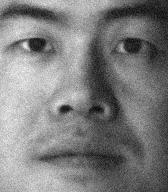}}
\end{subfigure}
\hspace{-0.9em}
\begin{subfigure}{
  \includegraphics[width = 0.17\textwidth]{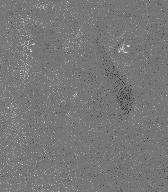}}
\end{subfigure}
\hspace{-0.9em}
\begin{subfigure}{
  \includegraphics[width = 0.17\textwidth]{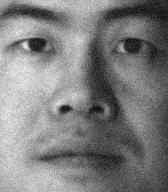}}
\end{subfigure}
\hspace{-0.9em}
\begin{subfigure}{
  \includegraphics[width = 0.17\textwidth]{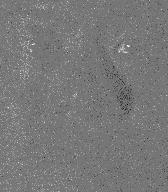}}
\end{subfigure}
\vspace{-1em}
\caption{yaleB02: recovered $\wh{\mb L},\wh{\mb S}$ for frame 20}
\vspace{-1em}
\label{fig:yaleB02_frame20_all}
\end{minipage}
\end{figure}

\begin{figure}[h!]
\begin{minipage}{0.5\textwidth}
\centering
\vspace{-2em}
\begin{subfigure}{
\makebox[20pt]{\raisebox{20pt}{\rotatebox[origin=c]{90}{}}}%
\makebox[0.18\textwidth]{face$ + \mb Z_0$}%
\makebox[0.18\textwidth]{$\wh{\mb L}_{\mathrm{root}}^{(\sigma)}$}%
\makebox[0.18\textwidth]{$\wh{\mb S}_{\mathrm{root}}^{(\sigma)}$}%
\makebox[0.18\textwidth]{$\wh{\mb L}_{\mathrm{stable}}^{(\sigma)}$}%
\makebox[0.18\textwidth]{$\wh{\mb S}_{\mathrm{stable}}^{(\sigma)}$}%
}
\end{subfigure}

\vspace{-0.5em}
\begin{subfigure}{
\makebox[20pt]{\raisebox{15pt}{\rotatebox[origin=c]{90}{$\sigma = 0$}}}%
\includegraphics[width = 0.17\textwidth]{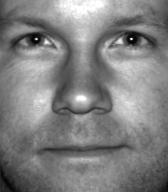}}
\end{subfigure}
\hspace{-0.9em}
\begin{subfigure}{
  \includegraphics[width = 0.17\textwidth]{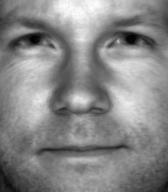}}
\end{subfigure}
\hspace{-0.9em}
\begin{subfigure}{
  \includegraphics[width = 0.17\textwidth]{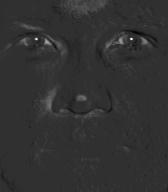}}
\end{subfigure}
\hspace{-0.9em}
\begin{subfigure}{
  \includegraphics[width = 0.17\textwidth]{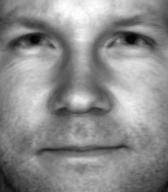}}
\end{subfigure}
\hspace{-0.9em}
\begin{subfigure}{
  \includegraphics[width = 0.17\textwidth]{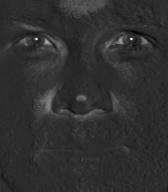}}
\end{subfigure}

\vspace{-0.9em}
\begin{subfigure}{
\makebox[20pt]{\raisebox{15pt}{\rotatebox[origin=c]{90}{$\sigma = 10$}}}%
\includegraphics[width = 0.17\textwidth]{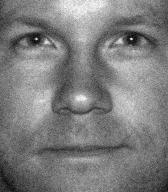}}
\end{subfigure}
\hspace{-0.9em}
\begin{subfigure}{
  \includegraphics[width = 0.17\textwidth]{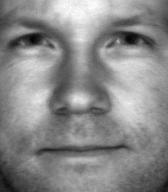}}
\end{subfigure}
\hspace{-0.9em}
\begin{subfigure}{
  \includegraphics[width = 0.17\textwidth]{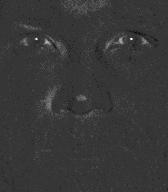}}
\end{subfigure}
\hspace{-0.9em}
\begin{subfigure}{
  \includegraphics[width = 0.17\textwidth]{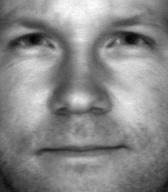}}
\end{subfigure}
\hspace{-0.9em}
\begin{subfigure}{
  \includegraphics[width = 0.17\textwidth]{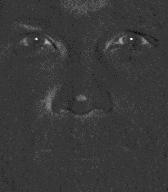}}
\end{subfigure}

\vspace{-0.9em}
\begin{subfigure}{
\makebox[20pt]{\raisebox{15pt}{\rotatebox[origin=c]{90}{$\sigma = 20$}}}%
\includegraphics[width = 0.17\textwidth]{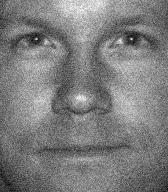}}
\end{subfigure}
\hspace{-0.9em}
\begin{subfigure}{
  \includegraphics[width = 0.17\textwidth]{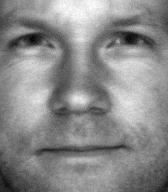}}
\end{subfigure}
\hspace{-0.9em}
\begin{subfigure}{
  \includegraphics[width = 0.17\textwidth]{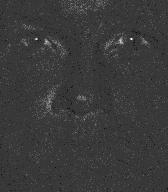}}
\end{subfigure}
\hspace{-0.9em}
\begin{subfigure}{
  \includegraphics[width = 0.17\textwidth]{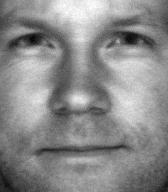}}
\end{subfigure}
\hspace{-0.9em}
\begin{subfigure}{
  \includegraphics[width = 0.17\textwidth]{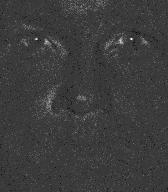}}
\end{subfigure}

\vspace{-0.9em}
\begin{subfigure}{
\makebox[20pt]{\raisebox{15pt}{\rotatebox[origin=c]{90}{$\sigma = 30$}}}%
\includegraphics[width = 0.17\textwidth]{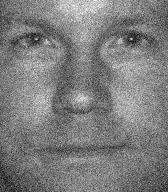}}
\end{subfigure}
\hspace{-0.9em}
\begin{subfigure}{
  \includegraphics[width = 0.17\textwidth]{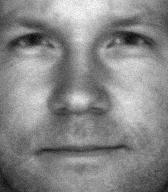}}
\end{subfigure}
\hspace{-0.9em}
\begin{subfigure}{
  \includegraphics[width = 0.17\textwidth]{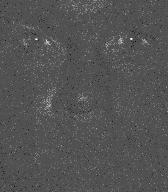}}
\end{subfigure}
\hspace{-0.9em}
\begin{subfigure}{
  \includegraphics[width = 0.17\textwidth]{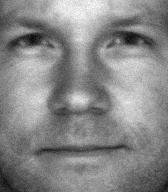}}
\end{subfigure}
\hspace{-0.9em}
\begin{subfigure}{
  \includegraphics[width = 0.17\textwidth]{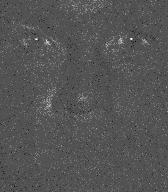}}
\end{subfigure}

\vspace{-0.9em}
\begin{subfigure}{
\makebox[20pt]{\raisebox{15pt}{\rotatebox[origin=c]{90}{$\sigma = 40$}}}%
\includegraphics[width = 0.17\textwidth]{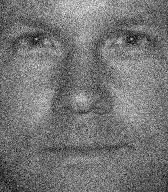}}
\end{subfigure}
\hspace{-0.9em}
\begin{subfigure}{
  \includegraphics[width = 0.17\textwidth]{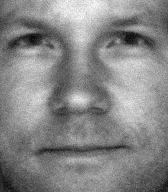}}
\end{subfigure}
\hspace{-0.9em}
\begin{subfigure}{
  \includegraphics[width = 0.17\textwidth]{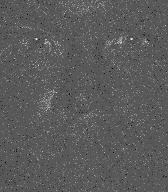}}
\end{subfigure}
\hspace{-0.9em}
\begin{subfigure}{
  \includegraphics[width = 0.17\textwidth]{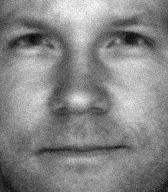}}
\end{subfigure}
\hspace{-0.9em}
\begin{subfigure}{
  \includegraphics[width = 0.17\textwidth]{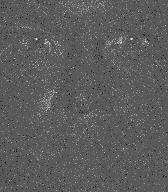}}
\end{subfigure}
\vspace{-1em}
\caption{yaleB03: recovered $\wh{\mb L},\wh{\mb S}$ for frame 1}
\vspace{-1em}
\label{fig:yaleB03_frame1_all}
\end{minipage}
\quad
\begin{minipage}{0.5\textwidth}
\centering
\vspace{-2em}
\begin{subfigure}{
\makebox[20pt]{\raisebox{20pt}{\rotatebox[origin=c]{90}{}}}%
\makebox[0.18\textwidth]{face$ + \mb Z_0$}%
\makebox[0.18\textwidth]{$\wh{\mb L}_{\mathrm{root}}^{(\sigma)}$}%
\makebox[0.18\textwidth]{$\wh{\mb S}_{\mathrm{root}}^{(\sigma)}$}%
\makebox[0.18\textwidth]{$\wh{\mb L}_{\mathrm{stable}}^{(\sigma)}$}%
\makebox[0.18\textwidth]{$\wh{\mb S}_{\mathrm{stable}}^{(\sigma)}$}%
}
\end{subfigure}

\vspace{-0.9em}
\begin{subfigure}{
\makebox[20pt]{\raisebox{15pt}{\rotatebox[origin=c]{90}{$\sigma = 0$}}}%
\includegraphics[width = 0.17\textwidth]{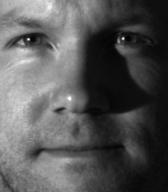}}
\end{subfigure}
\hspace{-0.9em}
\begin{subfigure}{
  \includegraphics[width = 0.17\textwidth]{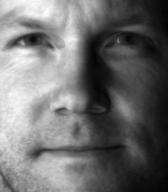}}
\end{subfigure}
\hspace{-0.9em}
\begin{subfigure}{
  \includegraphics[width = 0.17\textwidth]{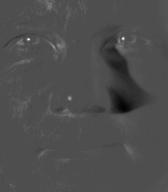}}
\end{subfigure}
\hspace{-0.9em}
\begin{subfigure}{
  \includegraphics[width = 0.17\textwidth]{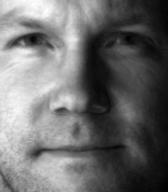}}
\end{subfigure}
\hspace{-0.9em}
\begin{subfigure}{
  \includegraphics[width = 0.17\textwidth]{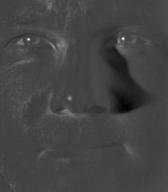}}
\end{subfigure}

\vspace{-0.9em}
\begin{subfigure}{
\makebox[20pt]{\raisebox{15pt}{\rotatebox[origin=c]{90}{$\sigma = 10$}}}%
\includegraphics[width = 0.17\textwidth]{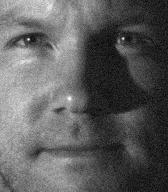}}
\end{subfigure}
\hspace{-0.9em}
\begin{subfigure}{
  \includegraphics[width = 0.17\textwidth]{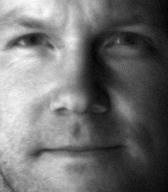}}
\end{subfigure}
\hspace{-0.9em}
\begin{subfigure}{
  \includegraphics[width = 0.17\textwidth]{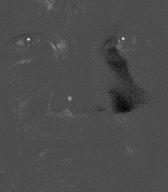}}
\end{subfigure}
\hspace{-0.9em}
\begin{subfigure}{
  \includegraphics[width = 0.17\textwidth]{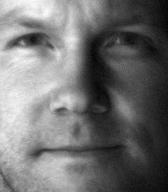}}
\end{subfigure}
\hspace{-0.9em}
\begin{subfigure}{
  \includegraphics[width = 0.17\textwidth]{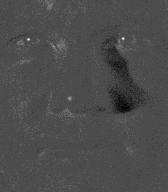}}
\end{subfigure}

\vspace{-0.9em}
\begin{subfigure}{
\makebox[20pt]{\raisebox{15pt}{\rotatebox[origin=c]{90}{$\sigma = 20$}}}%
\includegraphics[width = 0.17\textwidth]{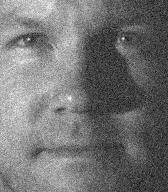}}
\end{subfigure}
\hspace{-0.9em}
\begin{subfigure}{
  \includegraphics[width = 0.17\textwidth]{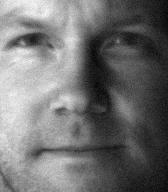}}
\end{subfigure}
\hspace{-0.9em}
\begin{subfigure}{
  \includegraphics[width = 0.17\textwidth]{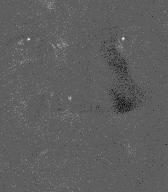}}
\end{subfigure}
\hspace{-0.9em}
\begin{subfigure}{
  \includegraphics[width = 0.17\textwidth]{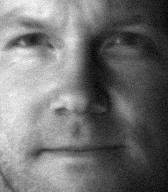}}
\end{subfigure}
\hspace{-0.9em}
\begin{subfigure}{
  \includegraphics[width = 0.17\textwidth]{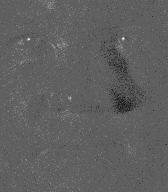}}
\end{subfigure}

\vspace{-0.9em}
\begin{subfigure}{
\makebox[20pt]{\raisebox{15pt}{\rotatebox[origin=c]{90}{$\sigma = 30$}}}%
\includegraphics[width = 0.17\textwidth]{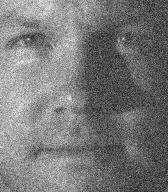}}
\end{subfigure}
\hspace{-0.9em}
\begin{subfigure}{
  \includegraphics[width = 0.17\textwidth]{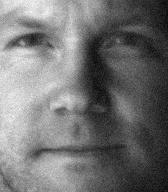}}
\end{subfigure}
\hspace{-0.9em}
\begin{subfigure}{
  \includegraphics[width = 0.17\textwidth]{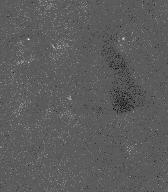}}
\end{subfigure}
\hspace{-0.9em}
\begin{subfigure}{
  \includegraphics[width = 0.17\textwidth]{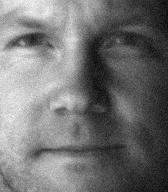}}
\end{subfigure}
\hspace{-0.9em}
\begin{subfigure}{
  \includegraphics[width = 0.17\textwidth]{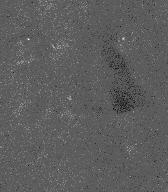}}
\end{subfigure}

\vspace{-0.9em}
\begin{subfigure}{
\makebox[20pt]{\raisebox{15pt}{\rotatebox[origin=c]{90}{$\sigma = 40$}}}%
\includegraphics[width = 0.17\textwidth]{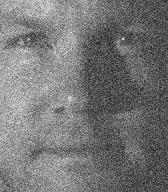}}
\end{subfigure}
\hspace{-0.9em}
\begin{subfigure}{
  \includegraphics[width = 0.17\textwidth]{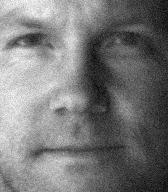}}
\end{subfigure}
\hspace{-0.9em}
\begin{subfigure}{
  \includegraphics[width = 0.17\textwidth]{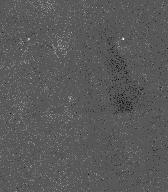}}
\end{subfigure}
\hspace{-0.9em}
\begin{subfigure}{
  \includegraphics[width = 0.17\textwidth]{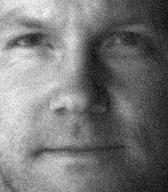}}
\end{subfigure}
\hspace{-0.9em}
\begin{subfigure}{
  \includegraphics[width = 0.17\textwidth]{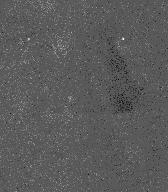}}
\end{subfigure}
\vspace{-1em}
\caption{yaleB03: recovered $\wh{\mb L},\wh{\mb S}$ for frame 20}
\vspace{-1em}
\label{fig:yaleB03_frame20_all}
\end{minipage}
\end{figure}

In Figures \ref{fig:yaleB01_frame1_all}, \ref{fig:yaleB01_frame20_all}, \ref{fig:yaleB02_frame1_all}, \ref{fig:yaleB02_frame20_all}, \ref{fig:yaleB03_frame1_all}, and \ref{fig:yaleB03_frame20_all}, we present the recovered low rank and sparse matrices for frame 1 and frame 20 using \rootpcp and \stableU.

In Table \ref{table:yale_root1}, \ref{table:yale_root2}, and \ref{table:yale_root3}, we show the relative error, running time and iteration for \rootpcp on yaleB01, yaleB02, and yaleB03 datasets. In Table \ref{table:yale_stable1}, \ref{table:yale_stable2}, and \ref{table:yale_stable3}, we show the relative error, running time and iteration for \stableU on yaleB01, yaleB02, and yaleB03 datasets. 
\begin{table}[h!]
\begin{minipage}{.5\linewidth}
\caption{\rootpcp: yaleB01}
\label{table:yale_root1}
\vskip 0.15in
\begin{center}
\begin{small}
\begin{sc}
\begin{tabular}{l||c|c|c|c}
$\sigma$ & $\frac{\|\wh{\mb L}-\mb L_0\|_F}{\|\mb L_0\|_F} $& $\frac{\|\wh{\mb S}-\mb S_0\|_F}{\|\mb S_0\|_F}$& Time (s)  & Iter\\
\hline
0     & 0.0298& 0.1934& 255.6449& 2054\\
10    & 0.0481& 0.3823& 171.0386& 1366\\ 
20    & 0.0863& 0.6117& 184.6412& 1480\\ 
30    & 0.1250& 0.8421& 210.4474& 1693\\ 
40    & 0.1613& 1.0692& 493.5861& 3953\\ 
\end{tabular}
\end{sc}
\end{small}
\end{center}
\vskip -0.1in
\end{minipage}
\quad
\begin{minipage}{.5\linewidth}
\caption{\stableU: yaleB01}
\label{table:yale_stable1}
\vskip 0.15in
\begin{center}
\begin{small}
\begin{sc}
\begin{tabular}{l||c|c|c|c}
$\sigma$ & $\frac{\|\wh{\mb L}-\mb L_0\|_F}{\|\mb L_0\|_F} $& $\frac{\|\wh{\mb S}-\mb S_0\|_F}{\|\mb S_0\|_F}$& Time (s)  & Iter\\
\hline
0     & 0.0298& 0.1934& 602.0781& 5000+\\ 
10    & 0.0461& 0.3808& 599.3534& 5000+\\ 
20    & 0.0838& 0.6441& 594.2583& 5000+\\ 
30    & 0.1225& 0.8898& 596.9568& 5000+\\ 
40    & 0.1591& 1.1276& 602.5028& 5000+\\ 
\end{tabular}
\end{sc}
\end{small}
\end{center}
\vskip -0.1in
\end{minipage}
\end{table}

\begin{table}[h!]
\begin{minipage}{.5\linewidth}
\caption{\rootpcp: yaleB02}
\label{table:yale_root2}
\vskip 0.15in
\begin{center}
\begin{small}
\begin{sc}
\begin{tabular}{l||c|c|c|c}
$\sigma$ & $\frac{\|\wh{\mb L}-\mb L_0\|_F}{\|\mb L_0\|_F} $& $\frac{\|\wh{\mb S}-\mb S_0\|_F}{\|\mb S_0\|_F}$& Time (s)  & Iter\\
\hline
0     & 0.0261& 0.1720& 272.5150& 2190\\                                         
10    & 0.0477& 0.3657& 167.7344& 1352\\ 
20    & 0.0873& 0.5993& 174.0215& 1409\\ 
30    & 0.1228& 0.8311& 407.7892& 3289\\ 
40    & 0.1545& 1.0599& 244.3114& 1970\\ 
\end{tabular}
\end{sc}
\end{small}
\end{center}
\vskip -0.1in
\end{minipage}
\quad
\begin{minipage}{.5\linewidth}
\caption{\stableU: yaleB02}
\label{table:yale_stable2}
\vskip 0.15in
\begin{center}
\begin{small}
\begin{sc}
\begin{tabular}{l||c|c|c|c}
$\sigma$ & $\frac{\|\wh{\mb L}-\mb L_0\|_F}{\|\mb L_0\|_F} $& $\frac{\|\wh{\mb S}-\mb S_0\|_F}{\|\mb S_0\|_F}$& Time (s)  & Iter\\
\hline
0     & 0.0261& 0.1720& 595.5254& 5000+\\                                     
10    & 0.0465& 0.3697& 595.0926& 5000+\\ 
20    & 0.0843& 0.6348& 594.2518& 5000+\\ 
30    & 0.1204& 0.8808& 599.8056& 5000+\\ 
40    & 0.1528& 1.1174& 593.7441& 5000+\\ 
\end{tabular}
\end{sc}
\end{small}
\end{center}
\vskip -0.1in
\end{minipage}
\end{table}

\begin{table}[h!]
\begin{minipage}{.5\linewidth}
\caption{\rootpcp: yaleB03}
\label{table:yale_root3}
\vskip 0.15in
\begin{center}
\begin{small}
\begin{sc}
\begin{tabular}{l||c|c|c|c}
$\sigma$ & $\frac{\|\wh{\mb L}-\mb L_0\|_F}{\|\mb L_0\|_F} $& $\frac{\|\wh{\mb S}-\mb S_0\|_F}{\|\mb S_0\|_F}$& Time (s)  & Iter\\
\hline
0     & 0.0336& 0.2128& 272.8555& 2194\\
10    & 0.0458& 0.3823& 175.4205& 1417\\ 
20    & 0.0780& 0.5887& 261.3707& 1519\\ 
30    & 0.1133& 0.8008& 221.5380& 1751\\ 
40    & 0.1469& 1.0114& 512.7160& 4051\\ 
\end{tabular}
\end{sc}
\end{small}
\end{center}
\vskip -0.1in
\end{minipage}
\quad
\begin{minipage}{.5\linewidth}
\caption{\stableU: yaleB03}
\label{table:yale_stable3}
\vskip 0.15in
\begin{center}
\begin{small}
\begin{sc}
\begin{tabular}{l||c|c|c|c}
$\sigma$ & $\frac{\|\wh{\mb L}-\mb L_0\|_F}{\|\mb L_0\|_F} $& $\frac{\|\wh{\mb S}-\mb S_0\|_F}{\|\mb S_0\|_F}$& Time (s)  & Iter\\
\hline
0     & 0.0336& 0.2128& 594.5691& 5000+\\                                         
10    & 0.0434& 0.3687& 594.4220& 5000+\\ 
20    & 0.0760& 0.6162& 596.9680& 5000+\\ 
30    & 0.1109& 0.8440& 839.4338& 5000+\\ 
40    & 0.1449& 1.0662& 601.5731& 5000+\\ 
\end{tabular}
\end{sc}
\end{small}
\end{center}
\vskip -0.1in
\end{minipage}
\end{table}

\subsection{Additional Results for Section \ref{section:low_light_video}}
\begin{figure}[h!]
\centering
\begin{subfigure}{
\makebox[20pt]{\raisebox{20pt}{\rotatebox[origin=c]{90}{}}}%
\makebox[0.16\textwidth]{$\mb D$}%
\makebox[0.16\textwidth]{$\wh{\mb L}+\wh{\mb S}$}%
\makebox[0.16\textwidth]{$\wh{\mb L}$}%
\makebox[0.16\textwidth]{$\wh{\mb S}$}%
\makebox[0.16\textwidth]{$\wh{\mb Z}$}%
}
\end{subfigure}
\vspace{-0.5em}
\begin{subfigure}{
\makebox[20pt]{\raisebox{20pt}{\rotatebox[origin=c]{90}{frame 30}}}%
\includegraphics[width = 0.15\textwidth]{plots/denoising_v01/frame_30_raw.jpeg}}
\end{subfigure}
\hspace{-0.5em}
\begin{subfigure}{
  \includegraphics[width = 0.15\textwidth]{plots/denoising_v01/frame_30_L_plus_S.jpeg}}
\end{subfigure}
\hspace{-0.5em}
\begin{subfigure}{
  \includegraphics[width = 0.15\textwidth]{plots/denoising_v01/frame_30_L.jpeg}}
\end{subfigure}
\hspace{-0.5em}
\begin{subfigure}{
  \includegraphics[width = 0.15\textwidth]{plots/denoising_v01/frame_30_S.jpeg}}
\end{subfigure}
\hspace{-0.5em}
\begin{subfigure}{
  \includegraphics[width = 0.15\textwidth]{plots/denoising_v01/frame_30_noise.jpeg}}
\end{subfigure}
\vspace{-0.5em}
\begin{subfigure}{
\makebox[20pt]{\raisebox{20pt}{\rotatebox[origin=c]{90}{frame 60}}}%
  \includegraphics[width = 0.15\textwidth]{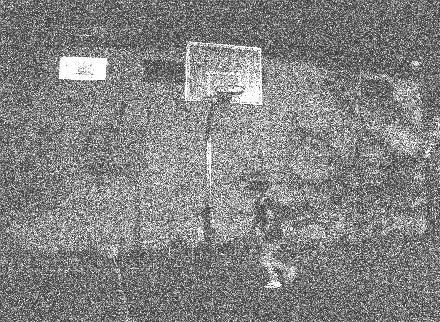}}
\end{subfigure}
\hspace{-0.5em}
\begin{subfigure}{
  \includegraphics[width = 0.15\textwidth]{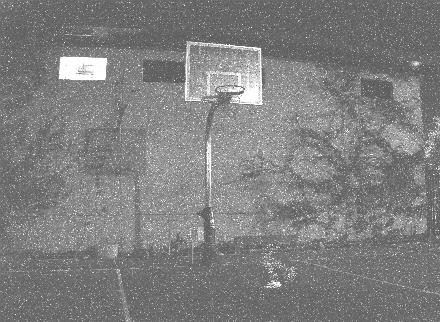}}
\end{subfigure}
\hspace{-0.5em}
\begin{subfigure}{
  \includegraphics[width = 0.15\textwidth]{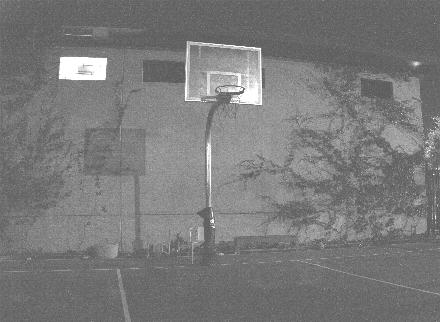}}
\end{subfigure}
\hspace{-0.5em}
\begin{subfigure}{
  \includegraphics[width = 0.15\textwidth]{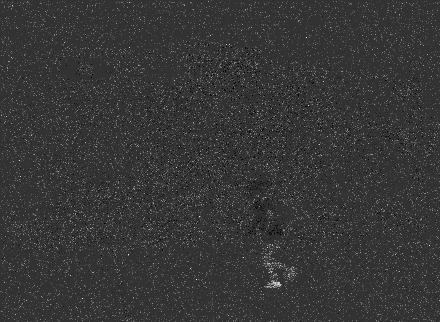}}
\end{subfigure}
\hspace{-0.5em}
\begin{subfigure}{
  \includegraphics[width = 0.15\textwidth]{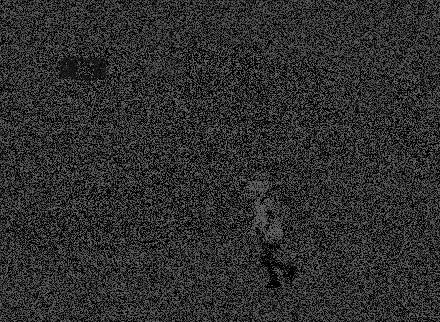}}
\end{subfigure}
\begin{subfigure}{
\hspace{.05cm}
\makebox[20pt]{\raisebox{20pt}{\rotatebox[origin=c]{90}{frame 90}}}%
  \includegraphics[width = 0.15\textwidth]{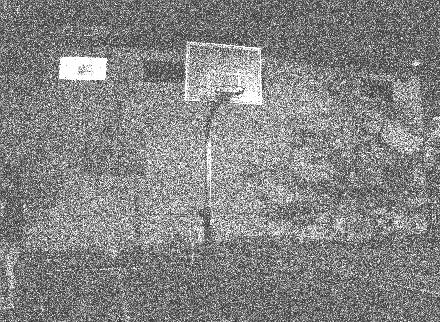}}
\end{subfigure}
\hspace{-0.5em}
\begin{subfigure}{
  \includegraphics[width = 0.15\textwidth]{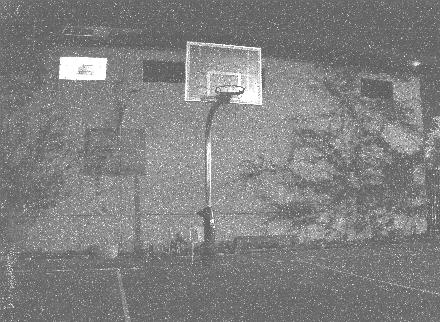}}
\end{subfigure}
\hspace{-0.5em}
\begin{subfigure}{
  \includegraphics[width = 0.15\textwidth]{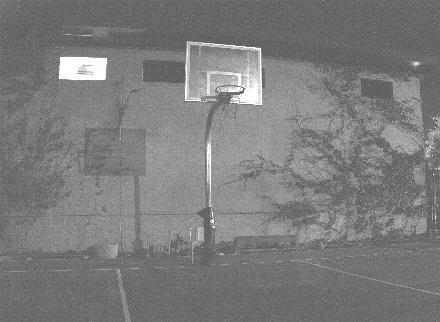}}
\end{subfigure}
\hspace{-0.5em}
\begin{subfigure}{
  \includegraphics[width = 0.15\textwidth]{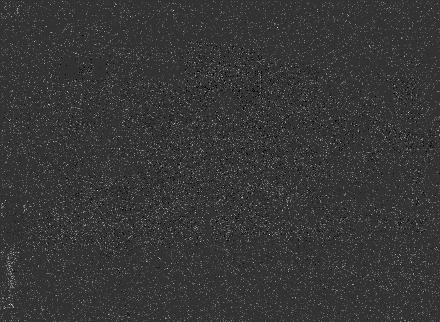}}
\end{subfigure}
\hspace{-0.5em}
\begin{subfigure}{
  \includegraphics[width = 0.15\textwidth]{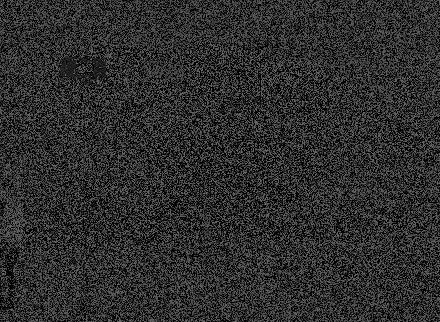}}
\end{subfigure}
\vspace{-1em}
\caption{Low light video frame 30, 60, 90 for M0001 ($\wh{\mb Z} = \wh{\mb L}+\wh{\mb S}-\wh{\mb D}$).}
\label{fig:denoising_video_M0001}
\end{figure}

\begin{figure}[h!]
\centering
\vspace{-2em}
\begin{subfigure}{
\makebox[20pt]{\raisebox{20pt}{\rotatebox[origin=c]{90}{}}}%
\makebox[0.16\textwidth]{$\mb D$}%
\makebox[0.16\textwidth]{$\wh{\mb L}+\wh{\mb S}$}%
\makebox[0.16\textwidth]{$\wh{\mb L}$}%
\makebox[0.16\textwidth]{$\wh{\mb S}$}%
\makebox[0.16\textwidth]{$\wh{\mb Z}$}%
}
\end{subfigure}
\vspace{-0.5em}
\begin{subfigure}{
\makebox[20pt]{\raisebox{15pt}{\rotatebox[origin=c]{90}{frame 30}}}%
\includegraphics[width = 0.15\textwidth]{plots/denoising_v04/frame_30_raw.jpeg}}
\end{subfigure}
\hspace{-0.5em}
\begin{subfigure}{
  \includegraphics[width = 0.15\textwidth]{plots/denoising_v04/frame_30_L_plus_S.jpeg}}
\end{subfigure}
\hspace{-0.5em}
\begin{subfigure}{
  \includegraphics[width = 0.15\textwidth]{plots/denoising_v04/frame_30_L.jpeg}}
\end{subfigure}
\hspace{-0.5em}
\begin{subfigure}{
  \includegraphics[width = 0.15\textwidth]{plots/denoising_v04/frame_30_S.jpeg}}
\end{subfigure}
\hspace{-0.5em}
\begin{subfigure}{
  \includegraphics[width = 0.15\textwidth]{plots/denoising_v04/frame_30_noise.jpeg}}
\end{subfigure}
\vspace{-0.5em}
\begin{subfigure}{
\makebox[20pt]{\raisebox{15pt}{\rotatebox[origin=c]{90}{frame 60}}}%
  \includegraphics[width = 0.15\textwidth]{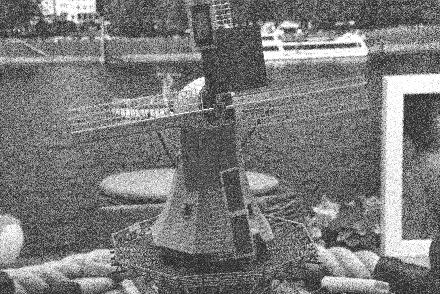}}
\end{subfigure}
\hspace{-0.5em}
\begin{subfigure}{
  \includegraphics[width = 0.15\textwidth]{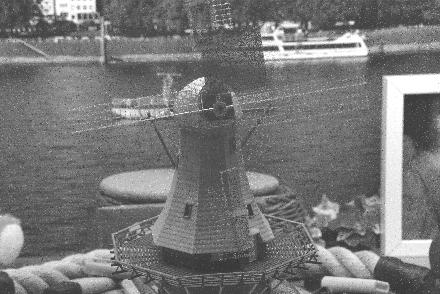}}
\end{subfigure}
\hspace{-0.5em}
\begin{subfigure}{
  \includegraphics[width = 0.15\textwidth]{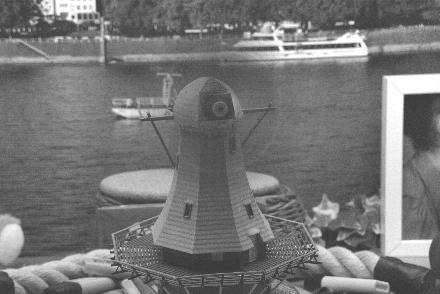}}
\end{subfigure}
\hspace{-0.5em}
\begin{subfigure}{
  \includegraphics[width = 0.15\textwidth]{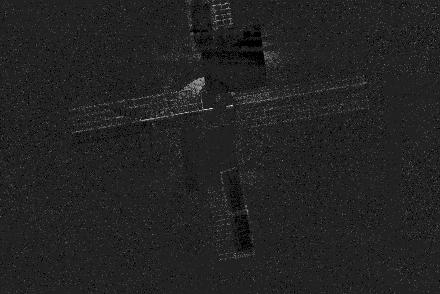}}
\end{subfigure}
\hspace{-0.5em}
\begin{subfigure}{
  \includegraphics[width = 0.15\textwidth]{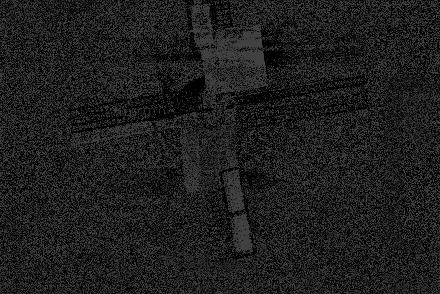}}
\end{subfigure}
\begin{subfigure}{
\hspace{.05cm}
\makebox[20pt]{\raisebox{15pt}{\rotatebox[origin=c]{90}{frame 90}}}%
  \includegraphics[width = 0.15\textwidth]{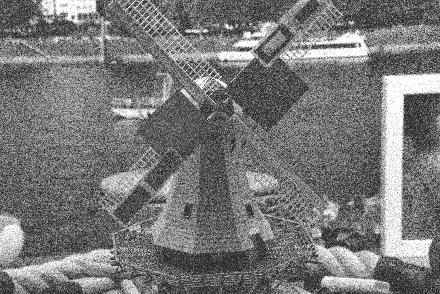}}
\end{subfigure}
\hspace{-0.5em}
\begin{subfigure}{
  \includegraphics[width = 0.15\textwidth]{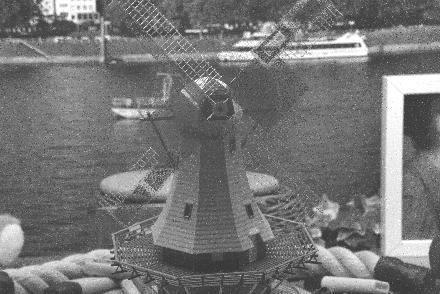}}
\end{subfigure}
\hspace{-0.5em}
\begin{subfigure}{
  \includegraphics[width = 0.15\textwidth]{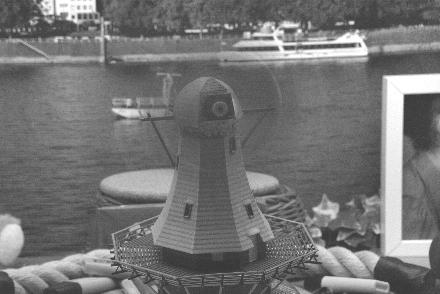}}
\end{subfigure}
\hspace{-0.5em}
\begin{subfigure}{
  \includegraphics[width = 0.15\textwidth]{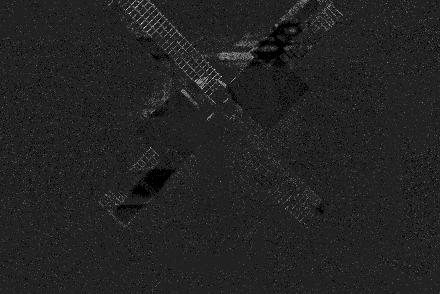}}
\end{subfigure}
\hspace{-0.5em}
\begin{subfigure}{
  \includegraphics[width = 0.15\textwidth]{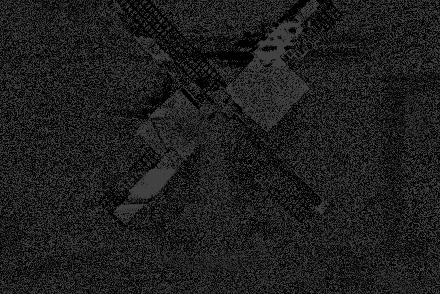}}
\end{subfigure}
\vspace{-1em}
\caption{Low light video frame 30, 60, 90 for M0004 ($\wh{\mb Z} = \wh{\mb L}+\wh{\mb S}-\wh{\mb D}$).}
\label{fig:denoising_video_M0004}
\end{figure}

\begin{figure}[h!]
\centering
\begin{subfigure}{
\makebox[20pt]{\raisebox{20pt}{\rotatebox[origin=c]{90}{}}}%
\makebox[0.16\textwidth]{$\mb D$}%
\makebox[0.16\textwidth]{$\wh{\mb L}+\wh{\mb S}$}%
\makebox[0.16\textwidth]{$\wh{\mb L}$}%
\makebox[0.16\textwidth]{$\wh{\mb S}$}%
\makebox[0.16\textwidth]{$\wh{\mb Z}$}%
}
\end{subfigure}
\vspace{-0.5em}
\begin{subfigure}{
\makebox[20pt]{\raisebox{15pt}{\rotatebox[origin=c]{90}{frame 30}}}%
\includegraphics[width = 0.15\textwidth]{plots/denoising_v09/frame_30_raw.jpeg}}
\end{subfigure}
\hspace{-0.5em}
\begin{subfigure}{
  \includegraphics[width = 0.15\textwidth]{plots/denoising_v09/frame_30_L_plus_S.jpeg}}
\end{subfigure}
\hspace{-0.5em}
\begin{subfigure}{
  \includegraphics[width = 0.15\textwidth]{plots/denoising_v09/frame_30_L.jpeg}}
\end{subfigure}
\hspace{-0.5em}
\begin{subfigure}{
  \includegraphics[width = 0.15\textwidth]{plots/denoising_v09/frame_30_S.jpeg}}
\end{subfigure}
\hspace{-0.5em}
\begin{subfigure}{
  \includegraphics[width = 0.15\textwidth]{plots/denoising_v09/frame_30_noise.jpeg}}
\end{subfigure}
\vspace{-0.5em}
\begin{subfigure}{
\makebox[20pt]{\raisebox{15pt}{\rotatebox[origin=c]{90}{frame 60}}}%
  \includegraphics[width = 0.15\textwidth]{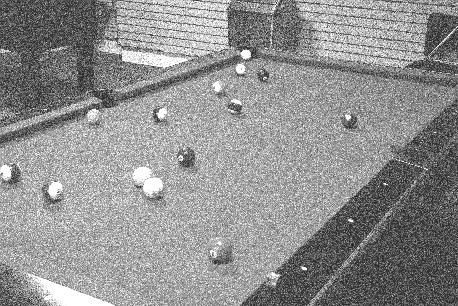}}
\end{subfigure}
\hspace{-0.5em}
\begin{subfigure}{
  \includegraphics[width = 0.15\textwidth]{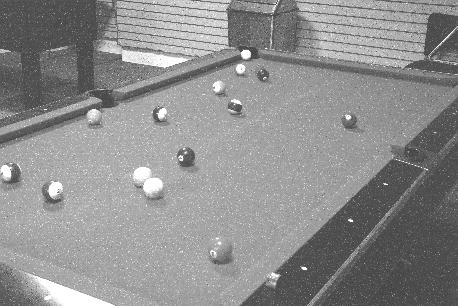}}
\end{subfigure}
\hspace{-0.5em}
\begin{subfigure}{
  \includegraphics[width = 0.15\textwidth]{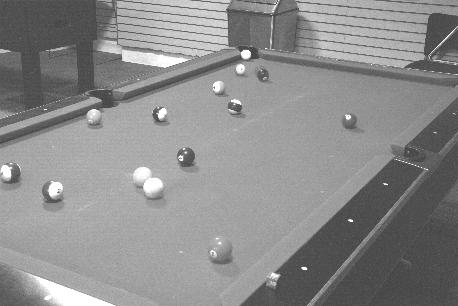}}
\end{subfigure}
\hspace{-0.5em}
\begin{subfigure}{
  \includegraphics[width = 0.15\textwidth]{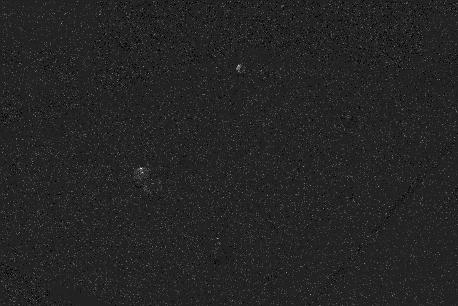}}
\end{subfigure}
\hspace{-0.5em}
\begin{subfigure}{
  \includegraphics[width = 0.15\textwidth]{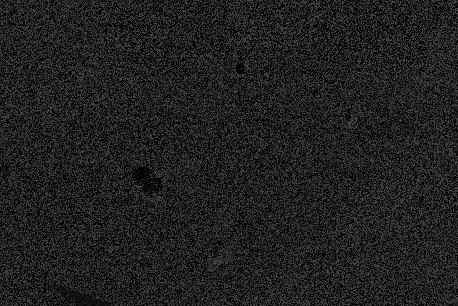}}
\end{subfigure}
\begin{subfigure}{
\hspace{.05cm}
\makebox[20pt]{\raisebox{15pt}{\rotatebox[origin=c]{90}{frame 90}}}%
  \includegraphics[width = 0.15\textwidth]{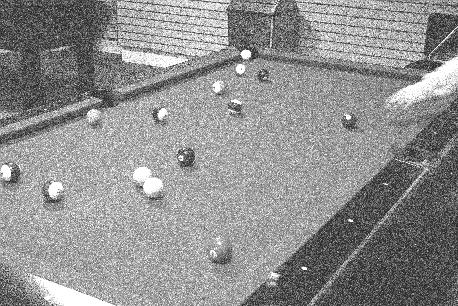}}
\end{subfigure}
\hspace{-0.5em}
\begin{subfigure}{
  \includegraphics[width = 0.15\textwidth]{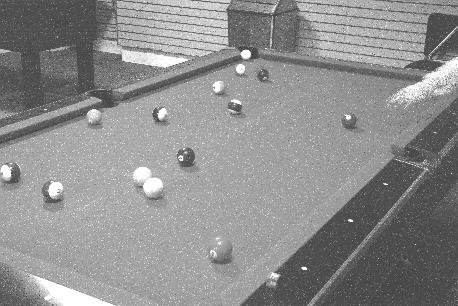}}
\end{subfigure}
\hspace{-0.5em}
\begin{subfigure}{
  \includegraphics[width = 0.15\textwidth]{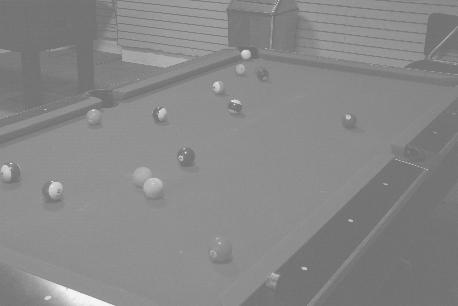}}
\end{subfigure}
\hspace{-0.5em}
\begin{subfigure}{
  \includegraphics[width = 0.15\textwidth]{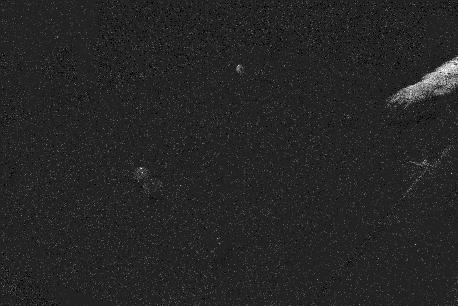}}
\end{subfigure}
\hspace{-0.5em}
\begin{subfigure}{
  \includegraphics[width = 0.15\textwidth]{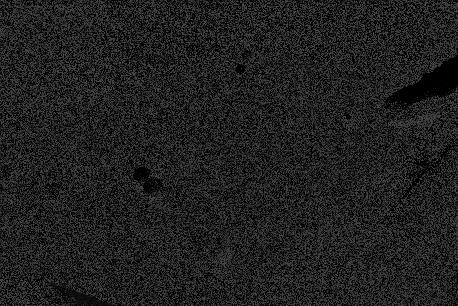}}
\end{subfigure}
\vspace{-1em}
\caption{Low light video frame 30, 60, 90 for M0009 ($\wh{\mb Z} = \wh{\mb L}+\wh{\mb S}-\wh{\mb D}$).}
\label{fig:denoising_video_M0009}
\end{figure}

We provide frame 30, 60, and 90 for the three videos in Figures \ref{fig:denoising_video_M0001}, \ref{fig:denoising_video_M0004}, and \ref{fig:denoising_video_M0009}.

\subsection{Additional Results for Section \ref{section:oct}}

\begin{figure}[h!]
\centering
\begin{subfigure}[$\mb D$]{
  \includegraphics[width = 0.09\textwidth]{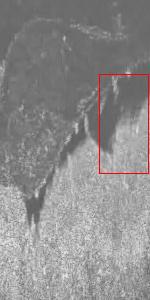}
  \label{fig:oct_raw150}}
\end{subfigure}
\hspace{-0.9em}
\begin{subfigure}[$\wh{\mb L}+\wh{\mb S}$]{
  \includegraphics[width = 0.09\textwidth]{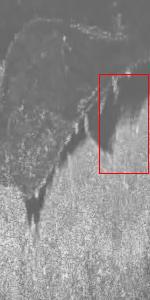}
\label{fig:oct_LS150}}
\end{subfigure}
\hspace{-0.9em}
\begin{subfigure}[$\wh{\mb L}$]{
  \includegraphics[width = 0.09\textwidth]{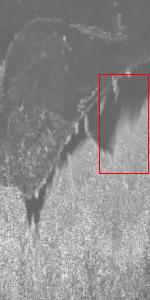}
\label{fig:oct_L150}}
\end{subfigure}
\hspace{-0.9em}
\begin{subfigure}[$\wh{\mb S}$]{
  \includegraphics[width = 0.09\textwidth]{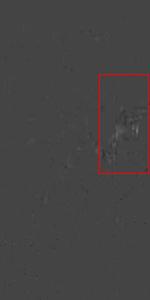}
\label{fig:oct_S150}}
\end{subfigure}
\hspace{-0.9em}
\begin{subfigure}[$\wh{\mb Z}$]{
  \includegraphics[width = 0.09\textwidth]{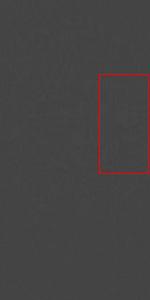}
\label{fig:oct_noise150}}
\end{subfigure}
\hspace{-0.9em}
\begin{subfigure}[$\mb D$]{
  \includegraphics[width = 0.09\textwidth]{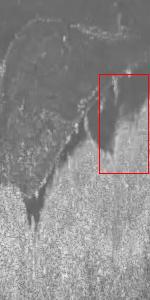}
  \label{fig:oct_raw200}}
\end{subfigure}
\hspace{-0.9em}
\begin{subfigure}[$\wh{\mb L}+\wh{\mb S}$]{
  \includegraphics[width = 0.09\textwidth]{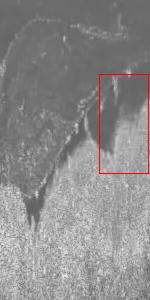}
\label{fig:oct_LS200}}
\end{subfigure}
\hspace{-0.9em}
\begin{subfigure}[$\wh{\mb L}$]{
  \includegraphics[width = 0.09\textwidth]{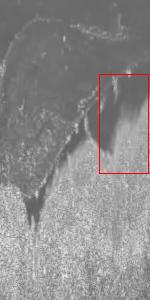}
\label{fig:oct_L200}}
\end{subfigure}
\hspace{-0.9em}
\begin{subfigure}[$\wh{\mb S}$]{
  \includegraphics[width = 0.09\textwidth]{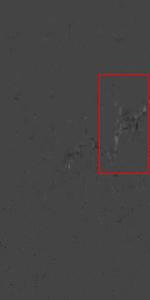}
\label{fig:oct_S200}}
\end{subfigure}
\hspace{-0.9em}
\begin{subfigure}[$\wh{\mb Z}$]{
  \includegraphics[width = 0.09\textwidth]{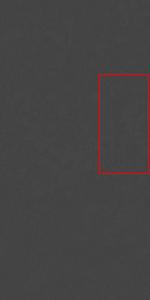}
\label{fig:oct_noise200}}
\end{subfigure}
\vspace{-1em}
\caption{OCT, a-e): frame 150, f-j): frame 200 ($\wh{\mb Z} = \wh{\mb L}+\wh{\mb S}-\wh{\mb D}$).}
\vspace{-1em}
\label{fig:oct_more}
\end{figure}

Results for frame 150 and 200 are presented in Figure \ref{fig:oct_more}.

\subsection{Settings and Additional Results for Section \ref{section:optimal_mu}}
Table \ref{tb:parameters} below lists the settings for this set of experiments. 

\begin{table}[h!]
\vspace{-1em}
  \caption{Parameters for Simulation: $\rho_S = 0.1$, $|(\mb S_0)_{(i,j)\in \Omega}|=0.05$}
  \label{tb:parameters}
  \centering
  \begin{tabular}{cccc}
    \toprule
    $n_1$     & $n_2$     & $\rho_L$ & $\sigma$ \\
    \midrule
    $\{200,300,\ldots,1000\}$ &$n_2 = n_1$ & 0.1 & 0.01\\
    $\{300,\ldots,1000\}$ &300 & 0.1 & 0.01\\
    300&300&$\{0.05,0.1,\ldots,0.5\}$&0.01\\
    300&300&0.1&$\{0.005,0.01,\ldots,0.05\}$\\
    \bottomrule
  \end{tabular}
  \vspace{-0.5em}
\end{table}

\begin{figure}[h!]
\centering
\begin{subfigure}[vary $n=n_1=n_2$]{
  \includegraphics[width = 0.22\textwidth]{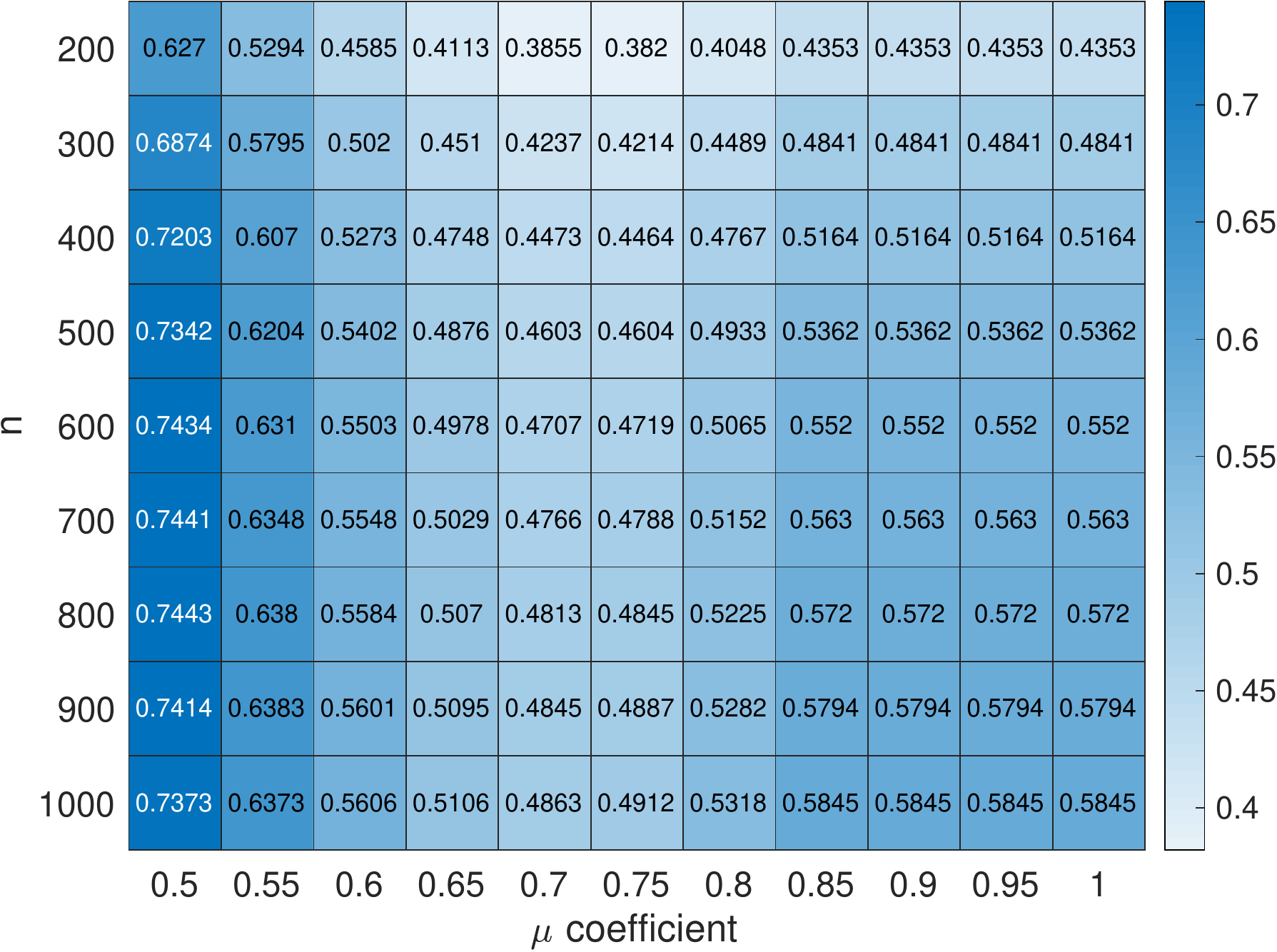}
\label{fig:vary_mu_n_relative}}
\end{subfigure}
\begin{subfigure}[vary $n_1$]{
  \includegraphics[width = 0.22\textwidth]{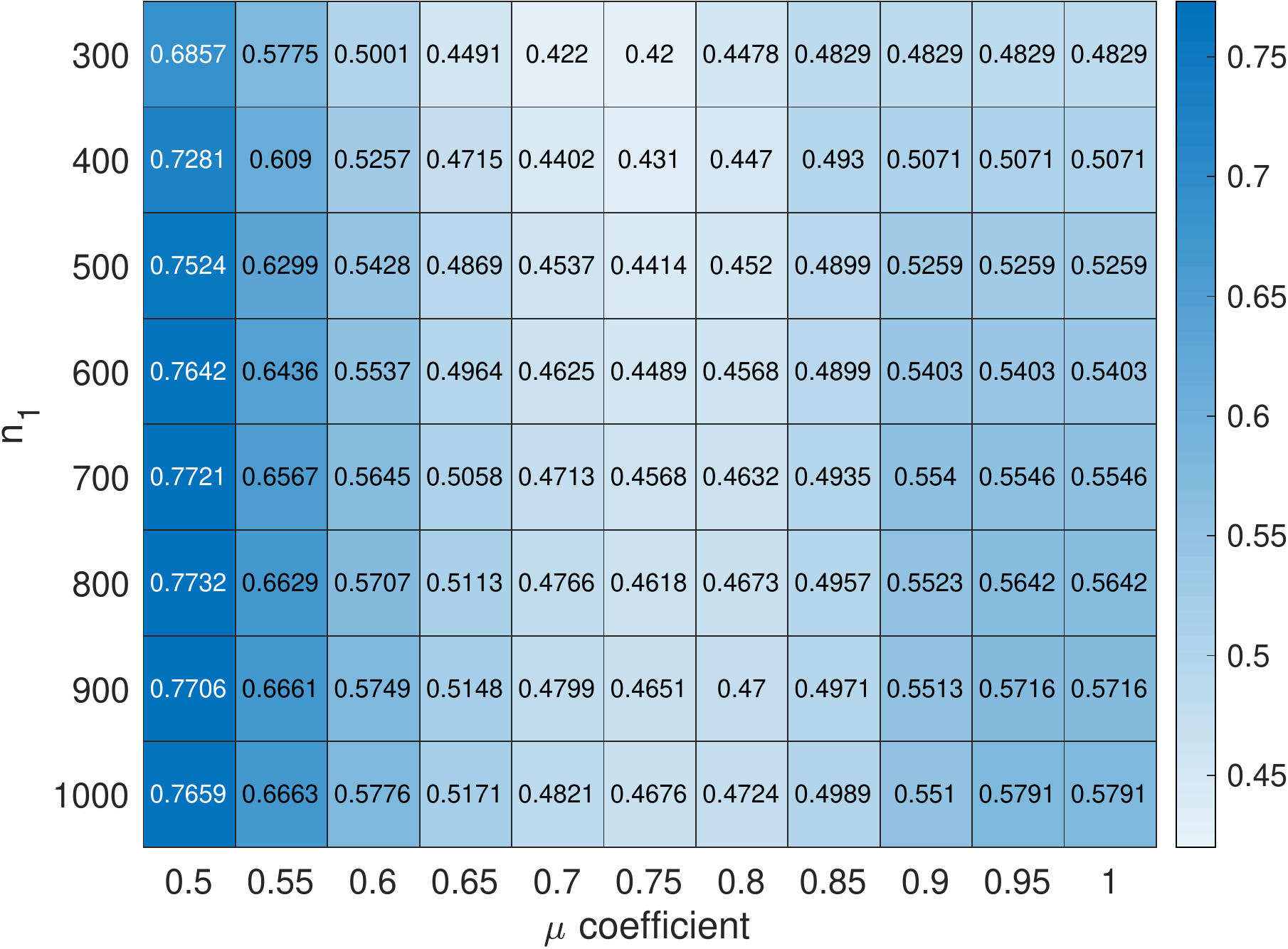}
\label{fig:vary_mu_p_relative}}
\end{subfigure}
\begin{subfigure}[vary $\rho_L$]{
  \includegraphics[width = 0.22\textwidth]{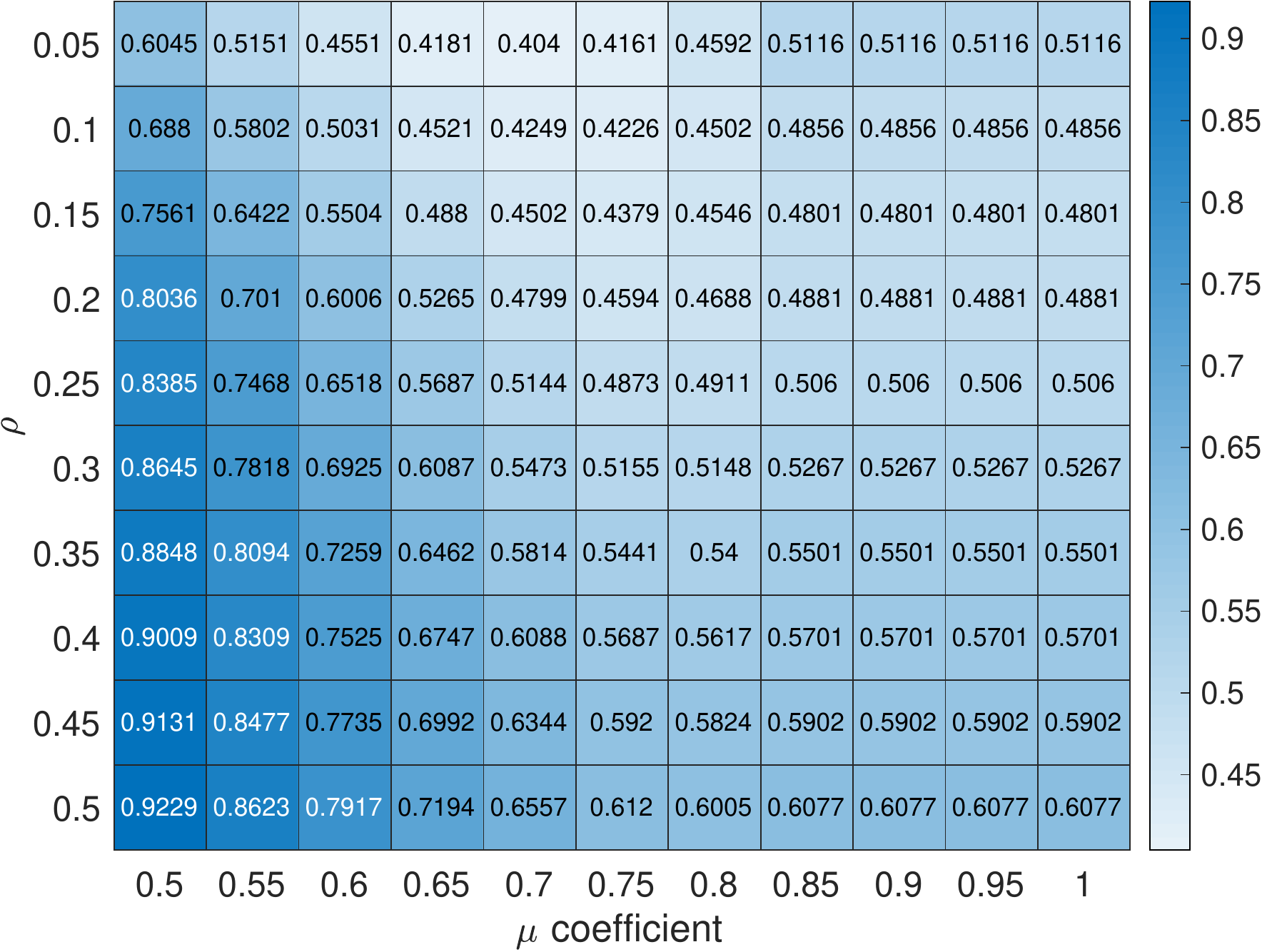}
\label{fig:vary_mu_rho_relative}}
\end{subfigure}
\begin{subfigure}[vary $\sigma$]{
  \includegraphics[width = 0.22\textwidth]{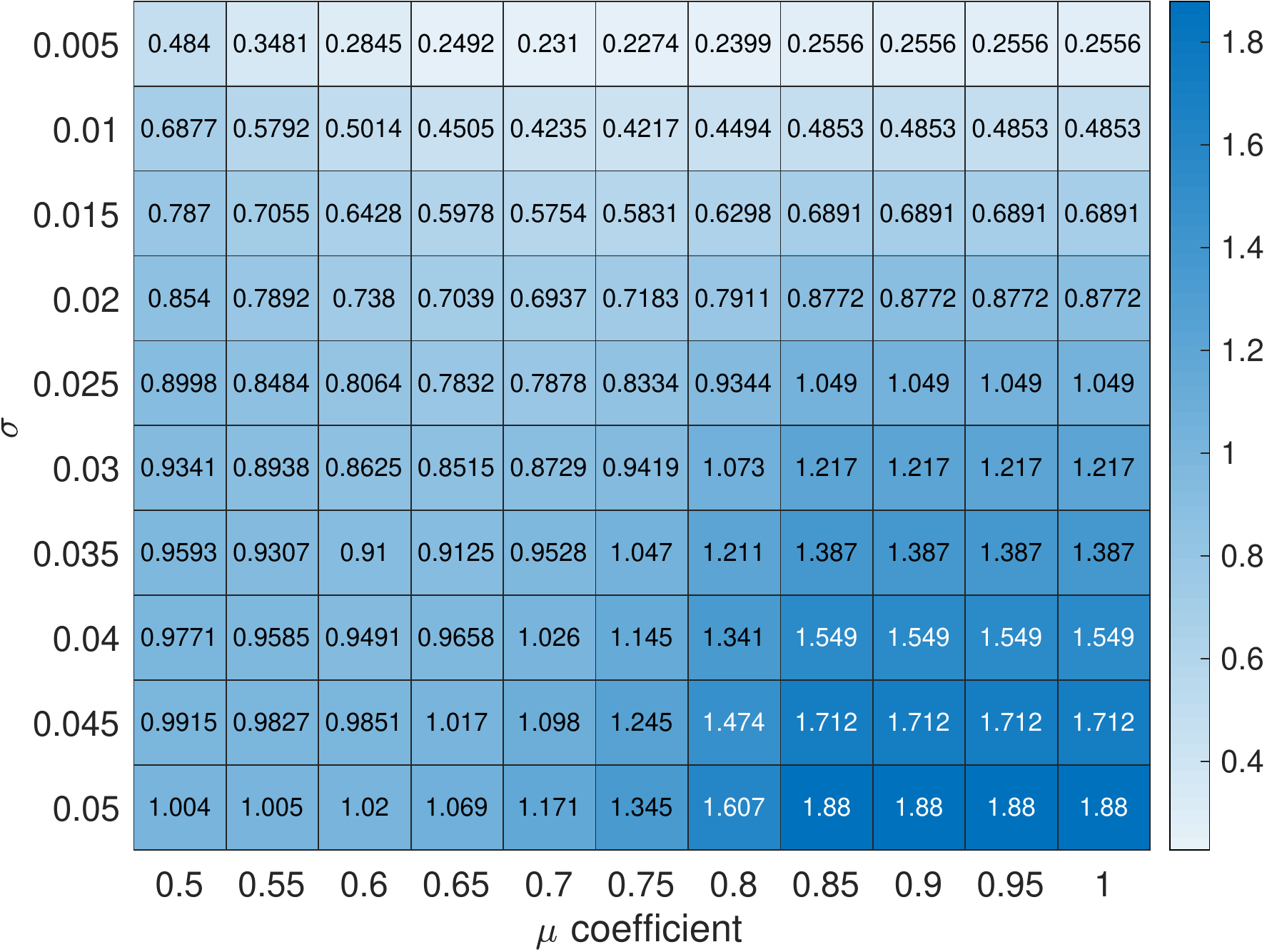}
\label{fig:vary_mu_sigma_relative}}
\end{subfigure}
\vspace{-1em}
\caption{$\eta_{\mathrm{rel}} (\mu)$ under different varying parameters}
\vspace{-1em}
\label{fig:optimal_mu_relative}
\end{figure}

We also provide heatmaps for the relative recovery error, i.e. the 10-average of $\eta_{\mathrm{rel}}(\mu):=\| (\wh {\mb L}(\mu), \wh {\mb S}(\mu)) - (\mb L_0,\mb S_0)\|_F/\| (\mb L_0,\mb S_0)\|_F$, in Figure \ref{fig:optimal_mu_relative}. 

\end{document}